\documentclass[12pt,letterpaper]{article}
\usepackage[margin=1in]{geometry}
\usepackage{amsthm,amsfonts,amsmath,amssymb}
\usepackage[mathscr]{eucal}
\usepackage{enumitem}
\usepackage{algorithm}
\usepackage{algpseudocode}
\usepackage{tikz}
\usetikzlibrary{decorations.pathreplacing, calc}
\usepackage{multirow}
\usepackage{booktabs}
\usepackage{graphicx}
\usepackage{makecell}
\usepackage[labelfont=bf]{caption}
\usepackage{subcaption}
\usepackage{placeins}
\usepackage{float}
\usepackage[numbers,sort&compress,round]{natbib}
\usepackage{jabbrv}
\usepackage[colorlinks=true,linkcolor=red,citecolor=blue,urlcolor=blue]{hyperref}

\newtheorem{theorem}{Theorem}[section]
\newtheorem{definition}{Definition}[section]
\newtheorem{proposition}[theorem]{Proposition}

\newtheorem{assumption}[theorem]{Assumption}

\newtheorem{example}[theorem]{Example}

\newcommand\ind{\protect\mathpalette{\protect\independenT}{\perp}}
\def\independenT#1#2{\mathrel{\rlap{$#1#2$}\mkern2mu{#1#2}}}

\newcommand{\BE}{\mathbb{E}}

\newcommand{\mart}{\mathcal{M}}

\newcommand{\X}{\mathcal{X}}
\newcommand{\Y}{\mathcal{Y}}
\newcommand{\Z}{\mathcal{Z}}

\newcommand{\M}{f}

\newcommand{\bbr}{\mathbb{R}}  
\newcommand{\bbn}{\mathbb{N}}
\newcommand{\bbp}{\mathbb{P}}

\newcommand{\bbe}{\mathbb{E}}

\newcommand{\be}{\begin{equation}}
\newcommand{\bew}{\begin{equation*}}
\newcommand{\eew}{\end{equation*}}

\newcommand{\ci}{\cite}

\begin{document}

\title{AICO: Feature Significance Tests for Supervised Learning}
\author{Kay Giesecke,\,\, Enguerrand Horel,\,\, Chartsiri Jirachotkulthorn\footnote{Giesecke (corresponding author, Kay.Giesecke@hpi.de) is with the Hasso Plattner Institute, University of Potsdam, 14482 Potsdam, Germany. Horel (enguerrand.horel@gmail.com) is with Upstart, Inc., San Mateo, CA 94402. Jirachotkulthorn (cjiracho@stanford.edu) is with the Institute for Computational and Mathematical Engineering, Stanford University, Stanford, CA 94305. We are very grateful to Pulkit Goel and Chenyu Song for exceptional research assistance. We thank Rakshitha Ireddi and Yashwanth Devavarapu for comments. We are also grateful for financial support from the Stanford Institute for Human-Centered Artificial Intelligence (HAI). E.H. is employed by Upstart, a company that might commercially use the methods described in this work; E.H. might own stock or options in Upstart, Inc. The other authors declare no competing interest. A Python code package is available at \url{https://github.com/ji-chartsiri/AICO}. A preliminary version of this work appeared under the title ``Computationally Efficient Feature Significance and Importance for Predictive Models'' (ICAIF 2022).}}
\date{July 31, 2026}
\maketitle

\begin{abstract}

Machine learning is central to modern science, industry, and policy, yet its predictive power often comes at the cost of transparency: we rarely know which input features drive a model’s predictions. Without such understanding, researchers cannot draw reliable conclusions, practitioners cannot ensure fairness or accountability, and policymakers cannot trust or govern model-based decisions. Existing tools for assessing feature influence are limited; most lack statistical guarantees, and many require costly retraining or surrogate modeling, making them impractical for large modern models. We introduce AICO, a broadly applicable framework that turns model interpretability into an efficient statistical exercise. AICO tests whether each feature contributes to predictive performance by masking its information and measuring the resulting change. The method provides exact, finite-sample feature $p$-values and confidence intervals for feature importance through a simple, non-asymptotic hypothesis testing procedure. It requires no retraining, surrogate modeling, or distributional assumptions, making it feasible for large-scale algorithms. In both controlled experiments and real applications, from credit scoring to mortgage-behavior prediction, AICO identifies variables that contribute to model behavior, providing a scalable and statistically principled path toward transparent and trustworthy machine learning.

\end{abstract}


\section*{Introduction}

\phantomsection
\label{intro}

Machine learning has become integral to modern science, industry, and public policy. Algorithms now predict molecular activity, identify disease markers, forecast financial defaults, and guide the allocation of social resources. Yet even as these systems achieve extraordinary predictive accuracy, their inner workings remain largely opaque. We often do not know which input features truly shape a model’s predictions, which contribute little to predictive performance, or which may encode unwanted biases. This opacity limits scientific understanding, hinders regulatory oversight, and undermines public trust in data-driven decision making. Despite extensive progress in explainable artificial intelligence, most tools for measuring feature importance lack statistical guarantees or depend on computationally costly retraining or surrogate modeling procedures, restricting their use in today’s large, complex models. As a result, reconciling interpretability with scalable inference remains an open challenge for modern machine-learning practice.


We introduce AICO, a tractable method for feature significance testing in any supervised learning model, whether for regression or classification. AICO transforms the question of interpretability into one of formal statistical inference: it tests whether a feature contributes meaningfully to a model’s predictive performance, using only the model’s fitted predictions and observed outcomes. AICO constructs test statistics based on the change in model performance when a feature is masked. It is model- and distribution-agnostic: it makes no assumptions about the model class, data distribution, or estimand (for example, conditional mean or quantile). The method does not rely on retraining or surrogate models, making it well suited for analyzing complex, high-dimensional algorithms that are costly to train. It bridges a gap between machine-learning practice and classical statistical inference, offering a tractable route to transparent, trustworthy prediction in today’s large-scale applications.\footnote{An open-source Python package is available at \url{https://github.com/ji-chartsiri/AICO}.}

The AICO feature effect—the change in model performance due to masking—quantifies a feature’s importance for each individual sample. For a null feature, these effects form a distribution across test samples whose median is non-positive. A uniformly most powerful randomized sign test achieves finite-sample exactness: its size (type-I error probability) exactly matches the nominal level (e.g., $5\%$). The procedure produces exact, randomized feature $p$-values whose conditional distribution, given the data, is uniform on an interval that can be reported in practice. It is simple, robust, and computationally efficient, relying on minimal assumptions and remaining valid under broad conditions. These properties enable valid inference on a model’s features without relying on large-sample approximations or parametric assumptions.



The median feature effect serves as a natural, robust, and global measure of feature importance. 
The choice of masking specification gives rise to alternative importance scores: a conditional score measures the incremental contribution of a feature given the others, whereas an unconditional score quantifies its stand-alone contribution. 
We develop randomized, exact-coverage confidence intervals for the median that are dual to the hypothesis test. 
Together with the test and corresponding $p$-values, these intervals provide a unified framework for feature inference in any learning model.

Extensive experiments on synthetic ``known-truth'' tasks demonstrate the statistical and computational performance of our approach in neural-network regression and classification settings. 
We compare AICO with powerful, model-agnostic, and non-asymptotic tests by \ci{berrett,tansey2018holdout,Lei2018predictive}, as well as with the model-X knockoff procedures of \ci{candes2018panning,deepknockoffs} for variable selection under false discovery control. 
Our test identifies relevant features comparably to these alternatives while producing fewer discoveries among features that are null in the data-generating process, all while reducing computational cost by orders of magnitude.
The test and median feature-importance measure remain stable across a range of feature distributions, including settings with strong correlations and interactions. 
The median is robust to noise, producing stable feature rankings. 




Empirical analyses demonstrate the ability of our framework to inform high-stakes decisions with legal, economic, and regulatory implications. In a credit scoring task using ensembles of gradient-boosting models, AICO identifies several statistically significant features describing past payment behavior and social factors. 
In a mortgage state–transition prediction task based on a recently proposed set–sequence model 
\cite{elliot} and trained on extensive panel data, AICO detects time-varying macroeconomic variables such as mortgage rates, as well as features summarizing past state transitions, as significant predictors of borrower behavior.


\subsection*{Related literature}\label{lit}
There is a large body of work developing algorithmic feature-importance measures; see \ci{burkart-huber} for an overview and \ci{binyu} for a unifying framework for interpretability methods. Popular examples are based on Shapley values (e.g., SHAP \ci{lundberg2017unified}), local approximations (e.g., LIME \ci{ribeiro2016should}), and permutation methods (e.g., \ci{breiman,fisher-rudin}). These methods provide useful insight into feature contribution but typically lack formal statistical guarantees and associated $p$-values, in contrast to the importance measures delivered by our hypothesis-testing framework. Moreover, algorithmic importance measures can be sensitive to feature correlation \ci{verdinelli-wasserman,hooker}, whereas the conditional AICO score is robust to both correlation and interaction. Finally, we show that AICO achieves significant improvements in computational efficiency relative to widely used algorithmic approaches.



The statistical and econometric literature has long studied significance testing for variables in nonparametric regression models. 
\ci{fan-li,lavergne2000nonparametric,ait-sahalia-bickel,delgado-manteiga,lavergne2015} and others develop kernel-based asymptotic testing procedures. 
\ci{white2001statistical,horel-giesecke,fallahgoul} and others develop gradient-based asymptotic tests using neural networks. 
Coleman et al.~\ci{mentch} develop permutation-based asymptotic tests for random forests; see also Mentch and Hooker~\ci{mentch-hooker} for an alternative approach based on $U$-statistics. 
In contrast, AICO provides a model-agnostic, non-asymptotic framework that delivers exact $p$-values and confidence intervals and applies to a broad range of data structures. 


Pursuing a model-agnostic approach, \ci{candes2018panning,berrett,tansey2018holdout,janson,bellot,wright} and others develop non-asymptotic tests for conditional independence in the model-X knockoff framework of \ci{candes2018panning}. 
\ci{pnas-test,lundborg,williamson2,williamson} and others study asymptotic tests for conditional {\it mean} independence. 
These procedures assess feature significance through the change in predictive performance when a feature is removed from the model. 
Related feature-importance measures based on this principle are studied—often via asymptotic analysis—in \ci{dai-shao-chen,gan,Lei2018predictive,rinaldo,williamson2,williamson}. 


AICO studies an alternative formulation of this incremental-value principle based on feature masking rather than feature removal, while delivering exact, non-asymptotic inferential guarantees. 
This masking strategy avoids the repeated model retraining or auxiliary modeling required by existing approaches.\footnote{Methods based on removing features typically require retraining the model for each feature under consideration, which is prohibitive for feature-rich models. The approach of \ci{gan} requires retraining the model once using a custom training algorithm. Model-X knockoff procedures require constructing an auxiliary model of the joint feature distribution, a task that can be as challenging as fitting the predictive model itself; see \ci{bates} for fundamental limitations.} 
As a result, AICO achieves substantial gains in computational efficiency and enables inference for large, feature-rich models that are otherwise computationally challenging. 
Our experiments show that these computational gains do not come at the expense of statistical performance.

Like several of the aforementioned methods, AICO employs data splitting into training and testing samples to facilitate inference after model selection; see \ci{rinaldo} and the references therein. 
While data splitting can reduce statistical efficiency, it is widely used in machine learning to evaluate the out-of-sample performance of fitted models, and our approach aligns with this standard practice. 
Moreover, a power analysis allows the training and testing sample sizes to be chosen to achieve a desired level of power. 
Even under unfavorable alternatives, the test sample sizes required for high power remain small relative to typical data sizes in machine learning. 
Finally, the AICO framework admits a cross-fitting implementation, which attains comparable performance while eliminating reliance on a single data split.




\section{Feature effect}\label{setup}
We consider a distribution-free setting, where the random pair $Z= (X,Y)$ of $d$-dimensional feature $X=(X^1,\ldots,X^d)\in\X$ and response $Y\in\Y$ has an unknown distribution. In a typical regression formulation with tabular data, $\X\subseteq\bbr^d$ and $\Y=\mathbb{R}$. We observe data $Z_1,\ldots,Z_m$, where the $Z_i = (X_i, Y_i)\in\Z=\X\times\Y$ are independent copies of $Z$. 


We assume a random split of the index set $[m]=\{1,\ldots,m\}$ into non-intersecting subsets $I_1$ and $I_2$. 
The dataset $D_1=\{Z_i\}_{i\in I_1}$ is used for training, while $D_2=\{Z_i\}_{i\in I_2}$ is reserved for testing. 
Such sample splitting is standard in machine learning and, in our setting, enables valid post-selection inference.\footnote{Applications of sample splitting to testing problems have a long history; Cox~\ci{cox-75} is an early reference. See \ci{rinaldo} and \ci{lundborg} for recent developments.} 
The results below provide guidance on selecting the size of $I_2$ to optimize testing power. 
As discussed in Appendix~\ref{cross-fitting}, our approach naturally extends to cross-fitting with multiple data splits, allowing each sample $Z_i$ to be used for both training and testing while maintaining valid inference.

We also take as given a model (algorithm) $\M: \X\to\mart$ trained on $D_1$, which is to be evaluated. 
Throughout, we treat $\M$ as a non-random function of $D_1$. 
We impose no restrictions on the form of the estimand: $\M$ may represent the conditional mean, a quantile, or any other statistic of the response. 
Nor do we assume a particular training criterion or optimization procedure for fitting $\M$.

We test the statistical significance of a feature $\ell\in[d]$ under the fitted $\M$. 
When $\M$ is linear, classical asymptotic procedures such as the $t$-test are available. 
In contrast, we develop a simple, non-asymptotic, and model-agnostic procedure that does not depend on a particular model structure and delivers valid inference for $\M$ of arbitrary form. 
As is standard in post-selection inference with data splitting, testing based on $D_2$ is conducted conditional on the training data $D_1$ and on the training and selection procedures used to construct $\M$ (e.g., \ci{rinaldo}).

\subsection{Main ideas}\label{main-ideas}
We sketch our approach in a standard regression setting in which $\M(x)$ estimates the conditional mean $\bbe(Y| X = x)$ of a continuous, real-valued response $Y$.\footnote{In this section, all expectations are taken with respect to the (unknown) distribution of $Z = (X,Y)$ from which the data are drawn.} 
Section~\ref{general} extends the framework to other types of responses and estimands. 
Following a growing literature and prediction-focused downstream applications of machine-learning models, our approach ties the statistical significance of a feature to its incremental predictive power \ci{pnas-test,Lei2018predictive,lundborg,rinaldo,williamson2,williamson}.
Rather than assessing this predictive contribution through feature removal and model retraining, we propose a masking-based strategy for measuring a feature's incremental predictive value. We posit that if feature $\ell$ has predictive power, then masking its values systematically affects prediction errors. 
We develop a simple, non-asymptotic hypothesis testing procedure to detect this effect.


We first measure this effect at the level of an individual test sample $i \in I_2$ with feature–response pair $(X_i, Y_i)$. 
Consider the {\it masked} feature vector
\begin{align}\label{masked}
m^\ell(X_i) = (X_i^1, \ldots, v^\ell, \ldots, X_i^d),
\end{align}
where the observed value $X_i^\ell$ of feature $\ell$ is replaced by a reference value $v^\ell \neq X_i^\ell$. 
A natural choice for $v^\ell$ is a measure of central tendency of $\ell$ in the training data $D_1$, such as the mean when $\ell$ has a continuous distribution on an interval. 
The model output $\M(m^\ell(X_i))$ represents the prediction when the masked feature vector (\ref{masked}) is used as input. 
The prediction error can be quantified by $\{\M(m^\ell(X_i)) - Y_i\}^2$, which is natural when $\M$ is trained using a squared-error criterion.
The {\it feature effect} for sample $i$ is the incremental error due to masking $\ell$, 
\begin{align}\label{residuals}
\{\M(m^\ell(X_i)) - Y_i\}^2 - \{\M(X_i) - Y_i\}^2.
\end{align}

The sign and magnitude of the feature effect (\ref{residuals}) characterize the change in predictive accuracy induced by masking feature $\ell$ for sample $i$ under $\M$. 
If Eq.~(\ref{residuals}) is positive, masking $\ell$ reduces accuracy, indicating that $\ell$ contributes to the fitted model's predictive performance for sample $i$.
In this case, larger values correspond to greater reductions in accuracy due to masking and hence greater predictive power of $\ell$. 
Thus, the magnitude of (\ref{residuals}) quantifies the effect size of feature $\ell$ at the sample level.

Moving beyond an individual sample, we deem feature $\ell$ significant when masking it systematically degrades predictive accuracy across the test set. Consider the median of the sample-level feature effects (\ref{residuals}) across $I_2$. A positive median is evidence that masking $\ell$ systematically reduces predictive accuracy and hence that $\ell$ is significant. Conversely, a non-positive median is treated as absence of evidence that $\ell$ is significant. A sign test of this property delivers exact $p$-values for feature significance (see Section~\ref{sec-test}). The median feature effect also serves as a point estimate of feature importance. Larger positive values correspond to greater reductions in predictive accuracy due to masking and hence greater predictive power of $\ell$. Confidence intervals for feature importance are obtained by inverting the test (see Section~\ref{sec-imp}).





An analysis of incremental prediction errors also underlies significance testing based on conditional mean independence \ci{pnas-test,Lei2018predictive,lundborg,rinaldo,williamson2,williamson}. 
This approach introduces a reduced feature vector $X^{-\ell} = (X^1, \ldots, X^{\ell-1}, X^{\ell+1}, \ldots, X^d)$ that excludes $X^\ell$. 
Feature $\ell$ is treated as having no incremental predictive value if the response $Y$ is conditionally mean independent of $\ell$ given $X^{-\ell}$, i.e., if
\begin{align}\label{cmi}
\{\bbe(Y \mid X^{-\ell}) - Y\}^2 - \{\bbe(Y \mid X) - Y\}^2
\end{align}
has mean zero. 
The quantity (\ref{cmi}) measures the incremental squared prediction error induced by removing $\ell$ from the feature set. 
Our feature effect (\ref{residuals}) is based on the same incremental prediction-error principle, but differs from (\ref{cmi}) in two important respects.\footnote{Note that Eq.~(\ref{residuals}) estimates Eq.~(\ref{cmi}) at the sample level if $\M(m^\ell(X))$, the predicted mean response after masking $\ell$, approximates $\bbe(Y \mid X^{-\ell})$, the true mean response after removing $\ell$.} 
First, Eq.~(\ref{residuals}) measures the change in prediction error induced by masking $\ell$ rather than removing it, thereby avoiding the need to construct an auxiliary model for $\bbe(Y \mid X^{-\ell})$. 
This yields substantial computational gains, especially when many features are analyzed or $\M$ is large. It also avoids the additional randomness introduced by auxiliary model fitting. 
Second, our approach focuses on sample-level prediction errors under masked features, rather than population-level errors. 
This sample-level formulation naturally supports an assumption-light, non-asymptotic framework for significance testing. 

\subsection{General formulation}\label{general}
We formulate our approach in greater generality, allowing for real and categorical responses $Y$ and estimands beyond the conditional mean such as conditional quantiles. Recall our premise: if feature $\ell$ is predictive, then masking its true values will systematically impact prediction errors under $\M$. The feature effect measures this impact at the sample level; its distribution across the test set will be used to construct test statistics. 

We begin by specifying a masking function $m^\ell: \X\to\X$ that masks the $\ell$th and possibly other components of an observed feature vector $X_i$ with suitable reference values, and an unmasking function $u^\ell: \X\to\X$ that restores some or all of the true feature values. The reference values are allowed to depend on the training data $D_1$, in which case $m^\ell$ and $u^\ell$ are random functions. The errors associated with the predictions $\M(m^\ell(X_i))$ and $\M(u^\ell(X_i))$ for masked and unmasked feature vectors, respectively, will then be used to define the {feature effect} for sample $i$. Before doing that, we provide two canonical examples for the masking and unmasking functions and discuss the choice of the reference values. The first example reflects the formulation in Section \ref{main-ideas} above.

\begin{example}\label{ex1}
For all $x=(x^1,\ldots,x^d)\in\X$, let
\begin{align}\label{conditional}
m^\ell(x) = (x^1,\ldots,v^\ell,\ldots,x^d),\quad u^\ell(x) = x.
\end{align} 
Feature $\ell$ is masked with a reference value $v^\ell$ so that the prediction $\M(m^\ell(X_i))$ is based on $v^\ell$ rather than the true value $X^\ell_i$. To unmask, the true value is {added back in}—an operation that gives rise to the name \emph{AICO (Add-In COvariates)}. 
Since masking affects only $\ell$, Eq.~(\ref{conditional}) can be used to construct a conditional feature effect, which measures the incremental contribution of $\ell$ given the others. 


\end{example}

\begin{example}\label{ex2}
For all $x\in\X$, let
\begin{align}\label{unconditional}
m^\ell(x) = (v^1,\ldots,v^d),\quad u^\ell(x) = (v^1, \ldots, x^\ell,\ldots,v^d).
\end{align} 
Each feature $j\in[d]$ is masked with a reference value $v^j$. 
Adding the true value of feature $\ell$ back into this context performs the same {AICO} operation. 
By masking all features, Eq.~(\ref{unconditional}) can be used to construct an unconditional feature effect, which measures $\ell$'s stand-alone contribution.
\end{example}

The masking and unmasking functions in (\ref{conditional}) and (\ref{unconditional}) require the specification of reference values $v^j$. These values should provide a neutral baseline for feature-effect measurement while mapping masked inputs to regions of the feature space where the model can be expected to provide reliable predictions. This suggests selecting suitable measures of central tendency from the observed values of each feature in the training set $I_1$, ensuring that masked feature values remain in the feature's support.\footnote{Even when a reference value is marginally in the feature support, masking can still produce covariate combinations that force the model to extrapolate into regions with limited data support. The risk of such off-manifold evaluations can be mitigated through localized or conditional masking schemes that construct reference values using observations with similar feature profiles. For a concrete example, see Appendix \ref{app-regress}.} Specifically, if feature $j$ has a continuous distribution with interval support in $\bbr$, a natural choice for $v^j$ is the mean of $\{X^j_k\}_{k\in I_1}$. If feature $j$ has a discrete distribution with support set $s^j$, such as a categorical feature, a natural choice is the adjusted mode $\arg\max_{w\in s^j\setminus\{x^j\}} p^j(w)$, where $x^j$ is the feature value being masked and $p^j$ denotes the probability mass function of $\{X^j_k\}_{k\in I_1}$. For example, if $j$ is a dummy variable, then $v^j = 1 - x^j$. The adjusted mode generalizes naturally to categorical variables encoded by multiple dummies and ensures that the masked feature value differs from the original value while remaining in the feature's support. Our numerical experiments in Section~\ref{sec-numerical} and Appendix~\ref{robustness} highlight the effectiveness and robustness of these choices for $v^j$.

We construct the feature effect using masking and unmasking functions, generalizing (\ref{residuals}). 
Formally, the feature effect $\Delta^\ell(Z_i)$ is given by a random function $\Delta^\ell: \Z \to \bbr$ of a test sample $Z_i$, depending on the training data $D_1$. 
To allow for different types of response variables $Y$ and estimands when specifying $\Delta^\ell$, we introduce a real-valued {\it loss function} $L$ on $\mart \times \Y$ that measures the discrepancy between the response predicted by the model $\M$ and the true response. 
A natural choice for $L$ is the loss function used to train $\M$, which ensures consistency across training and testing. 
Examples include the squared loss in regression used in (\ref{residuals}), the cross-entropy loss in classification (i.e., categorical $Y$), and the pinball loss in quantile regression.\footnote{With these specifications, $L$ measures predictive accuracy. Other choices of $L$ can quantify model performance in downstream screening or decision-making applications of $\M$ (e.g., revenue or cost), such as the financial loss resulting from a poor decision due to misclassification.}

\begin{definition}
The \emph{feature effect} $\Delta^\ell(Z_i)$ is defined as the incremental loss for test sample $i \in I_2$ induced by masking feature $\ell$. 
For $z = (x,y) \in \Z$, it is given by
\begin{align}\label{old-delta}
\Delta^\ell(z) = L(f(m^\ell(x)),y) - L(f(u^\ell(x)),y).
\end{align}
\end{definition}

The definition (\ref{old-delta}) can be paired with the masking functions in Examples~\ref{ex1} and~\ref{ex2}. 
If Eq.~(\ref{conditional}) is used, $\Delta^\ell(Z_i)$ defines a conditional feature effect that measures the incremental loss due to masking $\ell$ given the true values of the other features and thus captures the conditional contribution of a feature. 
A special case is our initial specification (\ref{residuals}), which arises from choosing the squared loss $L(b,y) = (b - y)^2$. 
If Eq.~(\ref{unconditional}) is used in (\ref{old-delta}), the feature effect $\Delta^\ell(Z_i)$ measures the incremental loss due to masking $\ell$ while keeping the other features masked and thus captures the stand-alone contribution of a feature. 
We numerically evaluate both formulations in Section~\ref{sec-numerical}.





The feature-effect framework extends naturally along several dimensions. 
First, while our discussion has focused on a single feature $\ell \in [d]$, the same approach applies to subsets of $[d]$, enabling joint testing of multiple features and the exploration of higher-order effects. 
Second, the framework accommodates panel data, where collections of units are observed over time; see Appendix~\ref{panel}. 
Third, it accommodates missing data; see Appendix~\ref{missing}.


\section{Hypothesis testing}\label{sec-test}
To construct a hypothesis test for the significance of feature $\ell$ under the model $f$, we examine the empirical distribution of the feature effects $\Delta^\ell(Z_i)$ across the test samples $i \in I_2$. 
A positive center of this distribution is treated as evidence of significance, since this implies that a majority of the test samples have positive feature effects.
Several classical tests could be used to assess such central tendencies, including $t$-, sign, Wilcoxon, and permutation tests. 
We propose a sign test for the median of the feature effects, which requires minimal assumptions on the data while retaining favorable asymptotic relative efficiency compared with a $t$-test.


\begin{assumption}\label{assumption-a}
Conditional on the training data $D_1$, the feature effects $\{\Delta^\ell(Z_i)\}_{i\in I_2}$ are conditionally independent draws from a distribution $\bbp_{M^\ell}$ with continuous distribution function $F^\ell$ and unique median $M^\ell$.
\end{assumption}

We note that $\bbp_{M^\ell}$ and $F^\ell$ in Assumption \ref{assumption-a} represent the conditional distribution and distribution function of the feature effects given the training data $D_1$, and that $M^\ell$ denotes the corresponding conditional median. A sufficient condition for uniqueness of $M^\ell$ is that $F^\ell$ be strictly increasing in a neighborhood of $M^\ell$.


Under Assumption~\ref{assumption-a}, the hypotheses for testing the significance of feature $\ell$ under $f$ can be formulated in terms of $M^\ell$. 
For real $M_0$, consider
\begin{align}\label{hypotheses}
H_0: M^\ell \le M_0 \quad \text{vs.} \quad H_1: M^\ell > M_0.
\end{align}
Take $M_0 = 0$. 
Under the null hypothesis, $M^\ell \le 0$, so masking feature $\ell$ does not systematically degrade predictive accuracy across the test set. 
Under the alternative, $M^\ell > 0$, so masking $\ell$ systematically increases prediction errors. 
While the choice $M_0 = 0$ is natural, Eq.~(\ref{hypotheses}) accommodates alternative values.

Let $B_{n,s}$ denote the Binomial distribution with parameters $n\in\bbn$ and $s\in(0,1)$. Write $\Pi_{n,s}$ for its distribution function, $\pi_{n,s}$ for its probability function, and $q_{\lambda}(B_{n,s})$ for its (lower) quantile at level $\lambda \in (0,1)$. Proposition \ref{ump} below provides an optimal sign test for (\ref{hypotheses}). 


\begin{proposition}\label{ump}Under Assumption \ref{assumption-a}, 
a Uniformly Most Powerful test of size $\alpha\in(0,1)$ rejects $H_0$ with probability $\phi_N(n^\ell_+(M_0),\alpha)$ where $n^\ell_+(w)=\#\{i\in I_2: \Delta^\ell(Z_i)> w\}$ is the number of feature effect samples in the test set exceeding $w\in \bbr$ and
\begin{align}\label{test}
 \phi_N(n,\alpha) =
    1\{n> T_{N,\alpha}\} + \gamma_N(n,\alpha)1\{n= T_{N,\alpha}\}        
\end{align}
where $N = |I_2|$ is the test sample size, $T_{N,\alpha}=q_{1-\alpha}(B_{N, 1/2})$ and 
\begin{align}\label{gamma}
\gamma_N(n,\alpha)=\frac{\Pi_{N,1/2}(n)-(1-\alpha)}{\pi_{N,1/2}(n)},\quad n=0,1,\ldots, N.
\end{align}
\end{proposition}









For brevity, we often suppress the dependence of the test statistic $n^\ell_+$ on $M_0$. 
The test rejects the null $H_0$ with probability one if $n^\ell_+$, the number of feature effects in the test set exceeding $M_0$, exceeds $T_{N,\alpha}$, the $(1-\alpha)$-quantile of $B_{N,1/2}$. 
If $n^\ell_+$ lies exactly on the boundary $T_{N,\alpha}$ of the rejection region, the test randomizes, rejecting $H_0$ with probability $\gamma_N(n^\ell_+,\alpha)$.
Relative to a non-randomized test of $H_0$ with test function $1\{\cdot > T_{N,\alpha}\}$, this randomization renders the procedure slightly more aggressive. 
Indeed, the randomized test achieves maximum power while guaranteeing \emph{exactness}, in the sense that the size satisfies 
\begin{align}\label{size}
\max_{M^\ell \le M_0} \bbe_{M^\ell}(\phi_N(n^\ell_+,\alpha)) = \alpha,
\end{align}
for any $M_0$ in (\ref{hypotheses}), with the maximum attained at $M^\ell = M_0$. 
Here and below, $\bbe_{M^\ell}$ denotes expectation under the feature effect sampling distribution $\bbp_{M^\ell}$ of Assumption~\ref{assumption-a}. 
Under $\bbp_{M^\ell}$, the test statistic $n^\ell_+$ has a Binomial distribution $B_{N,s}$, where $s = 1 - F^\ell(M_0) = 1/2$ when $M^\ell = M_0$ and $s < 1/2$ when $M^\ell < M_0$. 



The test statistic depends on the masking specification through the feature-effect construction (\ref{old-delta}). 
The specification in~(\ref{conditional}) yields a {conditional} significance test tied to the incremental contribution of a feature given the others, whereas Eq.~(\ref{unconditional}) yields a test based on its stand-alone contribution. 
Our numerical results in Section~\ref{sec-numerical} and Appendix \ref{robustness} suggest that the conditional test exhibits greater robustness in the presence of feature interactions and correlations.


In the context of the randomized test in Proposition \ref{ump} the notion of $p$-value is ambiguous. A conventional $p$-value could be defined as $\inf\{\alpha: \phi_N(n^\ell_+,\alpha)=1\}$, the smallest level of significance $\alpha$ at which $H_0$ is rejected with probability one, which is $1-\Pi_{N,1/2}(n^\ell_+-1)$. (This is also a $p$-value for a non-randomized test of $H_0$ with test function $1\{\cdot> T_{N,\alpha}\}$.) 
However, this formulation ignores the randomness of the test's decision rule (\ref{test}). We follow \ci{geyer-meeden} and consider a {\it randomized} $p$-value $p(n^\ell_+)$ with conditional distribution function given the test statistic specified by 
\begin{align}\label{unif}
\bbp_{M^\ell}(p(n^\ell_+)\le \alpha\,|\,n^\ell_+ )=\phi_N(n^\ell_+,\alpha),\quad \alpha\in(0,1)
\end{align}
for any $M^\ell \le M_0$. From (\ref{test}), this function represents a uniform distribution on an interval $(a_N(n^\ell_+),b_N(n^\ell_+))$ of length $\pi_{N,1/2}(n^\ell_+)$ where
\begin{align*}
a_N(n)&=1-\Pi_{N,1/2}(n),\\ 
b_N(n)&=1-\Pi_{N,1/2}(n-1), 
\end{align*}
the right endpoint being the conventional $p$-value. 
By iterated expectations and Eq.~(\ref{size}), 
\begin{equation}\label{p-ex}
\max_{M^\ell \le M_0} \bbp_{M^\ell}(p(n^\ell_+)\le \alpha)=\alpha. 
\end{equation}

\begin{figure}[t!]
\centering
\resizebox{0.72\textwidth}{!}{%
\begin{tikzpicture}[scale=0.45]
  \def\h{10}
  \def\w{24}
  \def\xz{\w*0.5/10}
  \def\xa{\w*3.75/10}
  \def\xb{\w*7.25/10}
  \def\yz{\h/10}
  \def\ya{\h*4/10}
  \def\yb{\h*8.25/10}
  \def\ytick{0.3}
  \def\ylabel{1}

  \draw[ultra thick,->,black] (0,\yz) -- (\w,\yz) node[right] {$\alpha$};
  \draw[ultra thick,->,black] (\xz,0) -- (\xz,\h)
    node[above] {$\phi_N(n^\ell_+,\alpha)$};
  \node[below left,black] at (\xz,\yz) {0};

  \draw[thick,-] (\xz-\ytick,\yb) -- (\xz+\ytick,\yb)
    node[left,xshift=-0.5cm] {$1$};

  \draw[ultra thick,-,blue!60!black] (\xz,\yz) -- (\xa,\yz);
  \draw[ultra thick,-,blue!60!black] (\xa,\yz) -- (\xb,\yb)
    node[midway, above, sloped, green!25!black, yshift=4pt]
    {\shortstack{\,\,\,\,\, Randomized\\\,\,\,\,\, Rejection}};
  \draw[ultra thick,-,blue!60!black] (\xb,\yb) -- (\w,\yb);

  \foreach \x in {\xa,\xb} {
    \draw[ultra thick,black] (\x,\yz+0.4) -- (\x,\yz-0.4);
  }

  \node[black, below] at (\xa,\yz - 0.5*\ylabel)
    {$1 - \Pi_{N,1/2}(n^\ell_+)$};
  \node[black, below] at (\xb,\yz - 0.5*\ylabel)
    {\;\; $1 - \Pi_{N,1/2}(n^\ell_+ - 1)$};

  \node[green!25!black] at ({(\xa + \xz)/2}, \yz + \ylabel) {Retention};
  \node[green!25!black] at ({(\xb + \w)/2},  \yb + \ylabel) {Rejection};

  \draw[dotted, thick, gray, dash pattern=on 8pt off 8pt] (\xa,\yz) -- (\xa,\yb);
  \draw[dotted, thick, gray, dash pattern=on 8pt off 8pt] (\xb,\yz) -- (\xb,\yb);

  \draw[decorate, decoration={brace, amplitude=6pt}, thick]
    (\xa,\yb + 0.5*\ylabel) -- node[above=8pt] {$\pi_{N,1/2}(n^\ell_+)$} (\xb,\yb + 0.5*\ylabel);

\end{tikzpicture}%
}
\caption{Distribution of randomized $p$-value and decision intervals.}
\label{pvalue-fig}
\end{figure}
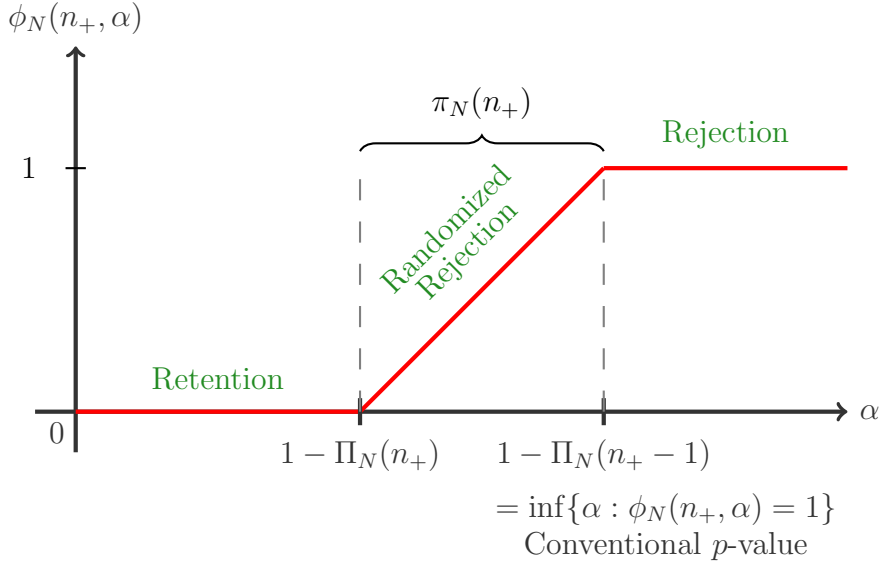

Equation (\ref{p-ex}) states that 
the randomized $p$-value $p(n^\ell_+)$ is standard uniform under the ``least favorable'' sampling distribution, for which $M^\ell = M_0$. 
In this sense, $p(n^\ell_+)$ is exact, mirroring the exactness of the test. If $p(n^\ell_+)$ is concentrated below a given level $\alpha$, or equivalently, if the conventional $p$-value $b_N(n^\ell_+)\le \alpha$, we reject the null. If $p(n^\ell_+)$ is concentrated above $\alpha$, or equivalently, if $\alpha<a_N(n^\ell_+)$, we retain the null. Else the evidence is equivocal; we thus randomize by drawing a sample $u$ from $U(a_N(n^\ell_+),b_N(n^\ell_+))$ and reject if $u\le \alpha$. See Figure \ref{pvalue-fig} for a visualization. In practice, it suffices to report the interval $(a_N(n^\ell_+),b_N(n^\ell_+))$ to fully characterize the $p$-value; there is often no need to force a decision.

The multiple testing problem can also be addressed in the randomized setting. A standard Bonferroni adjustment with test function $\phi_N(n,\alpha/d)$ and $p$-value being conditionally uniform on the interval $(d a_N(n^\ell_+)\wedge 1, d b_N(n^\ell_+)\wedge 1)$, where $d$ is the dimension of the feature vector, controls the family-wise error rate (the probability of having at least one false discovery), see \ci{kulinskaya2009fuzzy}. 
For a less conservative approach, one can also control the False Discovery Rate \cite{benjamini1995controlling}. Kulinskaya and Lewin~\ci{kulinskaya2009fuzzy} develop a formula and an algorithm for the adjusted rejection probability. Habiger~\ci{habiger2015multiple} proposes an alternative Monte Carlo approach. Cousido-Rocha et al.~\ci{cousido2022multiple} provide a comparison of methods. 

The power function $H_{N,\alpha}(s)$ of the test is given by  
\begin{align*}
H_{N,\alpha}(s)
=1-\Pi_{N,s}(T_{N,\alpha})+ \gamma_N(T_{N,\alpha},\alpha)\pi_{N,s}(T_{N,\alpha})
\end{align*}
where $s=1-F^\ell(M_0)\in(1/2,1)$ is the parameter of the test statistic's Binomial distribution $B_{N,s}$ under $H_1$.
We can invert this function to determine the test set size $N$ required to achieve a desired power level. Even when $s$ is only slightly larger than $1/2$, achieving power close to 1 requires $N$ on the order of $10^4$, which remains modest relative to typical data sizes in machine learning. A pilot sample can be used to estimate $s$ \ci{noether}. 



\section{Measuring feature importance}\label{sec-imp}
We provide tractable measures of feature importance (effect size) that complement our significance test and the associated feature $p$-values. 
A {\it local} measure of importance is given by the feature effect $\Delta^\ell(Z_i)$, which captures the impact of a feature on predictive performance at the level of an individual sample $i \in I_2$. 
Larger feature effects indicate greater influence at the sample level.

At the global level, a natural measure of feature importance is the median $M^\ell$ of the conditional feature-effect distribution $F^\ell$ from which the $\Delta^\ell(Z_i)$ are drawn under Assumption~\ref{assumption-a}. This median defines the feature significance hypotheses~(\ref{hypotheses}) being tested using the $\Delta^\ell(Z_i)$. When positive, $M^\ell$ measures the extent to which masking feature $\ell$ systematically degrades predictive accuracy across the test set. Larger positive values indicate a greater contribution of $\ell$ to predictive performance and hence greater importance.

The value of $M^\ell$ is unknown. A robust estimator is the empirical median of the feature effects across the test set,
\begin{equation}\label{emp-median}
\widehat M^\ell = \mathrm{Median}\{\Delta^\ell(Z_i): i \in I_2\}.
\end{equation}
Under Assumption~\ref{assumption-a}, this estimator is consistent and (approximately) equally likely to overestimate $M^\ell$ as to underestimate it. Moreover, it is unbiased when $F^\ell$ is symmetric about $M^\ell$; see Section~4.4 of \ci{lehmann}. The test, $p$-value, and importance score (\ref{emp-median}) together provide a unified framework for feature inference in any model.

The choice of masking specification in the construction of the feature effect $\Delta^\ell(Z_i)$ in (\ref{old-delta}) gives rise to alternative importance scores. 
The specifications in~(\ref{conditional}) and~(\ref{unconditional}) induce conditional and unconditional scores, respectively. 
The conditional score measures the incremental contribution of a feature given the others, whereas the unconditional score quantifies its stand-alone contribution. 
Our numerical results in Section~\ref{sec-numerical} and Appendix~\ref{robustness} indicate that the conditional score is more robust to feature interactions and correlations. 
This robustness represents an advantage over alternative importance measures such as SHAP~\cite{lundberg2017unified}, LOCO~\cite{rinaldo}, and permutation importance~\cite{breiman,fisher-rudin}, which are sensitive to feature correlation~\cite{verdinelli-wasserman,hooker}.

The resulting importance scores also depend on the choice of masking reference values. This dependence is natural and reflects the fact that predictive importance is inherently context-dependent. Accordingly, Eq.~(\ref{emp-median}) quantifies the predictive importance of a feature relative to a specified masking scheme and study cohort. This interpretation is consistent with the broader feature importance literature, where variable importance generally depends on the data distribution, prediction task, and feature perturbation mechanism.

The choice of loss function $L$ in the feature-effect construction (\ref{old-delta}) can change the magnitude of the scores but generally preserves the feature rankings, indicating robustness (see Section~\ref{sec-numerical}). 
The choice of $L$ should therefore be guided by the aspect of prediction quality of interest in the application. 
In practice, setting $L$ equal to the training loss function provides a natural default and ensures consistency across training and inference.


Since there is no ground truth for feature importance~\cite{hama-mase-owen}, multiple measures can be meaningful. 
Our score (\ref{emp-median}) offers an alternative to existing approaches. 
Relative to algorithmic measures such as SHAP and permutation-based scores, Eq.~(\ref{emp-median}) has the additional advantage of being tied to a hypothesis test that also delivers feature $p$-values. 
Moreover, as shown below, the testing framework yields confidence intervals for feature importance that algorithmic measures lack. 
Finally, as demonstrated in Section~\ref{sec-numerical}, our approach achieves an unprecedented degree of computational efficiency. 
Unlike existing methods, it can be used to measure feature importance even for the largest, feature-rich models.

\subsection*{Confidence intervals}
To complement the point estimator (\ref{emp-median}), we construct confidence intervals (CIs) for $M^\ell$ by inverting the test (\ref{test}). Let $\Delta^\ell_{(1)}\le \Delta^\ell_{(2)}\le \cdots\le \Delta^\ell_{(N)}$ denote the ordered sequence of the $\Delta^\ell(Z_i)$ across $I_2$. Define $\Delta^\ell_{(0)}=-\infty$ and $\Delta^\ell_{(N+1)}=\infty$. 

\begin{proposition}\label{ci-prop}
Let $U$ be a standard uniform variable that is independent of all other random elements including the data. Define a {randomized} CI for $M^\ell$ by  
\begin{equation}\label{ci}
    C^\ell_N(\alpha;U) 
			=\begin{cases}
			[\Delta^\ell_{(N - T_{N,\alpha})}, \infty),  & 1 - \gamma_N(T_{N,\alpha}, \alpha) \geq U\\
			[\Delta^\ell_{(N - T_{N,\alpha}+1)}, \infty), & \text{else} 		
			\end{cases}
\end{equation}
where $T_{N,\alpha}$ and $\gamma_N(n,\alpha)$ are as defined in Proposition \ref{ump}.
Under Assumption \ref{assumption-a}, the CI (\ref{ci}) has coverage probability $1-\alpha$ for $\alpha\in(0,1)$.
\end{proposition}

In analogy to the randomized $p$-value, the randomized CI (\ref{ci}) inherits the exactness of the test’s decision rule (\ref{test}). If $M_0<\Delta^\ell_{(N - T_{N,\alpha})}$ we reject the null. If $\Delta^\ell_{(N - T_{N,\alpha}+1)}\le M_0$ we retain the null. Else we reject with probability $\gamma_N(T_{N,\alpha}, \alpha)$. 
In practice, it suffices to report the interval endpoints along with the rejection probability. 


Standard arguments lead to alternative two-sided CIs, which are not, however, dual to the test (\ref{test}). Observe that the probability of 
$M^\ell\in[\Delta^\ell_{(k)},\Delta^\ell_{(k+1)}]$ equals $\pi_{N,1/2}(k)$.
Using these probabilities, we can combine intervals of this form 
so as to attain a desired confidence coefficient. Choosing the intervals symmetrically, we obtain intervals
\begin{equation}\label{naive-cis}
	c^\ell_{m,N}=[\Delta^\ell_{(1+m)}, \Delta^\ell_{(N-m)}]
\end{equation}	
with coverage probability $1-2 \Pi_{N,1/2}(m)$,
where we select $m\in\bbn$ so that $\Pi_{N,1/2}(m)\le \alpha/2$ for given $\alpha\in(0,1)$, 
implying a conservative approach. 
Zieli\'nski and Zieli\'nski~\ci{zielinski} extend this argument using randomization, resulting in CIs with exact $(1-\alpha)$ coverage and minimum expected length. Alternative asymptotic CIs can be obtained by bootstrapping (e.g. \ci{efron} and \ci{falk-kaufmann}), empirical likelihood \cite{chen-hall}, sectioning \cite{asmussen-glynn} and other methods.




\section{Synthetic tasks}\label{sec-numerical}
We illustrate the robust performance of our test and importance measure on a synthetic ``known-truth'' regression task implemented with neural networks. Appendix \ref{classification-sect} provides results for a related classification task.

We consider a feature vector $X\in\bbr^d$ of dimension $d=19$ and a response 
	$Y = \mu(X) + \epsilon$,
where the error $\epsilon \sim N(0,1)$ is independent of $X$, and for $x=(x^1,\ldots,x^{d})$, the conditional mean function $\mu$ is given by
\begin{align}\label{mu}
	\mu(x)= 3 &+ 4x^1 + x^1x^2 + 3(x^3)^2 + 2x^4x^5 + 6x^6 + 2\sin(x^7) \nonumber\\ 
	&+  \exp(x^8) + 5x^9 + 3x^{10} + 4x^{11} + 5x^{12}.
\end{align}

The features $X^1$ and $X^6$ are jointly normal with mean zero, unit variances, and covariance $0.85$. 
The features $X^2,\ldots,X^5$ and $X^7$ are independent $N(0,1)$ variables, and $X^8$ is uniformly distributed on $(-1,1)$. 
The binary feature $X^9 = 1\{X^2 + W < 0\}$ is generated from an independent $W \sim N(0,1)$. 
Feature $X^{10}$ follows a Poisson distribution with mean~3, while $X^{11}$ and $X^{12}$ are independent $t$-distributed variables with five degrees of freedom. 
The null variables $X^{13}$ and $X^{14}$ are independent $N(0,1)$, $(X^{15}, X^{16})$ are jointly normal with mean zero, unit variances, and covariance~$0.85$, and $X^{17}, X^{18}$, and $X^{19}$ are independent $t$-distributed variables with five degrees of freedom.

We generate a training set $I_1$ of $10^6$ independent samples $Z_i=(X_i,Y_i)$ and a test set $I_2$ of $5\cdot10^5$ independent samples. 
We train a single-layer neural network model $\M$ estimating Eq.~(\ref{mu}), defined as
\begin{align}\label{nn}
	\M(x)=b_0 + \sum_{u=1}^U b_u \psi(a_{0,u} + a_u^\top x),
\end{align}
where $U$ is the number of hidden units, $\psi(y)=1/(1+e^{-y})$ is the activation function, and $b_0, b_u, a_{0,u} \in \mathbb{R}$ and $a_u \in \mathbb{R}^d$ are parameters. 
Training uses a standard $L_2$ loss criterion with $L_2$ regularization on the weights $b_u$, the Adam optimization method, a batch size of 32, and 500 epochs with early stopping. 
We iterate across five random initializations. 
A validation set (25\% of $I_1$, excluded from training) and a Bayesian search are used to optimize the regularization weight, learning rate, and number of units~$U$. 
The optimal regularization weight is $7\cdot10^{-4}$, the optimal learning rate is $5\cdot10^{-4}$, and the optimal number of units is $U=300$. 
The resulting model achieves an $R^2$ exceeding 99\% on both $I_1$ and $I_2$.

Consistent with the $L_2$ training criterion, we use the squared loss $L(m,y)=(m-y)^2$ in (\ref{old-delta}). 
We adopt the masking and unmasking functions described in Example~\ref{ex1}, yielding a conditional feature effect (\ref{old-delta}). 
For continuous features, the masking reference value is set to the feature mean over the training set $I_1$ (see Appendix \ref{app-regress} for an alternative choice). 
For discrete features ($\ell=9,10$), we use the adjusted mode over $I_1$, as specified in Section~\ref{general}. 
Importantly, the validity of the test does not depend on these particular specifications. 
Below, we assess robustness to alternative loss functions, masking schemes, and reference values.

\begin{table}[h]
\centering
\resizebox{0.8\columnwidth}{!}{
\renewcommand{\arraystretch}{1} 
\setlength{\tabcolsep}{4.5pt}      
\begin{tabular}{@{}lcccccccccccc@{}}
\toprule
\multirow{3}{*}{$\ell$} 
& \multicolumn{2}{c}{\multirow{2}{*}{\textbf{AICO}}}
& \multicolumn{2}{c}{\multirow{2}{*}{\textbf{CPT}}} 
& \multicolumn{2}{c}{\multirow{2}{*}{\textbf{HRT}}} 
& \multicolumn{2}{c}{\multirow{2}{*}{\textbf{LOCO}}} 
& \multicolumn{2}{c}{\textbf{Deep}} 
& \multicolumn{2}{c}{\textbf{Gaussian}} \\
& & & & & & & & & \multicolumn{2}{c}{\textbf{Knockoff}} & \multicolumn{2}{c}{\textbf{Knockoff}} \\
\cmidrule(lr){2-3} \cmidrule(lr){4-5} \cmidrule(lr){6-7} \cmidrule(lr){8-9} \cmidrule(lr){10-11} \cmidrule(lr){12-13}
& 5\% & 1\% & 5\% & 1\% & 5\% & 1\% & 5\% & 1\% & 5\% & 1\% & 5\% & 1\% \\
\midrule
\textbf{1}  & 10 & 10 & 10 & 10 & 10 & 10 & 10 & 10 & 10 & 10 & 10 & 10 \\
\textbf{2}  & 10 & 10 & 10 & 10 & 10 & 10 & 10 & 10 & 10 & 10 & 10 & 10 \\
\textbf{3}  & 10 & 10 & 10 & 10 & 10 & 10 & 10 & 10 & 10 & 10 & 10 & 10 \\
\textbf{4}  & 10 & 10 & 10 & 10 & 10 & 10 & 10 & 10 & 8  & 8  & 10 & 10 \\
\textbf{5}  & 10 & 10 & 10 & 10 & 10 & 10 & 10 & 10 & 8  & 8  & 10 & 10 \\
\textbf{6}  & 10 & 10 & 10 & 10 & 10 & 10 & 10 & 10 & 10 & 10 & 10 & 10 \\
\textbf{7}  & 10 & 10 & 10 & 10 & 10 & 10 & 10 & 10 & 10 & 10 & 10 & 10 \\
\textbf{8}  & 10 & 10 & 10 & 10 & 10 & 10 & 10 & 10 & 9  & 9  & 10 & 10 \\
\textbf{9}  & 10 & 10 & 10 & 10 & 10 & 10 & 10 & 10 & 10 & 10 & 10 & 10 \\
\textbf{10} & 10 & 10 & 10 & 10 & 10 & 10 & 10 & 10 & 10 & 10 & 10 & 10 \\
\textbf{11} & 10 & 10 & 10 & 10 & 10 & 10 & 10 & 10 & 10 & 10 & 10 & 10 \\
\textbf{12} & 10 & 10 & 10 & 10 & 10 & 10 & 10 & 10 & 10 & 10 & 10 & 10 \\
\midrule
\textbf{13} & 0  & 0 & 0 & 0 & 0  & 0  & 0  & 0  & 3  & 3  & 4  & 4  \\
\textbf{14} & 0  & 0 & 1 & 0 & 1  & 0  & 1  & 1  & 2  & 2  & 3  & 3  \\
\textbf{15} & 0  & 0 & 2 & 1 & 1  & 0  & 0  & 0  & 0  & 0  & 4  & 4  \\
\textbf{16} & 0  & 0 & 1 & 0 & 1  & 1  & 1  & 0  & 1  & 1  & 3  & 3  \\
\textbf{17} & 0  & 0 & 0 & 0 & 0  & 0  & 0  & 0  & 2  & 2  & 3  & 3  \\
\textbf{18} & 0  & 0 & 0 & 0 & 1  & 0  & 0  & 0  & 0  & 0  & 2  & 2  \\
\textbf{19} & 0  & 0 & 1 & 0 & 0  & 0  & 0  & 0  & 0  & 0  & 2  & 2  \\
\bottomrule
\end{tabular}}
\caption{Results for 10 independent training/testing experiments at two standard significance levels ($\alpha = 5\%$ and $\alpha = 1\%$). The table compares AICO, CPT, HRT, LOCO, Deep Knockoff, and Gaussian Knockoff procedures. For each feature, we report the total number of rejections across 10 trials. The test set size is $N = 5 \cdot 10^5$.}
\label{table-fo-regression}
\end{table}


We run ten independent training/testing experiments to assess the performance of the test, using the default value $M_0=0$ for our hypotheses (\ref{hypotheses}).
The second and third columns of Table~\ref{table-fo-regression} report results for significance levels $\alpha=0.05$ and $0.01$.\footnote{\label{fn-ties}Limited numerical precision can lead to ties, i.e., samples $\Delta^\ell(Z_i)$ whose true values are distinct being rounded to the same value. If the value of a tie is strictly positive, we count each sample in the test statistic $n^\ell_+$. Another numerical artifact arises when samples $\Delta^\ell(Z_i)$ with true, nonzero values are rounded to zero. We treat these as nonpositive, excluding them from $n^\ell_+$---a conservative choice. Alternative, potentially less conservative treatments include dropping ties (adjusting $N$) or randomly breaking them. See also \ci{wittkowski} for another approach.}
Across all experiments, the fitted model and the AICO test identify as significant each feature that enters the data-generating process (DGP) through (\ref{mu}), while no feature that is null in the DGP is identified as significant.
Feature correlation (e.g., $\ell=1,6$), discrete feature distributions (e.g., $\ell=9,10$), and interaction effects (e.g., $\ell=2,4,5$) do not impair performance.


We conduct an extensive set of additional robustness experiments. Results from 2{,}000 independent trials, reported in Appendix~\ref{app-regress}, confirm the strong statistical performance documented in Table~\ref{table-fo-regression}. 
Performance remains stable under alternative masking reference values—for example, using 
the mean instead of the adjusted mode for discrete features. 
Similarly, results are robust to the choice of loss function $L$, including the use of absolute rather than squared loss. 
Performance is, however, moderately sensitive to the masking specification: replacing the conditional formulation~(\ref{conditional}) with the unconditional specification~(\ref{unconditional}) makes features $\ell=2,4,5$ less likely to be identified as significant and increases the number of discoveries among features that are null in the DGP.
These features affect the response primarily through interaction terms, making their contribution harder to detect under the unconditional formulation. 
This behavior underscores the importance of conditioning via (\ref{conditional}) when features contribute through interactions and illustrates how the masking choice influences testing in nonlinear settings.

Finally, we assess robustness with respect to the distributions of the features.
The results remain stable when replacing the Gaussian distributions of features $\ell=1,\ldots,7$ with log-normal, exponential, and Beta distributions.
The results are also robust to varying the correlation between the jointly Gaussian features $\ell=1,6$ between 0.25 and 1.
To further analyze the behavior under feature correlation, we consider a specification in which feature 1 and the DGP-null feature 13 are jointly Gaussian with correlation varying between 0.25 and 1 (while $\ell=1,6$ are independent).
Performance remains robust unless the correlation is close to 1, in which case the model can no longer reliably distinguish between the two features, and feature 13, although null in the DGP, is increasingly identified as significant.
In the degenerate case of perfect correlation, the two features are identical and both are identified as significant.


We also evaluate the procedure in a cross-fitting framework with five (rather than one) data splits, allowing each observation to contribute to both training and testing while preserving valid inference. Appendix~\ref{cross-fitting} provides details; Table~\ref{table-cv-aico} in Appendix~\ref{cross-fitting} confirms that the results remain stable under this more efficient data-splitting strategy.

 

We contrast our test with several recently developed model-agnostic alternatives, including the Leave-Out-Covariates Test (LOCO) of \ci{Lei2018predictive}, the Conditional Permutation Test (CPT) of \ci{berrett}, and the Holdout Randomization Test (HRT) of \ci{tansey2018holdout}. 
LOCO evaluates the change in predictive performance induced by removing a feature from the model. 
Testing all features requires retraining the neural network $d=19$ times, once for each feature subset $X^{-\ell}=(X^1,\ldots,X^{\ell-1},X^{\ell+1},\ldots,X^d)$, $\ell\in[d]$.\footnote{In our numerical experiments, each of the $d=19$ submodels has the same architecture as the full model $f$ and is trained using the same procedure.} 
The CPT and HRT procedures build on the Conditional Randomization Test of \ci{candes2018panning} and evaluate the conditional independence hypothesis $Y \ind X^\ell \,|\, X^{-\ell}$. 
Both methods require sampling from the (unknown) conditional law of $X^\ell$ given $X^{-\ell}$. 
Following \ci{tansey2018holdout}, we fit $d=19$ mixture-density neural networks to approximate these conditional distributions. 


Table~\ref{table-fo-regression} reports results for these alternative procedures across ten independent training/testing experiments.
CPT, HRT, and LOCO identify as significant the same features as AICO in all experiments, but more frequently identify as significant features that are null in the DGP.
Table~\ref{table-fo-regression-1000} in Appendix~\ref{robustness-comp-large} reports results for 1{,}000 independent trials, confirming that, overall, AICO produces the fewest such discoveries among the testing procedures considered. 

For additional perspective, we implement two prominent model-agnostic variable-selection procedures: the Gaussian Model-X Knockoff method of \ci{candes2018panning} and the Deep Model-X Knockoff method of \ci{deepknockoffs}. 
The Gaussian variant assumes a multivariate normal distribution for the features, whereas the deep variant employs a neural network to learn a generative model of the feature distribution. 
As shown in Table~\ref{table-fo-regression}, the Gaussian method consistently selects features~1 through~12 in every experiment, though it also frequently selects features that are null in the DGP, whereas the Deep method selects fewer such features but shows modest difficulty identifying features~4 and~5 that enter the response function~(\ref{mu}) through interaction terms.

We examine the performance of the AICO procedure as a function of the test-set size $N = |I_2|$.
Table~\ref{table-fo-regression-testset} shows that all features entering the DGP are identified as significant for sample sizes as small as $N=500$.
For smaller $N$, identification rates decline moderately, while features that are null in the DGP are identified as significant only slightly more often.
These results indicate reliable performance even in small-sample settings. 

\begin{table}[t!]
\centering
\resizebox{0.8\columnwidth}{!}{
\renewcommand{\arraystretch}{1.1} 
\setlength{\tabcolsep}{4.5pt}      
\begin{tabular}{@{}lcccccccccccc@{}} 
\toprule
\multicolumn{13}{c}{\textbf{AICO}} \\ 
\toprule
\multirow{2}{*}{$\ell$} 
& \multicolumn{2}{c}{$N = 10^5$} 
& \multicolumn{2}{c}{$N = 10^4$} 
& \multicolumn{2}{c}{$N = 10^3$} 
& \multicolumn{2}{c}{$N = 500$} 
& \multicolumn{2}{c}{$N = 250$} 
& \multicolumn{2}{c}{$N = 100$} \\
\cmidrule(lr){2-3} \cmidrule(lr){4-5} \cmidrule(lr){6-7} \cmidrule(lr){8-9} \cmidrule(lr){10-11} \cmidrule(lr){12-13}
& 5\% & 1\% & 5\% & 1\% & 5\% & 1\% & 5\% & 1\% & 5\% & 1\% & 5\% & 1\% \\
\midrule
\textbf{1}  & 10 & 10 & 10 & 10 & 10 & 10 & 10 & 10 & 10 & 10 & 10 & 10 \\
\textbf{2}  & 10 & 10 & 10 & 10 & 10 & 10 & 10 & 10 & 10 & 9  & 8  & 5  \\
\textbf{3}  & 10 & 10 & 10 & 10 & 10 & 10 & 10 & 10 & 10 & 10 & 10 & 10 \\
\textbf{4}  & 10 & 10 & 10 & 10 & 10 & 10 & 10 & 10 & 10 & 10 & 9  & 9  \\
\textbf{5}  & 10 & 10 & 10 & 10 & 10 & 10 & 10 & 10 & 10 & 10 & 10 & 9  \\
\textbf{6}  & 10 & 10 & 10 & 10 & 10 & 10 & 10 & 10 & 10 & 10 & 10 & 10 \\
\textbf{7}  & 10 & 10 & 10 & 10 & 10 & 10 & 10 & 10 & 10 & 10 & 10 & 10 \\
\textbf{8}  & 10 & 10 & 10 & 10 & 10 & 10 & 10 & 10 & 9  & 8  & 8  & 7  \\
\textbf{9}  & 10 & 10 & 10 & 10 & 10 & 10 & 10 & 10 & 10 & 10 & 10 & 10 \\
\textbf{10} & 10 & 10 & 10 & 10 & 10 & 10 & 10 & 10 & 10 & 10 & 10 & 10 \\
\textbf{11} & 10 & 10 & 10 & 10 & 10 & 10 & 10 & 10 & 10 & 10 & 10 & 10 \\
\textbf{12} & 10 & 10 & 10 & 10 & 10 & 10 & 10 & 10 & 10 & 10 & 10 & 10 \\
\midrule
\textbf{13} & 0  & 0  & 0  & 0  & 0  & 0  & 0  & 0  & 2  & 1  & 0  & 0  \\
\textbf{14} & 0  & 0  & 0  & 0  & 0  & 0  & 1  & 0  & 0  & 0  & 1  & 0  \\
\textbf{15} & 0  & 0  & 1  & 0  & 1  & 1  & 0  & 0  & 1  & 0  & 1  & 0  \\
\textbf{16} & 0  & 0  & 0  & 0  & 0  & 0  & 1  & 1  & 0  & 0  & 1  & 0  \\
\textbf{17} & 1  & 0  & 0  & 0  & 1  & 0  & 0  & 0  & 0  & 0  & 0  & 0  \\
\textbf{18} & 0  & 0  & 1  & 0  & 0  & 0  & 0  & 0  & 0  & 0  & 0  & 0  \\
\textbf{19} & 0  & 0  & 1  & 0  & 0  & 0  & 0  & 0  & 0  & 0  & 2  & 2  \\
\bottomrule
\end{tabular}}
\caption{AICO test results for 10 independent training/testing experiments, evaluated at two significance levels ($\alpha = 5\%$ and $1\%$) for several test sample sizes $N$. For each feature, we report the number of times the null was rejected out of 10 trials.}
\label{table-fo-regression-testset}
\end{table}

The experiments above reference the DGP and therefore do not directly assess the finite-sample validity result of Proposition~\ref{ump}. A separate numerical exercise in Appendix~\ref{app-validity}, conducted on fitted models satisfying the AICO null $M^\ell\le 0$, yields rejection frequencies consistent with the exact finite-sample size property (\ref{size}).

We now turn to measuring feature importance. 
Table~\ref{table-cis} presents a feature-importance ranking based on the score~(\ref{emp-median}) for one randomly selected training/testing experiment out of ten. 
It also reports the endpoints of the randomized confidence interval~(\ref{ci}) at the $99\%$ level, along with the standard two-sided interval~(\ref{naive-cis}) and their corresponding coverage probabilities. 
In addition, Table~\ref{table-cis} provides the distribution of the randomized $p$-values.\footnote{Ties among samples $\Delta^\ell(Z_i)$ due to limited numerical precision are retained in $\{\Delta^\ell(Z_i): i\in I_2\}$ when computing confidence intervals and coverage probabilities. 
This treatment is consistent with the handling of ties in the testing procedure (footnote~\ref{fn-ties}).}

\begin{table}[t!]
\centering
\resizebox{\columnwidth}{!}{
\centering
\renewcommand{\arraystretch}{1.1} 
\setlength{\tabcolsep}{2pt}       
\small
\resizebox{\textwidth}{!}{
\begin{tabular}{@{}lcccccccccccc@{}}
\toprule
\multicolumn{12}{c}{\textbf{AICO, Level $\alpha=1\%$}} \\
\toprule
\multicolumn{4}{c}{} & \multicolumn{2}{c}{$p$-value Distribution} & \multicolumn{3}{c}{Randomized One-Sided CI} & \multicolumn{3}{c}{Non-Randomized Two-Sided CI} \\
\cmidrule(lr){5-6} \cmidrule(lr){7-9} \cmidrule(lr){10-12}
Rank & Feature & Median & P(reject) & Lower & Upper & 98.8\% & 1.2\% & Coverage & Lower & Upper & Coverage \\
\midrule
1  & $X_9$ & 24.993 & 1.000 & 0.000 & 0.000 & 24.951 & 24.951 & 0.99 & 24.947 & 25.040 & 0.99007 \\
2  & $X_6$ & 15.788 & 1.000 & 0.000 & 0.000 & 15.661 & 15.661 & 0.99 & 15.647 & 15.938 & 0.99007 \\
3  & $X_{10}$ & 14.572 & 1.000 & 0.000 & 0.000 & 14.511 & 14.511 & 0.99 & 14.506 & 14.640 & 0.99007 \\
4  & $X_{12}$ & 12.898 & 1.000 & 0.000 & 0.000 & 12.789 & 12.789 & 0.99 & 12.775 & 13.022 & 0.99007 \\
5  & $X_{11}$ & 8.137  & 1.000 & 0.000 & 0.000 & 8.063  & 8.063  & 0.99 & 8.054  & 8.218  & 0.99007 \\
6  & $X_1$ & 6.210  & 1.000 & 0.000 & 0.000 & 6.155  & 6.155  & 0.99 & 6.148  & 6.273  & 0.99007 \\
7  & $X_3$ & 1.697  & 1.000 & 0.000 & 0.000 & 1.669  & 1.669  & 0.99 & 1.666  & 1.730  & 0.99007 \\
8  & $X_7$ & 0.953  & 1.000 & 0.000 & 0.000 & 0.941  & 0.941  & 0.99 & 0.940  & 0.964  & 0.99007 \\
9  & $X_4$ & 0.416  & 1.000 & 0.000 & 0.000 & 0.409  & 0.409  & 0.99 & 0.409  & 0.424  & 0.99007 \\
10 & $X_5$ & 0.416  & 1.000 & 0.000 & 0.000 & 0.409  & 0.409  & 0.99 & 0.408  & 0.424  & 0.99007 \\
11 & $X_8$ & 0.143  & 1.000 & 0.000 & 0.000 & 0.140  & 0.140  & 0.99 & 0.139  & 0.146  & 0.99007 \\
12 & $X_2$ & 0.084  & 1.000 & 0.000 & 0.000 & 0.082  & 0.082  & 0.99 & 0.082  & 0.086  & 0.99007 \\
\midrule
- & $X_{13}$ & -1.154e-07 & 0.000 & 0.880 & 0.881 & -2.442e-06 & -2.437e-06 & 0.99 & -2.708e-06 & 1.478e-06 & 0.99007 \\
- & $X_{14}$ & -2.054e-07 & 0.000 & 0.928 & 0.928 & -2.176e-06 & -2.175e-06 & 0.99 & -2.389e-06 & 9.014e-07 & 0.99007 \\
- & $X_{15}$ & -3.060e-07 & 0.000 & 0.845 & 0.846 & -3.339e-06 & -3.338e-06 & 0.99 & -3.639e-06 & 2.127e-06 & 0.99007 \\
- & $X_{16}$ & -5.512e-07 & 0.000 & 0.937 & 0.937 & -2.966e-06 & -2.965e-06 & 0.99 & -3.275e-06 & 9.626e-07 & 0.99007 \\
- & $X_{17}$ & -1.340e-06 & 0.000 & 0.996 & 0.997 & -3.773e-06 & -3.773e-06 & 0.99 & -4.022e-06 & 0.000      & 0.99150 \\
- & $X_{18}$ & 0.000      & 0.000 & 0.860 & 0.861 & -1.426e-06 & -1.423e-06 & 0.99 & -1.634e-06 & 1.114e-06 & 0.99007 \\
- & $X_{19}$ & 0.000      & 0.000 & 0.815 & 0.816 & -1.517e-06 & -1.516e-06 & 0.99 & -1.716e-06 & 1.333e-06 & 0.99007 \\
\bottomrule
\end{tabular}}
}
\caption{AICO feature importance ranking according to the empirical median (\ref{emp-median}) for one of the independent training/testing experiments that was randomly chosen. The table also reports the endpoints of the randomized CI (\ref{ci}) for the median at a $99\%$ level, the standard (non-randomized) two-sided CI (\ref{naive-cis}) along with coverage probabilities, and the distribution of the randomized $p$-value. Percentage heads: probabilities of the two possible lower endpoints of the randomized CI.}\label{table-cis}
\end{table}

The $p$-value distributions provide a clear separation between features identified as significant and those for which the null is retained.
For features $1-12$, the distributions are concentrated near zero, far below the level $\alpha = 1\%$.
This provides strong evidence against the null hypothesis for these features, with rejection probability equal to one.
In contrast, the $p$-value distributions for features $13,\ldots,19$ are concentrated well above $\alpha$, providing clear support for retaining the null.
No equivocal cases are observed, eliminating the need for randomization.

Feature $9$ emerges as the most important variable, followed by features $6$, $10$, and~$12$.
The remaining features are relatively less influential compared with this group.
Features $4$ and~$5$ exhibit (almost) identical medians and confidence intervals, consistent with their symmetric roles.

As shown in Table~\ref{rank-table-reg-narrow}, the AICO scores~(\ref{emp-median}) and resulting rankings are stable across independent training/testing experiments, demonstrating robustness to sampling variability.
Table~\ref{rank-table-reg-narrow} also reports the corresponding rankings for CPT, HRT, and LOCO importance measures~\cite{berrett,tansey2018holdout,verdinelli-wasserman}; Deep and Gaussian Knockoff importance measures~\cite{candes2018panning,deepknockoffs}; permutation importance~\cite{breiman,fisher-rudin}; and Deep and Kernel SHAP importance measures~\cite{lundberg2017unified}.
Different measures may induce different feature rankings, which is expected since there is generally no unique ground truth for feature importance~\cite{hama-mase-owen}.
Across all measures, the symmetric roles of features~4 and~5 in~(\ref{mu}) lead to minor variation in their relative rankings across experiments.




\begin{table}[t!]
\centering
\resizebox{\columnwidth}{!}{
\renewcommand{\arraystretch}{1.10}
\setlength{\tabcolsep}{2pt}
\scriptsize
\begin{tabular}{@{}lccccccccc@{}}
\toprule
\multirow{2}{*}{$\ell$}
& \multicolumn{1}{c}{\multirow{2}{*}{\textbf{AICO}}}
& \multicolumn{1}{c}{\multirow{2}{*}{\textbf{CPT}}}
& \multicolumn{1}{c}{\multirow{2}{*}{\textbf{HRT}}}
& \multicolumn{1}{c}{\multirow{2}{*}{\textbf{LOCO}}}
& \multicolumn{1}{c}{\textbf{Deep}}
& \multicolumn{1}{c}{\textbf{Gaussian}}
& \multicolumn{1}{c}{\textbf{Permutation}}
& \multicolumn{1}{c}{\textbf{Deep}}
& \multicolumn{1}{c}{\textbf{Kernel}} \\
& & & & & \multicolumn{1}{c}{\textbf{Knockoff}} & \multicolumn{1}{c}{\textbf{Knockoff}} & \multicolumn{1}{c}{\textbf{Importance}} & \multicolumn{1}{c}{\textbf{SHAP}} & \multicolumn{1}{c}{\textbf{SHAP}} \\
\midrule
\textbf{1} & 6 & 7 & 7 & 7 & 6 & 6 & 6 & 5 & 5 \\
\textbf{2} & 12 & 11 & 11 & 12 & 9 & 9\,{\fontsize{5.8pt}{6.8pt}\selectfont (9--10)} & 11 & 12 & 12 \\
\textbf{3} & 7 & 4 & 4 & 4 & 5 & 5 & 5 & 6 & 6 \\
\textbf{4} & 9\,{\fontsize{5.8pt}{6.8pt}\selectfont (9--10)} & 8\,{\fontsize{5.8pt}{6.8pt}\selectfont (8--9)} & 8\,{\fontsize{5.8pt}{6.8pt}\selectfont (8--9)} & 9\,{\fontsize{5.8pt}{6.8pt}\selectfont (9--10)} & 12\,{\fontsize{5.8pt}{6.8pt}\selectfont (11--12)} & 12\,{\fontsize{5.8pt}{6.8pt}\selectfont (11--12)} & 8\,{\fontsize{5.8pt}{6.8pt}\selectfont (8--9)} & 9\,{\fontsize{5.8pt}{6.8pt}\selectfont (9--10)} & 9\,{\fontsize{5.8pt}{6.8pt}\selectfont (9--10)} \\
\textbf{5} & 10\,{\fontsize{5.8pt}{6.8pt}\selectfont (9--10)} & 9\,{\fontsize{5.8pt}{6.8pt}\selectfont (8--9)} & 9\,{\fontsize{5.8pt}{6.8pt}\selectfont (8--9)} & 10\,{\fontsize{5.8pt}{6.8pt}\selectfont (9--10)} & 11\,{\fontsize{5.8pt}{6.8pt}\selectfont (10--12)} & 11\,{\fontsize{5.8pt}{6.8pt}\selectfont (11--12)} & 9\,{\fontsize{5.8pt}{6.8pt}\selectfont (8--9)} & 10\,{\fontsize{5.8pt}{6.8pt}\selectfont (9--10)} & 10\,{\fontsize{5.8pt}{6.8pt}\selectfont (9--10)} \\
\textbf{6} & 2 & 5 & 5 & 5 & 1 & 1 & 2 & 1 & 1 \\
\textbf{7} & 8 & 10 & 10 & 8 & 8 & 8 & 10 & 8 & 8 \\
\textbf{8} & 11 & 12 & 12 & 11 & 10\,{\fontsize{5.8pt}{6.8pt}\selectfont (10--11)} & 10\,{\fontsize{5.8pt}{6.8pt}\selectfont (9--10)} & 12 & 11 & 11 \\
\textbf{9} & 1 & 6 & 6 & 6 & 7 & 7 & 7 & 7 & 7 \\
\textbf{10} & 3 & 2 & 3\,{\fontsize{5.8pt}{6.8pt}\selectfont (2--3)} & 2 & 4 & 4 & 3 & 3 & 3 \\
\textbf{11} & 5 & 3 & 2\,{\fontsize{5.8pt}{6.8pt}\selectfont (2--3)} & 3 & 3 & 3 & 4 & 4 & 4 \\
\textbf{12} & 4 & 1 & 1 & 1 & 2 & 2 & 1 & 2 & 2 \\
\bottomrule
\end{tabular}}
\caption{Ranking stability across ten independent training/testing experiments. For each measure and trial, the 12 features entering the response function are ranked by importance among themselves (1 = most important). Each cell reports the median rank across the 10 trials and, when the rank varies across trials, the corresponding range (Min--Max) in parentheses. Ties between features within a method are broken using the mean rank so that all 12 reported ranks are distinct.}
\label{rank-table-reg-narrow}
\end{table}

We run additional experiments to evaluate the robustness of the AICO feature importance scores to the choice of loss function $L$ and masking specification. 
Replacing squared loss with absolute loss changes the magnitude of (\ref{emp-median}) but preserves feature rankings. 
This is expected because both loss functions quantify prediction error, albeit on different scales. 
Applying the training mean instead of the adjusted mode as the masking reference value for the discrete features ($\ell=9,10$) moderately decreases the score (\ref{emp-median}) and the relative importance of these two features. 
This is also expected because the observed feature value tends to be closer to the mean than to the adjusted mode, reducing the magnitude of the feature effect. 
While using the mean as a reference value for discrete and categorical variables is generally not recommended---due to potential out-of-support values and the reference value being too close to some feature values---this experiment highlights the importance of carefully choosing these reference values.
 

Switching from the conditional formulation~(\ref{conditional}) to the unconditional specification~(\ref{unconditional}) affects the magnitude of the scores, while the corresponding ranking remains largely stable.
As discussed above in the testing context, using Eq.~(\ref{unconditional}) makes interaction features $\ell=2,4,5$ less likely to be identified as significant, and this behavior is also reflected in the scores.
Specifically, across all trials features $\ell=4,5$ have negative scores and are never identified as significant, while the score of feature $\ell=2$ is highly volatile.
These findings further highlight the importance of the masking specification for importance scoring.

An additional benefit of the conditional formulation~(\ref{conditional}) is that the corresponding score~(\ref{emp-median}) is relatively robust to varying correlation between significant features and between significant and null features. 
In the degenerate case of perfect correlation, the corresponding scores collapse to identical values. 
This robustness contrasts with SHAP, LOCO, and permutation importance, which can be sensitive to feature correlation \cite{verdinelli-wasserman,hooker}.

\begin{figure}[t!]
\centering
\includegraphics[width=0.6\textwidth]{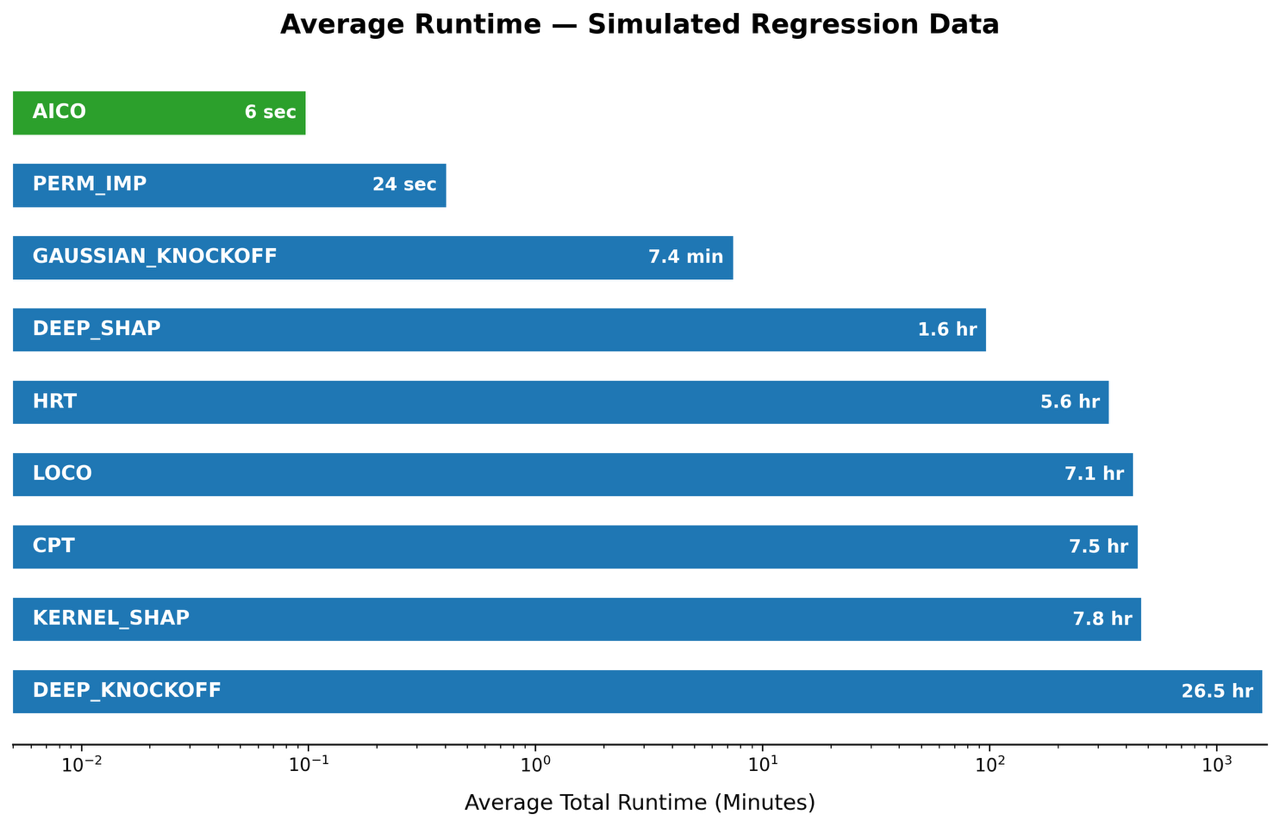}
\caption{Average computation times for testing/screening all 19 features (log-scale). The compute environment consists of 8GB memory and four CPU cores (Intel Xeon CPU E5-2670, 2.60GHz, and Silver 4110 CPU, 2.10GHz).}
\label{eff}
\end{figure}


Figure~\ref{eff} reports average computation times across ten experiments. In each experiment, we apply AICO, CPT, HRT, and LOCO to test all 19 features; apply the knockoff procedures to screen all 19 features; and compute SHAP and permutation importance scores for all 19 features. The reported times include any training of submodels or auxiliary models but exclude the time required to train the original neural network~(\ref{nn}).

AICO achieves an unprecedented degree of computational efficiency. 
It is more than three orders of magnitude faster than HRT, the second most efficient testing method considered. 
It is nearly two orders of magnitude faster than Gaussian knockoffs, the fastest screening procedure among those implemented, and four times faster than permutation importance, the most efficient algorithmic feature-importance method considered in our experiments. 
As shown earlier in this section, AICO’s computational efficiency does not come at the expense of statistical performance.


\section{Empirical applications}\label{sec-empirical}
\subsection{Credit card default}\label{credit}
This section complements the theoretical and numerical results with an empirical application to credit-card payment data from 30{,}000 cardholders of a Taiwanese issuer, collected in October~2005.\footnote{The dataset is available at  \url{https://archive.ics.uci.edu/dataset/350/default+of+credit+card+clients}.} 
The response variable is a binary indicator of payment default, and there are 23 explanatory variables (see Table~\ref{table-uci-features}). 
The dataset is split into training (60\%), validation (15\%), and testing (25\%) subsets, maintaining class proportions through stratified sampling. 
Categorical variables are one-hot encoded, and continuous variables are standardized to mean~0 and unit variance using training-set statistics. 


\begin{table}[t!]
\centering
\renewcommand{\arraystretch}{1.1} 
\setlength{\tabcolsep}{2pt} 
\footnotesize
\resizebox{\columnwidth}{!}{ 
\begin{tabular}{@{}llp{5cm}p{5cm}@{}}
\toprule
\textbf{Feature} & \textbf{Data Type} & \textbf{Levels} & \textbf{Description} \\
\midrule
SEX                    & Categorical & Male, Female & Sex of the cardholder. \\
MARRIAGE               & Categorical & Married, Single, Other & Marital status of the cardholder. \\
EDUCATION              & Categorical & Graduate, University, High school, Other & Education level of the cardholder. \\
PAY\_0, PAY\_2--PAY\_6 & Categorical & Paid in full, $1 \ldots 9+$ months delinquent & Repayment status for the six months from Sept~2005 (PAY\_0) back to Apr~2005 (PAY\_6). \\
\midrule
AGE                    & Continuous & -- & Age of the cardholder (in years). \\
BILL\_AMT1--BILL\_AMT6 & Continuous & -- & Bill statement amount (NT dollars) from Sept~2005 (BILL\_AMT1) back to Apr~2005 (BILL\_AMT6). \\
LIMIT\_BAL             & Continuous & -- & Amount of the given credit (NT dollars). \\
PAY\_AMT1--PAY\_AMT6   & Continuous & -- & Amount paid (NT dollars) from Sept~2005 (PAY\_AMT1) back to Apr~2005 (PAY\_AMT6). \\
\bottomrule
\end{tabular}} 
\caption{Raw features of the credit card default dataset.}
\label{table-uci-features}
\end{table}

We implement an XGBoost classifier~\cite{chen2016xgboost} using a cross-entropy loss function. 
Model performance is evaluated by the area under the receiver operating characteristic (ROC) curve (AUC) on the validation set, which serves as the criterion for hyper-parameter optimization. 
We use the Optuna package~\cite{akiba2019optuna} with the Tree-Structured Parzen Estimator~\cite{bergstra2011algorithms}, conducting 250 trials. 
The search space includes the boosting algorithm, tree-specific parameters such as maximum depth and minimum child weight, and regularization terms. 
Class imbalance is mitigated using a scaling-weight parameter, and training is capped at 1{,}000 boosting rounds with early stopping after 50 rounds without improvement in validation AUC. 
The optimal configuration employs gradient tree boosting~\cite{friedman2001greedy} with an approximation-based tree-building strategy, an $L_2$ regularization parameter of 0.2, an $L_1$ regularization parameter of 0.001, a subsampling ratio of 100\%, a column-sampling ratio per tree of 65\%, a maximum tree depth of six, a minimum child weight of two, a learning rate of 0.05, a minimum loss reduction of 0.00004, and a loss-guided tree-growth policy. 

We train an ensemble of 100 XGBoost models on 100 bootstrap samples of the training data, all using the tuned hyper-parameters. 
Ensemble predictions are obtained by averaging the outputs of all 100 models. 
The resulting ensemble achieves AUCs of $0.837$ on the training set, $0.782$ on the validation set, and $0.786$ on the test set. 
The results are robust to the model choice; for example, an ensemble of neural networks yields nearly identical outcomes.

We adopt the same conditional feature-effect framework as in Section~\ref{sec-numerical}. 
Specifically, we use the masking and unmasking functions described in Example~\ref{ex1}, with reference values defined by the training mean for continuous features and the adjusted training mode for discrete features.\footnote{We treat each categorical variable as a single discrete feature rather than analyzing its associated one-hot encodings separately. 
If higher resolution is desired, the test can alternatively be run for each individual level of a categorical variable.} We use the cross-entropy loss in (\ref{old-delta}). 

Table~\ref{table-uci-result} summarizes the test results for $\alpha=1\%$ ($N=7{,}500$ and $M_0=0$ in (\ref{hypotheses})).
The features identified as significant, ranked by the empirical median~(\ref{emp-median}), are \texttt{PAY\_0} and \texttt{PAY\_2} through \texttt{PAY\_6}, \texttt{MARRIAGE}, \texttt{SEX}, and \texttt{AGE}.
Past delinquencies are strong predictors of future payment default, with more recent payment records exerting greater influence.
For actual monthly bill and payment amounts (e.g., \texttt{PAY\_AMT} and \texttt{BILL\_AMT}), the test does not reject the null.
Marital status, sex, and age are identified as significant predictors of default even after controlling for past payment behavior.
However, their influence is relatively small, especially for age: the corresponding medians, which measure effect size, are at least an order of magnitude smaller than those for the payment-history features.
These findings suggest that personal circumstances exert a statistically significant, but comparatively modest, predictive influence on borrower behavior beyond patterns in prior payment.\footnote{In the United States, marital status, sex, and age are protected characteristics under fair lending laws and therefore are not used---indeed, usually not even collected---for estimating default probabilities or credit scores in credit underwriting.}


\begin{table}[t!]
\centering
\renewcommand{\arraystretch}{1.1} 
\setlength{\tabcolsep}{2.5pt}       
\small
\resizebox{\columnwidth}{!}{%
\begin{tabular}{@{}lcccccccccccc@{}}
\toprule
\multicolumn{12}{c}{\textbf{AICO, Level} $\alpha=1\%$} \\
\toprule
\multicolumn{4}{c}{} 
 & \multicolumn{2}{c}{$p$-value Distribution} 
 & \multicolumn{3}{c}{Randomized One-Sided CI} 
 & \multicolumn{3}{c}{Non-Randomized Two-Sided CI} \\
\cmidrule(lr){5-6} \cmidrule(lr){7-9} \cmidrule(lr){10-12}
Rank & Variable   & Median     & P(reject) 
     & Lower      & Upper      
     & 23.4\%     & 76.6\%     & Coverage   
     & Lower      & Upper      & Coverage   \\
\midrule
1    & PAY\_0     & 0.150      & 1.000   
     & 0.000      & 0.000      
     & 0.143      & 0.143      & 0.99    
     & 0.142      & 0.155      & 0.99063    \\
2    & PAY\_2     & 0.121      & 1.000   
     & 0.000      & 0.000      
     & 0.117      & 0.117      & 0.99    
     & 0.116      & 0.125      & 0.99063    \\
3    & PAY\_3     & 0.095      & 1.000   
     & 0.000      & 0.000      
     & 0.091      & 0.091      & 0.99    
     & 0.091      & 0.099      & 0.99063    \\
4    & PAY\_4     & 0.059      & 1.000   
     & 0.000      & 0.000      
     & 0.055      & 0.055      & 0.99    
     & 0.054      & 0.064      & 0.99063    \\
5    & PAY\_5     & 0.037      & 1.000   
     & 2.520e-322 & 6.423e-322 
     & 0.035      & 0.035      & 0.99    
     & 0.035      & 0.039      & 0.99063    \\
6    & PAY\_6     & 0.030      & 1.000   
     & 1.020e-297 & 2.493e-297 
     & 0.029      & 0.029      & 0.99    
     & 0.029      & 0.032      & 0.99063    \\
7    & MARRIAGE   & 0.002      & 1.000   
     & 3.912e-11  & 4.561e-11  
     & 0.002      & 0.002      & 0.99    
     & 0.001      & 0.003      & 0.99063    \\
8    & SEX        & 0.002      & 1.000   
     & 1.026e-37  & 1.383e-37  
     & 0.001      & 0.002      & 0.99    
     & 0.001      & 0.002      & 0.99063    \\
9    & AGE        & 3.796e-04  & 1.000   
     & 1.788e-09  & 2.056e-09  
     & 2.200e-04  & 2.204e-04  & 0.99    
     & 2.036e-04  & 5.814e-04  & 0.99063    \\
\midrule
-    & LIMIT\_BAL & 0.000      & 0.000   
     & 0.794      & 0.800      
     & 0.000      & 0.000      & 0.99    
     & 0.000      & 3.603e-04  & 0.99496    \\
-    & BILL\_AMT1 & -0.006     & 0.000   
     & 1.000      & 1.000      
     & -0.007     & -0.007     & 0.99    
     & -0.008     & -0.005     & 0.99063    \\
-    & BILL\_AMT2 & -0.001     & 0.000   
     & 1.000      & 1.000      
     & -0.002     & -0.002     & 0.99    
     & -0.002     & -9.540e-04 & 0.99063    \\
-    & BILL\_AMT3 & 0.000      & 0.000   
     & 0.975      & 0.976      
     & -1.599e-04 & -1.592e-04 & 0.99    
     & -1.751e-04 & 5.542e-05  & 0.99063    \\
-    & BILL\_AMT4 & -4.094e-04 & 0.000   
     & 1.000      & 1.000      
     & -5.298e-04 & -5.284e-04 & 0.99    
     & -5.504e-04 & -2.773e-04 & 0.99063    \\
-    & BILL\_AMT5 & -3.203e-04 & 0.000   
     & 1.000      & 1.000      
     & -5.070e-04 & -5.054e-04 & 0.99    
     & -5.220e-04 & -1.887e-04 & 0.99063    \\
-    & BILL\_AMT6 & -1.176e-04 & 0.000   
     & 1.000      & 1.000      
     & -2.041e-04 & -2.023e-04 & 0.99    
     & -2.160e-04 & -3.890e-05 & 0.99063    \\
-    & PAY\_AMT1  & -0.012     & 0.000   
     & 1.000      & 1.000      
     & -0.014     & -0.014     & 0.99    
     & -0.014     & -0.010     & 0.99063    \\
-    & PAY\_AMT2  & -0.008     & 0.000   
     & 1.000      & 1.000      
     & -0.009     & -0.009     & 0.99    
     & -0.010     & -0.007     & 0.99063    \\
-    & PAY\_AMT3  & -0.006     & 0.000   
     & 1.000      & 1.000      
     & -0.006     & -0.006     & 0.99    
     & -0.006     & -0.005     & 0.99063    \\
-    & PAY\_AMT4  & -3.764e-04 & 0.000   
     & 1.000      & 1.000      
     & -5.811e-04 & -5.779e-04 & 0.99    
     & -5.998e-04 & -1.935e-04 & 0.99063    \\
-    & PAY\_AMT5  & -0.002     & 0.000   
     & 1.000      & 1.000      
     & -0.002     & -0.002     & 0.99    
     & -0.002     & -0.002     & 0.99063    \\
-    & PAY\_AMT6  & -0.001     & 0.000   
     & 1.000      & 1.000      
     & -0.002     & -0.001     & 0.99    
     & -0.002     & -0.001     & 0.99063    \\
-    & EDUCATION  & -3.891e-04 & 0.000   
     & 0.848      & 0.853      
     & -0.001     & -0.001     & 0.99    
     & -0.001     & 4.611e-04  & 0.99063    \\
\bottomrule
\end{tabular}%
}
\caption{AICO test summary for the credit card default dataset. Percentage heads: probabilities of the two possible lower endpoints of the randomized CI.}
\label{table-uci-result}
\end{table}

We can further explore the role of individual features by running the test on subsets of the test set. 
For instance, Table~\ref{table-result-uci-single} reports results for the 4{,}029 (out of 7{,}500) samples for which $\texttt{MARRIAGE}=\text{Single}$. 
While the test does not reject the null for $\texttt{EDUCATION}$ at the $1\%$ level, it rejects at the $5\%$ level, as the distribution of its $p$-value is concentrated below $0.05$. 
When the test is restricted to samples with $\texttt{MARRIAGE}=\text{Married}$, the null for $\texttt{EDUCATION}$ is no longer rejected. 
These findings provide evidence that educational level contributes predictive information for unmarried borrowers, but provide no comparable evidence for married borrowers.

\begin{table}[t!]
\centering
\renewcommand{\arraystretch}{1.1} 
\setlength{\tabcolsep}{2.5pt}       
\small
\resizebox{\columnwidth}{!}{%
\begin{tabular}{@{}lcccccccccccc@{}}
\toprule
\multicolumn{12}{c}{\textbf{AICO, Conditional on MARRIAGE = Single, Level} $\alpha=1\%$} \\
\toprule
\multicolumn{4}{c}{} & \multicolumn{2}{c}{$p$-value Distribution} & \multicolumn{3}{c}{Randomized One-Sided CI} & \multicolumn{3}{c}{Non-Randomized Two-Sided CI} \\
\cmidrule(lr){5-6} \cmidrule(lr){7-9} \cmidrule(lr){10-12}
Rank & Variable & Median & P(reject) & Lower & Upper & 83.0\% & 17.0\% & Coverage & Lower & Upper & Coverage \\
\midrule
1  & PAY\_0   & 0.151    & 1.000 & 1.833e-248 & 5.763e-248 & 0.143  & 0.143  & 0.99 & 0.142  & 0.158  & 0.99023 \\
2  & PAY\_2   & 0.130    & 1.000 & 0.000 & 0.000 & 0.125  & 0.125  & 0.99 & 0.124  & 0.136  & 0.99023 \\
3  & PAY\_3   & 0.100    & 1.000 & 1.072e-295 & 3.822e-295 & 0.095  & 0.095  & 0.99 & 0.095  & 0.105  & 0.99023 \\
4  & PAY\_4   & 0.063    & 1.000 & 1.402e-260 & 4.555e-260 & 0.056  & 0.057  & 0.99 & 0.056  & 0.069  & 0.99023 \\
5  & PAY\_5   & 0.040    & 1.000 & 4.099e-183 & 1.071e-182 & 0.037  & 0.037  & 0.99 & 0.037  & 0.044  & 0.99023 \\
6  & PAY\_6   & 0.029    & 1.000 & 3.112e-171 & 7.848e-171 & 0.028  & 0.028  & 0.99 & 0.027  & 0.031  & 0.99023 \\
7  & AGE      & 0.002    & 1.000 & 2.601e-48  & 4.147e-48  & 0.002  & 0.002  & 0.99 & 0.002  & 0.002  & 0.99023 \\
8  & SEX      & 0.001    & 1.000 & 1.972e-16  & 2.560e-16  & 0.001  & 0.001  & 0.99 & 0.001  & 0.002  & 0.99023 \\
\midrule
-  & LIMIT\_BAL & -4.292e-04 & 0.000 & 1.000      & 1.000      & -0.002 & -0.002 & 0.99 & -0.002     & 0.000      & 0.99511 \\
-  & BILL\_AMT1 & -0.005     & 0.000 & 1.000      & 1.000      & -0.007 & -0.007 & 0.99 & -0.007     & -0.003     & 0.99023 \\
-  & BILL\_AMT2 & -0.001     & 0.000 & 1.000      & 1.000      & -0.002 & -0.002 & 0.99 & -0.002     & -8.918e-04 & 0.99023 \\
-  & BILL\_AMT3 & 4.366e-05  & 0.000 & 0.376      & 0.388      & -1.597e-05 & -1.038e-05 & 0.99 & -3.777e-05 & 2.814e-04  & 0.99023 \\
-  & BILL\_AMT4 & -4.339e-04 & 0.000 & 1.000      & 1.000      & -6.109e-04 & -6.105e-04 & 0.99 & -6.237e-04 & -2.882e-04 & 0.99023 \\
-  & BILL\_AMT5 & -1.737e-04 & 0.000 & 1.000      & 1.000      & -3.338e-04 & -3.327e-04 & 0.99 & -3.612e-04 & -3.330e-05 & 0.99023 \\
-  & BILL\_AMT6 & -1.147e-04 & 0.000 & 1.000      & 1.000      & -2.326e-04 & -2.315e-04 & 0.99 & -2.509e-04 & 0.000      & 0.99474 \\
-  & PAY\_AMT1  & -0.014     & 0.000 & 1.000      & 1.000      & -0.015     & -0.015     & 0.99 & -0.016     & -0.011     & 0.99023 \\
-  & PAY\_AMT2  & -0.010     & 0.000 & 1.000      & 1.000      & -0.012     & -0.012     & 0.99 & -0.012     & -0.009     & 0.99023 \\
-  & PAY\_AMT3  & -0.006     & 0.000 & 1.000      & 1.000      & -0.008     & -0.008     & 0.99 & -0.008     & -0.005     & 0.99023 \\
-  & PAY\_AMT4  & -4.392e-04 & 0.000 & 1.000      & 1.000      & -7.181e-04 & -7.118e-04 & 0.99 & -7.488e-04 & -1.825e-04 & 0.99023 \\
-  & PAY\_AMT5  & -0.002     & 0.000 & 1.000      & 1.000      & -0.003     & -0.003     & 0.99 & -0.003     & -0.002     & 0.99023 \\
-  & PAY\_AMT6  & -0.002     & 0.000 & 1.000      & 1.000      & -0.002     & -0.002     & 0.99 & -0.002     & -0.001     & 0.99023 \\
-  & EDUCATION  & 0.001      & 0.000 & 0.012      & 0.013      & -7.817e-05 & -4.987e-05 & 0.99 & -2.034e-04 & 0.002      & 0.99023 \\
\bottomrule
\end{tabular}%
}
\caption{AICO test summary for the credit card default dataset. The test is conditioned on the samples for which $\texttt{MARRIAGE}=\text{Single}$. Percentage heads: probabilities of the two possible lower endpoints of the randomized CI.}
\label{table-result-uci-single}
\end{table}


\subsection{Mortgage risk}
We next examine a more challenging classification setting, predicting the state of a mortgage loan in the next month given a history of time-varying features observed at the beginning of the month. 
The seven states are \emph{Current}, \emph{30~Days Delinquent}, \emph{60~Days Delinquent}, \emph{90+~Days Delinquent}, \emph{Foreclosure}, \emph{Paid Off}, and \emph{Real Estate Owned}. 
The analysis is based on an extensive dataset of mortgages from four ZIP codes in the greater Los Angeles area, tracked monthly from 1/1994 through 12/2022. 
This dataset constitutes a subset of the CoreLogic Loan-Level Market Analytics data \cite{dlmr}.
There are 52 borrower and macroeconomic features such as current loan state, credit score, loan balance, market interest rate, and local unemployment rate. 

We apply the data-filtering procedure of \ci{dlmr}, resulting in roughly five million loan-month transitions from 117{,}523 active loans. 
Following \ci{dlmr} and related work, samples with missing feature values are handled using missingness indicators and forward-filling based on the last observed value for each loan; see \ci{vanness2023missing} for an analysis of this approach. 
The training, validation, and test splits are 1/1994--6/2009, 7/2009--12/2009, and 1/2010--12/2022, respectively. 
We adopt the set--sequence model of \ci{elliot} with LongConv~\cite{elliot-seq} for the sequence layer, which represents the current state of the art in mortgage transition modeling. 
Training and validation procedures follow those of \ci{elliot}. 



The loan-month samples form a panel dataset. 
Appendix~\ref{panel} explains how such a data structure can be handled in AICO. 
Specifically, we follow the grouping approach in Appendix~\ref{grouping}, treating each mortgage as a unit $i$ with an associated group consisting of loan-month samples observed over its time horizon. 
Any dependence across these monthly samples is thereby contained within each group.

For testing, we use the \emph{mean} multi-class cross-entropy loss, averaged over monthly time steps within each loan group. 
The multi-class loss accommodates all possible next-month states, while the temporal averaging enables joint testing of feature significance across all transitions in a loan’s trajectory.\footnote{Alternative formulations are also possible for both the per-period loss and the temporal aggregation. For example, one may use binary cross-entropy as the per-period loss to target whether the next state is \emph{Paid Off} versus any other state, or aggregate per-period losses via the maximum to emphasize extreme misclassification events.}

We adopt the same conditional feature-effect framework as in Section~\ref{credit}. 
Specifically, we use the masking specification~(\ref{conditional}), with masking reference values $v^\ell$ constructed from the training set $I_1$ via a two-step aggregation: first from the loan-month level to the loan level, and then across loans. 
For each tested feature $\ell$, the unit-level feature value $X_i^\ell$ is obtained by aggregating observations within unit $i$'s group, using the mean for continuous features and the adjusted mode for discrete and categorical features. 
The reference value $v^\ell$ in (\ref{conditional}) is then constructed from the unit-level values $\{X_i^\ell : i \in I_1\}$. 
This two-step aggregation ensures uniform weighting across loans when constructing reference values, regardless of the number of observations available for each loan.

To avoid ambiguity, bias, and dependence on the imputation method, loan-month samples with missing values are excluded from the test set, as discussed in Appendix~\ref{missing}. 
Because the extent of missingness varies across features, each test is conducted on a different effective sample size. 
Moreover, the test is performed at the variable level, so all one-hot (dummy) features corresponding to a categorical variable are tested jointly. 
This yields a total of 28 tested variables.

Table~\ref{table-mortgage-risk} reports the results ($M_0=0$ in (\ref{hypotheses})).
The most important feature by far is the current state of the mortgage.
The market mortgage rate (\texttt{natl\_mortgage\_rate}) is the second most important feature, underscoring the influence of time-varying macroeconomic conditions on borrower behavior.
Borrowers have a strong incentive to refinance (i.e., prepay) their mortgage when the market rate falls below their own loan rate, which is ranked ninth in importance.
The results also reveal pronounced path dependence:
the number of months a borrower remained current and the number of months 30~days delinquent during the past 12~months are both identified as significant and are economically important predictors of borrower behavior.

\begin{table}[t!]
\centering
\renewcommand{\arraystretch}{1.1}
\setlength{\tabcolsep}{1.5pt}
\small
\resizebox{\columnwidth}{!}{%
\begin{tabular}{@{}lcccccccccccccc@{}}
\toprule
\multicolumn{14}{c}{\textbf{AICO, Level} $\alpha=1\%$} \\
\toprule
\multicolumn{4}{c}{} 
 & \multicolumn{2}{c}{$p$-value Distribution} 
 & \multicolumn{4}{c}{Randomized One-Sided CI} 
 & \multicolumn{3}{c}{Non-Randomized Two-Sided CI} 
 & \multicolumn{1}{c}{} \\
\cmidrule(lr){5-6} \cmidrule(lr){7-10} \cmidrule(lr){11-13}
Rank & Variable   & Median     & P(reject) 
     & Lower      & Upper      
     & Lower-1    & Lower-2    & P(lwr-2) & Coverage
     & Lower      & Upper      & Coverage
     & Missing rate \\
\midrule
1  & current\_state                  & 0.560       & 1.000   & 0.000       & 0.000       & 0.557       & 0.557       & 0.625  & 0.99 & 0.556       & 0.564       & 0.9902 & 0.000 \\
2  & natl\_mortgage\_rate        & 0.006       & 1.000   & 5.372e-66   & 6.297e-66   & 0.005       & 0.005       & 0.625  & 0.99 & 0.005       & 0.006       & 0.9902 & 0.000 \\
3  & pool\_insurance\_flag    & 0.005       & 1.000   & 1.031e-240  & 1.560e-240  & 0.005       & 0.005       & 0.032  & 0.99 & 0.005       & 0.006       & 0.9903 & 0.443 \\
4  & times\_current                  & 0.002       & 1.000   & 5.920e-178  & 7.708e-178  & 0.002       & 0.002       & 0.625  & 0.99 & 0.002       & 0.003       & 0.9902 & 0.000 \\
5  & times\_30dd\_value              & 0.002       & 1.000   & 0.000       & 0.000       & 0.001       & 0.001       & 0.393  & 0.99 & 0.001       & 0.002       & 0.9902 & 0.014 \\
6  & loan\_age                       & 7.823e-04   & 1.000   & 3.830e-18   & 4.151e-18   & 5.401e-04   & 5.407e-04   & 0.625  & 0.99 & 5.230e-04   & 0.001       & 0.9902 & 0.000 \\
7  & collateral\_type & 7.683e-04 & 1.000   & 2.425e-70   & 2.882e-70   & 6.749e-04   & 6.750e-04   & 0.535  & 0.99 & 6.652e-04   & 8.658e-04   & 0.9902 & 0.096 \\
8  & lagged\_prepay\_rate        & 7.518e-04   & 1.000   & 7.103e-127  & 8.870e-127  & 6.730e-04   & 6.732e-04   & 0.625  & 0.99 & 6.662e-04   & 8.298e-04   & 0.9903 & 0.000 \\
9  & current\_interest\_rate  & 6.571e-04   & 1.000   & 8.549e-85   & 1.024e-84   & 5.789e-04   & 5.792e-04   & 0.530  & 0.99 & 5.716e-04   & 7.453e-04   & 0.9902 & 0.003 \\
10 & mba\_days\_delinquent    & 3.738e-04   & 1.000   & 1.890e-131  & 2.370e-131  & 3.379e-04   & 3.380e-04   & 0.625  & 0.99 & 3.341e-04   & 4.136e-04   & 0.9902 & 0.000 \\
11 & prepay\_penalty\_flag    & 2.733e-04   & 1.000   & 4.209e-56   & 4.903e-56   & 2.365e-04   & 2.366e-04   & 0.078  & 0.99 & 2.317e-04   & 3.184e-04   & 0.9901 & 0.085 \\
12 & initial\_interest\_rate         & 2.199e-04   & 1.000   & 1.219e-27   & 1.349e-27   & 1.775e-04   & 1.776e-04   & 0.625  & 0.99 & 1.751e-04   & 2.671e-04   & 0.9902 & 0.000 \\
13 & buydown\_flag            & 2.192e-04   & 1.000   & 1.214e-68   & 1.461e-68   & 1.882e-04   & 1.883e-04   & 0.160  & 0.99 & 1.856e-04   & 2.537e-04   & 0.9903 & 0.235 \\
14 & times\_foreclosure              & 2.032e-04   & 1.000   & 2.492e-46   & 2.844e-46   & 1.680e-04   & 1.681e-04   & 0.625  & 0.99 & 1.631e-04   & 2.454e-04   & 0.9903 & 0.000 \\
15 & original\_balance               & 1.815e-04   & 1.000   & 4.651e-43   & 5.283e-43   & 1.505e-04   & 1.505e-04   & 0.625  & 0.99 & 1.479e-04   & 2.178e-04   & 0.9902 & 0.000 \\
16 & fico\_score\_at\_orig    & 1.233e-04   & 1.000   & 4.995e-36   & 5.609e-36   & 9.650e-05   & 9.669e-05   & 0.625  & 0.99 & 9.399e-05   & 1.515e-04   & 0.9902 & 0.000 \\
17 & sched\_monthly\_pi   & 3.616e-05   & 1.000   & 4.464e-32   & 4.979e-32   & 2.805e-05   & 2.809e-05   & 0.067  & 0.99 & 2.688e-05   & 4.472e-05   & 0.9900 & 0.003 \\
\midrule
-  & original\_ltv                   & 1.874e-05   & 0.000   & 0.017       & 0.018       & -2.449e-06  & -2.418e-06  & 0.625  & 0.99 & -4.904e-06  & 4.245e-05   & 0.9902 & 0.000 \\
-  & unemployment\_rate              & -3.907e-04  & 0.000   & 1.000       & 1.000       & -4.822e-04  & -4.813e-04  & 0.625  & 0.99 & -4.909e-04  & -2.838e-04  & 0.9902 & 0.000 \\
-  & current\_balance         & -1.298e-06  & 0.000   & 0.520       & 0.524       & -3.206e-05  & -3.195e-05  & 0.625  & 0.99 & -3.507e-05  & 3.639e-05   & 0.9902 & 0.000 \\
-  & sched\_principal     & -5.314e-05  & 0.000   & 1.000       & 1.000       & -5.989e-05  & -5.983e-05  & 0.980  & 0.99 & -6.061e-05  & -4.456e-05  & 0.9901 & 0.573 \\
-  & convertible\_flag        & -5.164e-04  & 0.000   & 1.000       & 1.000       & -6.553e-04  & -6.551e-04  & 0.034  & 0.99 & -6.722e-04  & -3.525e-04  & 0.9900 & 0.555 \\
-  & io\_flag\_value                  & -4.553e-05  & 0.000   & 0.973       & 0.973       & -1.026e-04  & -1.020e-04  & 0.868  & 0.99 & -1.091e-04  & 1.431e-05   & 0.9902 & 0.324 \\
-  & neg\_amortiz\_flag & -8.336e-04 & 0.000 & 1.000       & 1.000       & -0.001      & -0.001      & 0.825  & 0.99 & -0.001      & -4.979e-04  & 0.9903 & 0.853 \\
-  & original\_term            & -1.041e-04  & 0.000   & 0.999       & 0.999       & -1.843e-04  & -1.843e-04  & 0.139  & 0.99 & -1.924e-04  & -1.752e-05  & 0.9900 & 0.0001 \\
-  & times\_60dd                       & -3.928e-04  & 0.000   & 1.000       & 1.000       & -5.032e-04  & -5.024e-04  & 0.625  & 0.99 & -5.118e-04  & -2.708e-04  & 0.9902 & 0.000 \\
-  & times\_90dd                       & -4.781e-04  & 0.000   & 1.000       & 1.000       & -5.630e-04  & -5.630e-04  & 0.625  & 0.99 & -5.723e-04  & -3.823e-04  & 0.9902 & 0.000 \\
-  & lagged\_forecl\_rate         & -1.049e-04  & 0.000   & 0.998       & 0.998       & -1.769e-04  & -1.765e-04  & 0.625  & 0.99 & -1.846e-04  & -9.468e-06  & 0.9902 & 0.000 \\
\bottomrule
\end{tabular}%
}
\caption{AICO test summary for the mortgage risk dataset using the set-sequence model. Training data is grouped and treated at the loan-month level. The \textit{missing rate} denotes the ratio of feature effects dropped from the test set due to missingness to the total number of loan-level feature effects after grouping (46,890). \textit{P(lwr-2)} represents the probability associated with the larger lower bound (\textit{lower-2}) of the one-sided randomized CI.}
\label{table-mortgage-risk}
\end{table}



\section{Conclusion}\label{conclusion}
This paper develops a simple and computationally efficient framework for assessing feature significance in supervised learning.
The proposed randomized sign test delivers exact $p$-values and confidence intervals for effect size under minimal assumptions, avoiding the limitations of retraining- and surrogate-model-based approaches.
Empirical studies demonstrate robust performance on synthetic data and practical value in real-world applications such as credit scoring and mortgage risk modeling, illustrating the method’s ability to inform high-stakes decisions with legal, economic, and regulatory implications.
More broadly, the framework provides a scalable, statistically grounded approach to interpretable machine learning that supports trustworthy, evidence-based decision-making in science, industry, and policy.

\section*{Acknowledgments}
We are very grateful to Pulkit Goel and Chenyu Song for exceptional research assistance. We thank Rakshitha Ireddi and Yashwanth Devavarapu for comments. We are also grateful for financial support from the Stanford Institute for Human-Centered Artificial Intelligence (HAI). A preliminary version of this work appeared under the title ``Computationally Efficient Feature Significance and Importance for Predictive Models'' (ICAIF 2022).

\section*{Data availability}
A Python code package is available at \url{https://github.com/ji-chartsiri/AICO}.

\FloatBarrier
\appendix
\renewcommand{\thesection}{S\arabic{section}}
\renewcommand{\thesubsection}{S\arabic{section}.\arabic{subsection}}
\renewcommand{\thesubsubsection}{S\arabic{section}.\arabic{subsection}.\arabic{subsubsection}}
\setcounter{section}{0}
\setcounter{equation}{0}
\renewcommand{\theequation}{S\arabic{equation}}
\setcounter{table}{0}
\renewcommand{\thetable}{S\arabic{table}}
\setcounter{figure}{0}
\renewcommand{\thefigure}{S\arabic{figure}}

\section{Proofs of technical results}\label{proofs}
\begin{proof}[Proof of Proposition \ref{ump}]
Due to Assumption \ref{assumption-a}, $n^\ell_+(w)\sim B_{N,\theta^\ell_w}$ under the sampling distribution $\bbp_{M^\ell}$, where $\theta^\ell_w=1-F^\ell(w)$. The Binomial distribution $B_{N,\theta^\ell_w}$ has a monotone likelihood ratio, so the Neyman-Pearson lemma \cite[Theorem 3.4.1]{lehmann2006testing} guarantees the existence of a uniformly most powerful (UMP) test for the hypotheses $$\theta^\ell_w\le \theta_0\quad\text{vs.}\quad\theta^\ell_w>\theta_0$$ for some $\theta_0\in[0,1]$. Due to the continuity of $F^\ell$, we have $F^\ell(M^\ell)=1/2$ so for $\theta_0=1/2$ and $w=M_0$ these hypotheses are equivalent to (\ref{hypotheses}). The UMP test has test function $\phi(n)=1\{n> T\} + \gamma 1\{n= T\}$ for some $T\in\bbn$ and $\gamma\in[0,1]$ determined by the requirement that the test's level of significance at the boundary $M^\ell = M_0$ be equal to $\alpha$. Specifically, for any $M_0\in \bbr$,
\begin{align*}
\alpha&=\bbe_{M_0}(\phi(n^\ell_+(M_0)))\\
&=\bbp_{M_0}(n^\ell_+(M_0)>T) + \bbe_{M_0}(\gamma | n^\ell_+(M_0)=T)\bbp_{M_0}(n^\ell_+(M_0)=T), \nonumber
\end{align*}
where $\bbe_{M_0}$ represents expectation under $\bbp_{M_0}$, the sampling distribution $\bbp_{M^\ell}$ of Assumption \ref{assumption-a} for $M^\ell=M_0$. Clearly $n^\ell_+(M_0)\sim B_{N,1/2}$ under $\bbp_{M_0}$. Moreover, 
it follows that $T=T_{N,\alpha}=q_{1-\alpha}(B_{N, 1/2})$ and 
\begin{equation*}
\gamma=\gamma_N(T,\alpha)=\frac{\alpha-\bbp_{M_0}(n^\ell_+(M_0)>T)}{\bbp_{M_0}(n^\ell_+(M_0)=T)}, 
\end{equation*}
where $\bbp_{M_0}(n^\ell_+(M_0)\le n)=\Pi_{N,1/2}(n)$ and $\bbp_{M_0}(n^\ell_+(M_0)=n)=\pi_{N,1/2}(n)$. This gives (\ref{gamma}).
\end{proof}

\begin{proof}[Proof of Proposition \ref{ci-prop}]
Let $U\sim U(0, 1)$ and $I^\ell(w) = 1 - \phi_N(n^\ell_+(w), \alpha)$, where we recall $n^\ell_+(w)=\#\{i\in I_2: \Delta^\ell(Z_i)> w\}$ for $w\in\bbr$, as in Proposition \ref{ump}. From \ci{geyer-meeden}, a randomized CI for $M^\ell$ is given by the $U$-cut of $I^\ell(w)$:
		$$C^\ell_N(\alpha;U) = \{w \in \mathbb{R} \; | \; I^\ell(w) \geq U\}.$$
Then, by iterated expectations, the independence of $U$, and Eq.~(\ref{size}) we get
\begin{align*}
\bbp_{M^\ell}(M^\ell\in C^\ell_N(\alpha;U)) 
&= \bbp_{M^\ell}(U\le I^\ell(M^\ell))\\
&=\bbe_{M^\ell}(1 - \phi_N(n^\ell_+(M^\ell), \alpha))\\
&=1-\alpha.
\end{align*}

Since $n^\ell_{+}(M^\ell) = N - i$ for $M^\ell \in [\Delta^\ell_{(i)}, \Delta^\ell_{(i+1)})$, we have
\begin{align*}
    C^\ell_N(\alpha;U)
    =  {\textstyle\bigcup} \: \{[\Delta^\ell_{(i)}, \Delta^\ell_{(i +1)}) \; | \; i = 0, \ldots, N;\;1 - \phi_N(N - i, \alpha) \geq U\}.
\end{align*}
Moreover, note that 
\begin{align*}
	1 - \phi_N(N - i, \alpha)
	&=1\{N-i\le T_{N,\alpha}\}-\gamma_N(T_{N,\alpha},\alpha)1\{N-i = T_{N,\alpha}\}\\
	&=1\{N-T_{N,\alpha}<i\}+(1-\gamma_N(T_{N,\alpha},\alpha))1\{N-T_{N,\alpha}=i\}.
\end{align*}
Thus $C^\ell_N(\alpha;U)$ is equal to the union of $[\Delta^\ell_{(N - T_{N,\alpha})}, \Delta^\ell_{(N-T_{N,\alpha} + 1)})$ and $[\Delta^\ell_{(N - T_{N,\alpha}+1)}, \infty)$ if $1 - \gamma_N(T_{N,\alpha}, \alpha) \geq U$ and $[\Delta^\ell_{(N - T_{N,\alpha}+1)}, \infty)$ otherwise. 
\end{proof}


\section{Additional experimental results}\label{robustness}
Appendix \ref{classification-sect} provides numerical results for a classification setting, complementing the regression results in Section \ref{sec-numerical}. Appendix \ref{app-regress} provides additional robustness results. Appendix \ref{robustness-comp-large} compares the testing performance of AICO with alternative procedures over a large number of trials. Appendix \ref{app-validity} examines the finite-sample validity of the AICO test.

\subsection{Classification}\label{classification-sect}
This section illustrates the robust performance of our test
and feature-importance measure on a synthetic “known-truth” classification task. Consider a binary response $Y \in \{0,1\}$ governed by the logistic model
\begin{align}\label{classification}
	\BE(Y|X) = \bbp(Y=1|X) = g(\mu(X)),
\end{align}
where $g(w)=1/(1+e^w)$, the function $\mu$ is given by~(\ref{mu}), and the features $X=(X^1,\ldots,X^d)$ are defined as in Section~\ref{sec-numerical}. 
We generate a training set $I_1$ of $10^6$ independent samples $Z_i$ and a test set $I_2$ of $5\cdot10^5$ independent samples. 
The data are imbalanced, as is common in practice: approximately 10\% of the samples satisfy $Y_i=1$. 

We train a single-layer neural network $\M$ estimating Eq.~(\ref{classification}), specified as
\begin{align}\label{classification-nn}
	\M(x) = g \bigg(b_0 + \sum_{u=1}^U b_u \psi(a_{0,u} + a_u^\top x)\bigg),
\end{align}
where $\psi(y)=1/(1+e^{-y})$, $U$ is the number of hidden units, and $b_0, b_u, a_{0,u}\in\mathbb{R}$ and $a_u\in\mathbb{R}^d$ are parameters. 
Training uses a standard cross-entropy loss with $L_2$ regularization (weight $10^{-3}$, approximately optimal in the regression setting), the Adam optimizer, a batch size of 32, and 500 epochs with early stopping. 
A validation set (25\% of $I_1$) and a Bayesian search procedure are used to tune the learning rate and number of units~$U$. 
The optimal learning rate is $5\cdot10^{-4}$ and the optimal $U=300$. 
The resulting model achieves an Area Under the ROC Curve exceeding 99\% on both the training and test data.

The testing procedure employs the cross-entropy loss $L(m,y)=-(y\log m + (1-y)\log(1-m))$ in (\ref{old-delta}), consistent with the training objective. 
We adopt the same conditional feature-effect framework as in Section~\ref{sec-numerical}. 
Specifically, we use the masking and unmasking functions described in Example~\ref{ex1}, with reference values defined by the training mean for continuous features and the adjusted training mode for discrete features.

Table~\ref{table-fo-classification} reports the results for significance levels $\alpha=0.01$ and $0.05$ across ten independent experiments, again using the default value $M_0=0$ in (\ref{hypotheses}).
Across all experiments, features entering the data-generating process (DGP) are identified as significant, while features that are null in the DGP are identified as significant only occasionally.
Feature correlation (e.g., $\ell=1,6$), discrete feature distributions (e.g., $\ell=9,10$), and interaction effects (e.g., $\ell=2,4,5$) do not impair performance.

\begin{table}[h]
\centering
\resizebox{0.5\columnwidth}{!}{
\renewcommand{\arraystretch}{1} 
\setlength{\tabcolsep}{4.5pt}      
\begin{tabular}{@{}lcccccccccccc@{}}
\toprule
\multirow{3}{*}{$\ell$} 
& \multicolumn{2}{c}{\multirow{2}{*}{\textbf{AICO}}} 
& \multicolumn{2}{c}{\multirow{2}{*}{\textbf{CPT}}} 
& \multicolumn{2}{c}{\multirow{2}{*}{\textbf{HRT}}} 
& \multicolumn{2}{c}{\multirow{2}{*}{\textbf{LOCO}}} 
& \multicolumn{2}{c}{\textbf{Deep}} 
& \multicolumn{2}{c}{\textbf{Gaussian}} \\
& & & & & & & & & \multicolumn{2}{c}{\textbf{Knockoff}} & \multicolumn{2}{c}{\textbf{Knockoff}} \\
\cmidrule(lr){2-3} \cmidrule(lr){4-5} \cmidrule(lr){6-7} \cmidrule(lr){8-9} \cmidrule(lr){10-11} \cmidrule(lr){12-13}
& 5\% & 1\% & 5\% & 1\% & 5\% & 1\% & 5\% & 1\% & 5\% & 1\% & 5\% & 1\% \\
\midrule
\textbf{1}  & 10 & 10 & 10 & 10 & 10 & 10 & 10 & 10 & 10 & 10 & 10 & 10 \\
\textbf{2}  & 10 & 10 & 10 & 10 & 10 & 10 & 10 & 10 & 10 & 10 & 10 & 10 \\
\textbf{3}  & 10 & 10 & 10 & 10 & 10 & 10 & 10 & 10 & 10 & 10 & 10 & 10 \\
\textbf{4}  & 10 & 10 & 10 & 10 & 10 & 10 & 10 & 10 & 9  & 9  & 10 & 10 \\
\textbf{5}  & 10 & 10 & 10 & 10 & 10 & 10 & 10 & 10 & 10 & 10 & 9  & 9  \\
\textbf{6}  & 10 & 10 & 10 & 10 & 10 & 10 & 10 & 10 & 10 & 10 & 10 & 10 \\
\textbf{7}  & 10 & 10 & 10 & 10 & 10 & 10 & 10 & 10 & 10 & 10 & 10 & 10 \\
\textbf{8}  & 10 & 10 & 10 & 10 & 10 & 10 & 10 & 10 & 9  & 9  & 9  & 9  \\
\textbf{9}  & 10 & 10 & 10 & 10 & 10 & 10 & 10 & 10 & 10 & 10 & 10 & 10 \\
\textbf{10} & 10 & 10 & 10 & 10 & 10 & 10 & 10 & 10 & 10 & 10 & 6  & 6  \\
\textbf{11} & 10 & 10 & 10 & 10 & 10 & 10 & 10 & 10 & 10 & 10 & 10 & 10 \\
\textbf{12} & 10 & 10 & 10 & 10 & 10 & 10 & 10 & 10 & 10 & 10 & 10 & 10 \\
\midrule
\textbf{13} & 1  & 1  & 0 & 0 & 0  & 0  & 4  & 4  & 2  & 2  & 1  & 1  \\
\textbf{14} & 0  & 0  & 2 & 0 & 0  & 0  & 5  & 5  & 1  & 1  & 1  & 1  \\
\textbf{15} & 0  & 0  & 1 & 0 & 2  & 0  & 6  & 6  & 0  & 0  & 3  & 3  \\
\textbf{16} & 0  & 0  & 1 & 1 & 2  & 1  & 5  & 5  & 1  & 1  & 2  & 2  \\
\textbf{17} & 1  & 1  & 0 & 0 & 0  & 0  & 5  & 5  & 0  & 0  & 1  & 1  \\
\textbf{18} & 1  & 1  & 1 & 0 & 0  & 0  & 5  & 5  & 0  & 0  & 2  & 2  \\
\textbf{19} & 1  & 1  & 0 & 0 & 1  & 1  & 3  & 3  & 0  & 0  & 1  & 1  \\
\bottomrule
\end{tabular}}
\caption{Classification model test results for 10 independent training/testing experiments at two standard significance levels ($\alpha = 5\%$ and $\alpha = 1\%$). The table compares AICO, CPT, HRT, LOCO, Deep Knockoff, and Gaussian Knockoff procedures, showing the total number of rejections across 10 trials. The test set size is $N = 5 \cdot 10^5$.}
\label{table-fo-classification}
\end{table}

To assess robustness, we perform 2{,}000 additional independent trials, reported in \ref{robustness-large}.
These experiments corroborate the performance shown in Table~\ref{table-fo-classification}.
Features entering the DGP continue to be identified as significant with high frequency, while features that are null in the DGP are generally identified as significant more frequently than in the regression setting.

The classification task appears to be more challenging than the regression task in this setting.
Whereas the regression model estimates a smooth conditional mean function, the classification model must learn a nonlinear decision boundary separating classes. 
Consequently, the fitted classifier is more sensitive to finite-sample variation and more likely to assign predictive value to features that are null in the DGP. 
This behavior helps explain why features that are null in the DGP are identified as significant somewhat more frequently than in the regression setting. 

The procedure remains stable under alternative choices of the loss function $L$, including the use of squared loss in place of cross-entropy loss.
Greater sensitivity arises with respect to the masking specification.
Specifically, using the mean rather than the adjusted mode for masking discrete features makes the Poisson-distributed feature $\ell=10$ less likely to be identified as significant.
This occurs because the mask often coincides with, or lies close to, the feature value being masked, rendering the masking operation less effective.
The effect is particularly pronounced in the classification setting, where the model output is compressed to $[0,1]$.

Replacing the conditional formulation~(\ref{conditional}) with the unconditional specification~(\ref{unconditional}) makes features $\ell=2,4,5$ less likely to be identified as significant and increases the frequency with which features that are null in the DGP are identified as significant.
In particular, features $\ell=4,5$ enter the response function through an interaction term, making their contribution harder to isolate under the unconditional formulation.
This behavior, also observed in the regression case (Section~\ref{sec-numerical}), highlights the importance of conditioning when features influence the response primarily through interactions.

Table~\ref{table-fo-classification} also reports results for CPT, HRT, LOCO, and the knockoff procedures.
AICO, CPT, HRT, and LOCO consistently identify as significant all features entering the DGP, while the knockoff approaches show modest difficulty identifying several such features.
LOCO, however, identifies as significant features that are null in the DGP substantially more frequently than the other methods.
A plausible explanation is that finite-sample estimation noise prevents the fitted model from being exactly invariant to DGP-null features.
Since LOCO compares predictions from two independently fitted models (the full model and a reduced model retrained without the candidate feature), small discrepancies between the fits may not fully cancel across test samples and can propagate into the test statistic.
The effect appears more pronounced in the classification setting, where the prediction problem is more difficult and the resulting finite-sample discrepancies between the fitted models may be larger. This mechanism may contribute to the higher frequency with which LOCO identifies as significant features that are null in the DGP.

\begin{table}[h]
\centering
\resizebox{0.5\columnwidth}{!}{
\renewcommand{\arraystretch}{1} 
\setlength{\tabcolsep}{4.5pt}      
\begin{tabular}{@{}lccccccccccccc@{}} 
\toprule
\multicolumn{13}{c}{\textbf{AICO}} \\ 
\toprule
\multirow{2}{*}{$\ell$}
  & \multicolumn{2}{c}{\textbf{$N=10^5$}}
  & \multicolumn{2}{c}{\textbf{$N=10^4$}}
  & \multicolumn{2}{c}{\textbf{$N=10^3$}}
  & \multicolumn{2}{c}{\textbf{$N=500$}}
  & \multicolumn{2}{c}{\textbf{$N=250$}}
  & \multicolumn{2}{c}{\textbf{$N=100$}} \\
\cmidrule(lr){2-3}
\cmidrule(lr){4-5}
\cmidrule(lr){6-7}
\cmidrule(lr){8-9}
\cmidrule(lr){10-11}
\cmidrule(lr){12-13}
  & 5\% & 1\% & 5\% & 1\% & 5\% & 1\% & 5\% & 1\% & 5\% & 1\% & 5\% & 1\% \\
\midrule
\textbf{1}  & 10 & 10 & 10 & 10 & 10 & 10 & 10 & 10 & 9  & 7  & 3 & 0 \\
\textbf{2}  & 10 & 10 & 10 & 10 & 6  & 5  & 3  & 2  & 6  & 5  & 3 & 1 \\
\textbf{3}  & 10 & 10 & 10 & 10 & 10 & 10 & 10 & 10 & 10 & 10 & 8 & 8 \\
\textbf{4}  & 10 & 10 & 10 & 10 & 10 & 10 & 10 & 9  & 6  & 2  & 4 & 0 \\
\textbf{5}  & 10 & 10 & 10 & 10 & 10 & 10 & 8  & 8  & 8  & 6  & 6 & 1 \\
\textbf{6}  & 10 & 10 & 10 & 10 & 10 & 10 & 10 & 10 & 8  & 6  & 2 & 1 \\
\textbf{7}  & 10 & 10 & 8  & 7  & 3  & 3  & 2  & 0  & 1  & 0  & 0 & 0 \\
\textbf{8}  & 10 & 10 & 5  & 2  & 0  & 0  & 0  & 0  & 0  & 0  & 0 & 0 \\
\textbf{9}  & 10 & 10 & 10 & 10 & 6  & 3  & 3  & 1  & 5  & 3  & 3 & 1 \\
\textbf{10} & 10 & 10 & 10 & 10 & 10 & 10 & 10 & 10 & 9  & 9  & 8 & 5 \\
\textbf{11} & 10 & 10 & 10 & 10 & 6  & 2  & 9  & 5  & 0  & 0  & 3 & 1 \\
\textbf{12} & 10 & 10 & 10 & 10 & 10 & 9  & 6  & 5  & 5  & 2  & 2 & 0 \\
\midrule
\textbf{13} & 0  & 0  & 1  & 0  & 1  & 0  & 3  & 0  & 0  & 0  & 0 & 0 \\
\textbf{14} & 0  & 0  & 0  & 0  & 0  & 0  & 1  & 0  & 1  & 0  & 2 & 1 \\
\textbf{15} & 0  & 0  & 0  & 0  & 0  & 0  & 0  & 0  & 1  & 1  & 1 & 0 \\
\textbf{16} & 0  & 0  & 0  & 0  & 0  & 0  & 0  & 0  & 0  & 0  & 2 & 0 \\
\textbf{17} & 1  & 1  & 1  & 0  & 1  & 0  & 2  & 0  & 0  & 0  & 0 & 0 \\
\textbf{18} & 1  & 1  & 1  & 0  & 1  & 0  & 2  & 0  & 0  & 0  & 0 & 0 \\
\textbf{19} & 0  & 0  & 1  & 0  & 1  & 0  & 0  & 0  & 0  & 0  & 0 & 0 \\
\bottomrule
\end{tabular}}
\caption{Classification model test results for 10 independent training/testing experiments, at two significance levels ($\alpha=5\%$ and $\alpha=1\%$), across various test sample sizes $N$. We report the total number of rejections over 10 trials. }
\label{table-fo-classification-N}
\end{table}

Table~\ref{table-fo-classification-N} reports test results for varying test-set sample sizes~$N$.
As $N$ decreases, identification rates for features entering the DGP decline more rapidly than in the regression case, particularly for features $\ell=7$ and~$\ell=8$.
The frequency with which features that are null in the DGP are identified as significant appears largely insensitive to~$N$.

As in the regression setting, the experiments above reference the DGP rather than the AICO null. A separate numerical demonstration of the finite-sample size property~(\ref{size}) is provided in Appendix~\ref{app-validity}.

Table~\ref{table-cis-class} presents a feature-importance ranking based on the empirical median~(\ref{emp-median}) for one randomly chosen training/testing experiment out of ten. 
It also reports the endpoints of the randomized confidence interval~(\ref{ci}) for the population median~$M$ at the $99\%$ level, together with the standard (non-randomized) two-sided interval~(\ref{naive-cis}) and their coverage probabilities. 
Table~\ref{table-cis-class} further shows the distribution of the randomized $p$-values.

\begin{table}[h]
\centering
\resizebox{0.75\columnwidth}{!}{
\renewcommand{\arraystretch}{1.1} 
\setlength{\tabcolsep}{2pt}       
\small
\begin{tabular}{@{}lcccccccccccc@{}}
\toprule
\multicolumn{12}{c}{\textbf{AICO, Level} $\alpha=1\%$} \\
\toprule
\multicolumn{4}{c}{} & \multicolumn{2}{c}{$p$-value Distribution} & \multicolumn{3}{c}{Randomized One-Sided CI} & \multicolumn{3}{c}{Non-Randomized Two-Sided CI} \\
\cmidrule(lr){5-6} \cmidrule(lr){7-9} \cmidrule(lr){10-12}
Rank & Feature & Median & P(reject) & Lower & Upper & 98.8\% & 1.2\% & Coverage & Lower & Upper & Coverage \\
\midrule
1  & $X_3$ & 3.577e-07 & 1.000 & 0.000 & 0.000 & 3.340e-07 & 3.340e-07 & 0.99 & 3.320e-07 & 3.829e-07 & 0.99007 \\
2  & $X_{10}$ & 2.324e-08 & 1.000 & 0.000 & 0.000 & 2.077e-08 & 2.077e-08 & 0.99 & 2.053e-08 & 2.619e-08 & 0.99007 \\
3  & $X_6$ & 6.722e-10 & 1.000 & 0.000 & 0.000 & 6.014e-10 & 6.016e-10 & 0.99 & 5.939e-10 & 7.656e-10 & 0.99007 \\
4  & $X_1$ & 1.630e-10 & 1.000 & 0.000 & 0.000 & 1.442e-10 & 1.443e-10 & 0.99 & 1.426e-10 & 1.862e-10 & 0.99007 \\
5  & $X_{12}$ & 6.989e-11 & 1.000 & 0.000 & 0.000 & 5.826e-11 & 5.828e-11 & 0.99 & 5.702e-11 & 8.597e-11 & 0.99007 \\
6  & $X_5$ & 4.339e-11 & 1.000 & 0.000 & 0.000 & 3.697e-11 & 3.697e-11 & 0.99 & 3.643e-11 & 5.135e-11 & 0.99007 \\
7  & $X_4$ & 3.169e-11 & 1.000 & 0.000 & 0.000 & 2.707e-11 & 2.707e-11 & 0.99 & 2.668e-11 & 3.733e-11 & 0.99007 \\
8  & $X_{11}$ & 1.001e-11 & 1.000 & 0.000 & 0.000 & 8.027e-12 & 8.030e-12 & 0.99 & 7.821e-12 & 1.275e-11 & 0.99007 \\
9  & $X_9$ & 6.787e-12 & 1.000 & 0.000 & 0.000 & 5.025e-12 & 5.025e-12 & 0.99 & 4.863e-12 & 9.439e-12 & 0.99007 \\
10 & $X_2$ & 2.575e-14 & 1.000 & 0.000 & 0.000 & 1.823e-14 & 1.825e-14 & 0.99 & 1.755e-14 & 3.704e-14 & 0.99007 \\
11 & $X_7$ & 1.100e-15 & 1.000 & 3.262e-71 & 3.431e-71 & 5.257e-16 & 5.261e-16 & 0.99 & 4.808e-16 & 2.326e-15 & 0.99007 \\
12 & $X_8$ & 1.344e-17 & 1.000 & 1.706e-14 & 1.743e-14 & 2.386e-18 & 2.394e-18 & 0.99 & 1.972e-18 & 4.921e-17 & 0.99007 \\
\midrule
-  & $X_{13}$ & 0.000      & 0.000 & 1.000 & 1.000 & 0.000       & 0.000       & 0.99 & 0.000       & 0.000       & 1.00000 \\
-  & $X_{14}$ & 0.000      & 0.000 & 1.000 & 1.000 & -1.396e-21  & -1.390e-21  & 0.99 & -2.680e-21  & 0.000       & 0.99485 \\
-  & $X_{15}$ & -3.193e-21 & 0.000 & 1.000 & 1.000 & -5.676e-20 & -5.672e-20 & 0.99 & -7.258e-20 & 0.000       & 0.99503 \\
-  & $X_{16}$ & 0.000      & 0.000 & 1.000 & 1.000 & -3.066e-22  & -3.052e-22  & 0.99 & -5.372e-22  & 0.000       & 0.99501 \\
-  & $X_{17}$ & 0.000      & 0.000 & 1.000 & 1.000 & 0.000       & 0.000       & 0.99 & 0.000       & 0.000       & 0.99992 \\
-  & $X_{18}$ & 0.000      & 0.000 & 0.940 & 0.940 & 0.000       & 0.000       & 0.99 & -1.849e-32  & 1.780e-22   & 0.99007 \\
-  & $X_{19}$ & 0.000      & 0.000 & 1.000 & 1.000 & 0.000       & 0.000       & 0.99 & 0.000       & 0.000       & 0.99999 \\
\bottomrule
\end{tabular}}
\caption{Classification model feature importance ranking according to the empirical median (\ref{emp-median}) for one of the independent training/testing experiments that was randomly chosen. The table also reports the endpoints of the randomized CI (\ref{ci}) for the median at a $99\%$ level, the standard (non-randomized) two-sided CI (\ref{naive-cis}) along with coverage probabilities, and the distribution of the randomized $p$-value. Percentage heads: probabilities of the two possible lower endpoints of the randomized CI.}
\label{table-cis-class}
\end{table}

As shown in Table~\ref{rank-table-class-narrow}, the AICO scores~(\ref{emp-median}) and importance rankings are relatively stable across independent training/testing experiments. Table~\ref{rank-table-class-narrow} also reports the corresponding rankings for CPT, HRT, and LOCO importance measures; Deep and Gaussian Knockoff importance measures; permutation importance; and Deep and Kernel SHAP importance measures. Compared with the regression setting, the rankings are somewhat more variable across experiments for all measures, reflecting the greater difficulty of the classification task. Across all measures, the symmetric roles of features~4 and~5 again lead to minor variation in their relative rankings across experiments. 

\begin{table}[h]
\centering
\resizebox{0.7\columnwidth}{!}{
\renewcommand{\arraystretch}{1.10}
\setlength{\tabcolsep}{4pt}
\scriptsize
\begin{tabular}{@{}lccccccccc@{}}
\toprule
\multirow{2}{*}{$\ell$}
& \multicolumn{1}{c}{\multirow{2}{*}{\textbf{AICO}}}
& \multicolumn{1}{c}{\multirow{2}{*}{\textbf{CPT}}}
& \multicolumn{1}{c}{\multirow{2}{*}{\textbf{HRT}}}
& \multicolumn{1}{c}{\multirow{2}{*}{\textbf{LOCO}}}
& \multicolumn{1}{c}{\textbf{Deep}}
& \multicolumn{1}{c}{\textbf{Gaussian}}
& \multicolumn{1}{c}{\textbf{Permutation}}
& \multicolumn{1}{c}{\textbf{Deep}}
& \multicolumn{1}{c}{\textbf{Kernel}} \\
& & & & & \multicolumn{1}{c}{\textbf{Knockoff}} & \multicolumn{1}{c}{\textbf{Knockoff}} & \multicolumn{1}{c}{\textbf{Importance}} & \multicolumn{1}{c}{\textbf{SHAP}} & \multicolumn{1}{c}{\textbf{SHAP}} \\
\midrule
\textbf{1} & 4 & 7 & 7 & 5\,{\fontsize{5.8pt}{6.8pt}\selectfont (5--7)} & 4\,{\fontsize{5.8pt}{6.8pt}\selectfont (3--4)} & 3 & 5 & 5 & 5 \\
\textbf{2} & 10 & 11 & 11 & 11\,{\fontsize{5.8pt}{6.8pt}\selectfont (9--12)} & 7 & 6\,{\fontsize{5.8pt}{6.8pt}\selectfont (6--7)} & 11 & 9\,{\fontsize{5.8pt}{6.8pt}\selectfont (9--10)} & 9\,{\fontsize{5.8pt}{6.8pt}\selectfont (9--11)} \\
\textbf{3} & 1 & 5 & 5 & 6\,{\fontsize{5.8pt}{6.8pt}\selectfont (5--6)} & 6 & 5 & 6 & 7 & 7 \\
\textbf{4} & 7\,{\fontsize{5.8pt}{6.8pt}\selectfont (6--9)} & 9\,{\fontsize{5.8pt}{6.8pt}\selectfont (8--10)} & 9\,{\fontsize{5.8pt}{6.8pt}\selectfont (8--10)} & 10\,{\fontsize{5.8pt}{6.8pt}\selectfont (7--11)} & 10\,{\fontsize{5.8pt}{6.8pt}\selectfont (9--12)} & 10\,{\fontsize{5.8pt}{6.8pt}\selectfont (9--12)} & 9\,{\fontsize{5.8pt}{6.8pt}\selectfont (8--10)} & 12\,{\fontsize{5.8pt}{6.8pt}\selectfont (11--12)} & 12\,{\fontsize{5.8pt}{6.8pt}\selectfont (10--12)} \\
\textbf{5} & 6\,{\fontsize{5.8pt}{6.8pt}\selectfont (5--8)} & 8\,{\fontsize{5.8pt}{6.8pt}\selectfont (8--9)} & 8\,{\fontsize{5.8pt}{6.8pt}\selectfont (8--9)} & 9\,{\fontsize{5.8pt}{6.8pt}\selectfont (8--11)} & 11\,{\fontsize{5.8pt}{6.8pt}\selectfont (9--12)} & 12\,{\fontsize{5.8pt}{6.8pt}\selectfont (10--12)} & 8\,{\fontsize{5.8pt}{6.8pt}\selectfont (8--9)} & 11\,{\fontsize{5.8pt}{6.8pt}\selectfont (11--12)} & 11\,{\fontsize{5.8pt}{6.8pt}\selectfont (11--12)} \\
\textbf{6} & 3 & 4 & 4 & 4 & 1\,{\fontsize{5.8pt}{6.8pt}\selectfont (1--2)} & 1 & 2 & 1\,{\fontsize{5.8pt}{6.8pt}\selectfont (1--2)} & 1\,{\fontsize{5.8pt}{6.8pt}\selectfont (1--2)} \\
\textbf{7} & 11 & 10\,{\fontsize{5.8pt}{6.8pt}\selectfont (8--10)} & 10\,{\fontsize{5.8pt}{6.8pt}\selectfont (8--10)} & 8\,{\fontsize{5.8pt}{6.8pt}\selectfont (7--12)} & 8\,{\fontsize{5.8pt}{6.8pt}\selectfont (8--9)} & 8\,{\fontsize{5.8pt}{6.8pt}\selectfont (7--9)} & 10\,{\fontsize{5.8pt}{6.8pt}\selectfont (8--10)} & 8 & 8 \\
\textbf{8} & 12 & 12 & 12 & 12\,{\fontsize{5.8pt}{6.8pt}\selectfont (10--12)} & 12\,{\fontsize{5.8pt}{6.8pt}\selectfont (10--12)} & 11\,{\fontsize{5.8pt}{6.8pt}\selectfont (9--12)} & 12 & 10\,{\fontsize{5.8pt}{6.8pt}\selectfont (9--10)} & 10\,{\fontsize{5.8pt}{6.8pt}\selectfont (9--10)} \\
\textbf{9} & 9\,{\fontsize{5.8pt}{6.8pt}\selectfont (7--9)} & 6 & 6 & 7\,{\fontsize{5.8pt}{6.8pt}\selectfont (6--11)} & 9\,{\fontsize{5.8pt}{6.8pt}\selectfont (8--12)} & 7\,{\fontsize{5.8pt}{6.8pt}\selectfont (6--9)} & 7 & 6 & 6 \\
\textbf{10} & 2 & 3 & 3 & 3 & 5 & 9\,{\fontsize{5.8pt}{6.8pt}\selectfont (6--12)} & 4 & 3\,{\fontsize{5.8pt}{6.8pt}\selectfont (3--4)} & 3 \\
\textbf{11} & 8\,{\fontsize{5.8pt}{6.8pt}\selectfont (6--9)} & 2 & 2 & 2 & 3\,{\fontsize{5.8pt}{6.8pt}\selectfont (3--4)} & 4 & 3 & 4\,{\fontsize{5.8pt}{6.8pt}\selectfont (3--4)} & 4 \\
\textbf{12} & 5\,{\fontsize{5.8pt}{6.8pt}\selectfont (5--7)} & 1 & 1 & 1 & 2\,{\fontsize{5.8pt}{6.8pt}\selectfont (1--2)} & 2 & 1 & 2\,{\fontsize{5.8pt}{6.8pt}\selectfont (1--2)} & 2\,{\fontsize{5.8pt}{6.8pt}\selectfont (1--2)} \\
\bottomrule
\end{tabular}}
\caption{Ranking stability across ten independent training/testing experiments. For each measure and trial, the 12 significant features are ranked by importance among themselves (1 = most important). Each cell reports the median rank across the 10 trials and, when the rank varies across trials, the corresponding range (Min--Max) in parentheses. Ties between features within a method are broken using the mean rank so that all 12 reported ranks are distinct.}
\label{rank-table-class-narrow}
\end{table}

The choice of loss function $L$ and masking specification can affect the scores (\ref{emp-median}) and the resulting feature importance rankings. 
The behavior is qualitatively similar to that in the regression case. 
In particular, switching from cross-entropy loss to squared loss reduces the magnitude of (\ref{emp-median}) but preserves the rankings. 
While not recommended, choosing the mean rather than the adjusted mode as the masking reference value for the discrete features $\ell=9,10$ also reduces the corresponding median values.



\begin{figure}[t]
\centering
\includegraphics[width=0.6\textwidth]{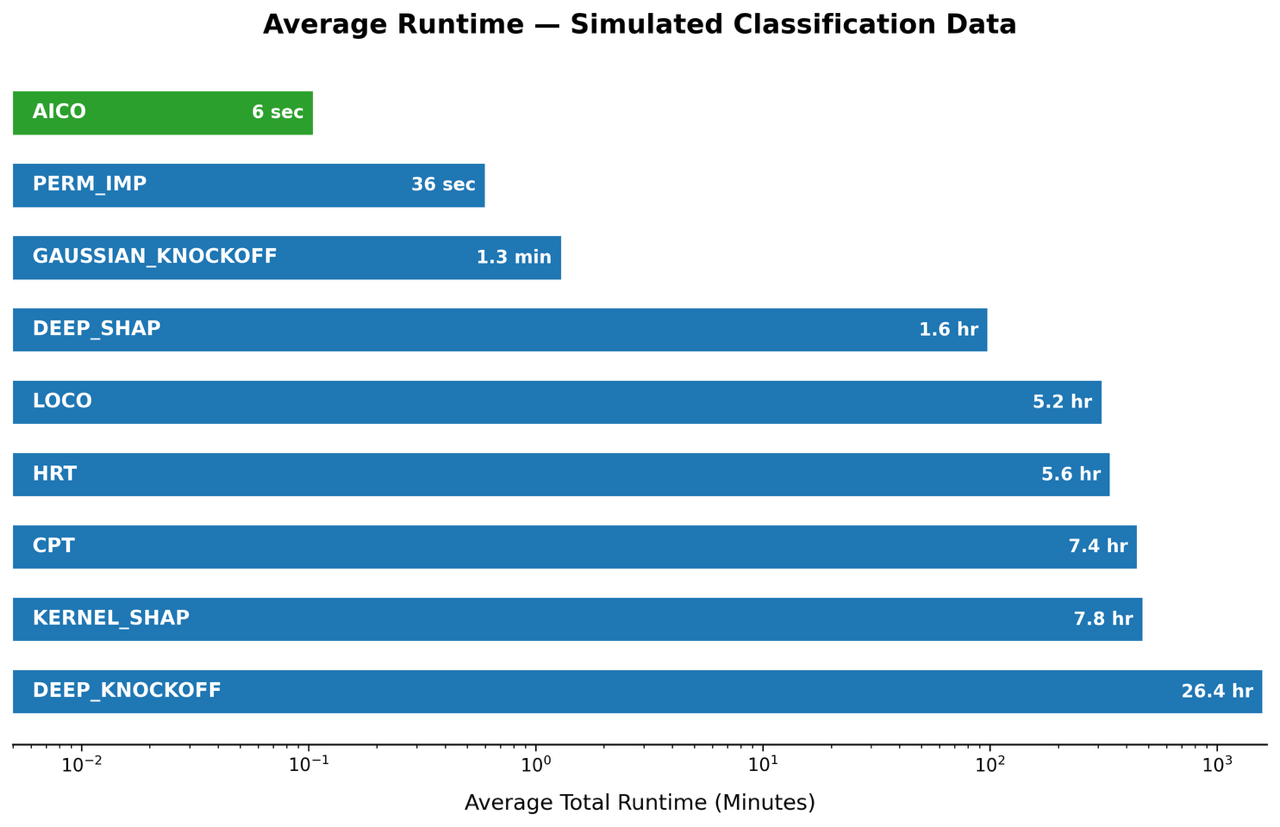}
\caption{Average computation times for testing/screening all 19 features (log-scale). The compute environment consists of 8GB memory and four CPU cores (Intel Xeon CPU E5-2670, 2.60GHz, and Silver 4110 CPU, 2.10GHz).}
\label{eff-class}
\end{figure}

Fig.~\ref{eff-class} reports average computation times across ten experiments. 
As in the regression case, AICO achieves exceptional computational efficiency. 
Among the testing procedures considered here, it is more than three orders of magnitude faster than LOCO, the next most efficient method. 
AICO is more than an order of magnitude faster than Gaussian knockoffs, the fastest screening approach among those implemented, and six times faster than permutation importance, the most efficient algorithmic feature-importance method considered in our experiments.

We also evaluate the procedure in a cross-fitting framework with five data splits. 
Table~\ref{table-cv-aico} below confirms that features entering the DGP continue to be identified as significant with high frequency under this more efficient data-splitting strategy, although features that are null in the DGP are identified as significant slightly more frequently.

\subsection{AICO test results for large number of trials}\label{app-regress}\label{robustness-large}

Table~\ref{table-robust-2000} reports AICO test results for 2,000 independent experiments for the regression task (Section~\ref{sec-numerical}) and the classification task (Appendix~\ref{classification-sect}). These results complement the results reported in columns two and three of Tables~\ref{table-fo-regression} and~\ref{table-fo-classification} for ten independent experiments. The Bonferroni-adjusted counts in panel~(b) demonstrate the effectiveness of standard multiplicity corrections in our setting. 

\begin{table}[h]
\centering
\begin{subtable}{0.34\columnwidth}
\centering
\resizebox{\columnwidth}{!}{
\renewcommand{\arraystretch}{1}
\setlength{\tabcolsep}{8pt}
\begin{tabular}{@{}lcccc@{}}
\toprule
\multicolumn{5}{c}{\textbf{AICO}} \\
\midrule
\multirow{3}{*}{$\ell$}
& \multicolumn{2}{c}{\textbf{Regression}}
& \multicolumn{2}{c}{\textbf{Classification}} \\
\cmidrule(lr){2-3} \cmidrule(lr){4-5}
& 5\% & 1\% & 5\% & 1\% \\
\midrule
\textbf{1}  & 2000 & 2000 & 2000 & 2000 \\
\textbf{2}  & 2000 & 2000 & 2000 & 2000 \\
\textbf{3}  & 2000 & 2000 & 1872 & 1872 \\
\textbf{4}  & 2000 & 2000 & 1851 & 1851 \\
\textbf{5}  & 2000 & 2000 & 1847 & 1846 \\
\textbf{6}  & 2000 & 2000 & 2000 & 2000 \\
\textbf{7}  & 2000 & 2000 & 2000 & 2000 \\
\textbf{8}  & 2000 & 2000 & 2000 & 2000 \\
\textbf{9}  & 2000 & 2000 & 2000 & 2000 \\
\textbf{10} & 2000 & 2000 & 2000 & 2000 \\
\textbf{11} & 2000 & 2000 & 2000 & 2000 \\
\textbf{12} & 2000 & 2000 & 2000 & 2000 \\
\midrule
\textbf{13} & 0    & 0    & 43   & 21   \\
\textbf{14} & 1    & 0    & 56   & 32   \\
\textbf{15} & 117  & 31   & 59   & 33   \\
\textbf{16} & 118  & 28   & 66   & 40   \\
\textbf{17} & 2    & 0    & 95   & 57   \\
\textbf{18} & 1    & 0    & 112  & 66   \\
\textbf{19} & 1    & 0    & 108  & 69   \\
\bottomrule
\end{tabular}}
\subcaption{Raw counts.}
\label{table-robust-2000}
\end{subtable}\hspace{0.04\columnwidth}\begin{subtable}{0.34\columnwidth}
\centering
\resizebox{\columnwidth}{!}{
\renewcommand{\arraystretch}{1}
\setlength{\tabcolsep}{8pt}
\begin{tabular}{@{}lcccc@{}}
\toprule
\multicolumn{5}{c}{\textbf{AICO}} \\
\midrule
\multirow{3}{*}{$\ell$}
& \multicolumn{2}{c}{\textbf{Regression}}
& \multicolumn{2}{c}{\textbf{Classification}} \\
\cmidrule(lr){2-3} \cmidrule(lr){4-5}
& 5\% & 1\% & 5\% & 1\% \\
\midrule
\textbf{1}  & 2000 & 2000 & 2000 & 2000 \\
\textbf{2}  & 2000 & 2000 & 2000 & 2000 \\
\textbf{3}  & 2000 & 2000 & 1872 & 1872 \\
\textbf{4}  & 2000 & 2000 & 1849 & 1847 \\
\textbf{5}  & 2000 & 2000 & 1845 & 1844 \\
\textbf{6}  & 2000 & 2000 & 2000 & 2000 \\
\textbf{7}  & 2000 & 2000 & 2000 & 2000 \\
\textbf{8}  & 2000 & 2000 & 2000 & 2000 \\
\textbf{9}  & 2000 & 2000 & 2000 & 2000 \\
\textbf{10} & 2000 & 2000 & 2000 & 2000 \\
\textbf{11} & 2000 & 2000 & 2000 & 2000 \\
\textbf{12} & 2000 & 2000 & 2000 & 2000 \\
\midrule
\textbf{13} & 0  & 0 & 12 & 11 \\
\textbf{14} & 0  & 0 & 22 & 13 \\
\textbf{15} & 8  & 1 & 27 & 19 \\
\textbf{16} & 14 & 3 & 35 & 30 \\
\textbf{17} & 0  & 0 & 43 & 30 \\
\textbf{18} & 0  & 0 & 46 & 29 \\
\textbf{19} & 0  & 0 & 51 & 34 \\
\bottomrule
\end{tabular}}
\subcaption{Bonferroni-adjusted at $\alpha/19$.}
\label{table-robust-2000-bonf}
\end{subtable}
\caption{Regression and classification AICO test results for $2{,}000$ independent experiments at two standard significance levels ($\alpha = 5\%$ and $\alpha = 1\%$). We report the total number of rejections (out of $2{,}000$) per feature $\ell$. Panel (a) shows the raw counts at nominal $\alpha$; panel (b) applies a Bonferroni adjustment with per-feature threshold $\alpha/19$. The training set size is $10^6$ and the
test set size is $N = 5 \cdot 10^5$. }
\label{table-robust-2000}
\end{table}

Several DGP-null features exhibit relatively elevated rejection counts. These include the correlated pair $(15,16)$ in the regression model and several features in the classification model. Moreover, in the classification model, several features entering the DGP have somewhat reduced rejection counts. An additional experiment suggests that variability induced by random initialization during training contributes to both phenomena. It also indicates that the elevated counts for the correlated pair $(15,16)$ are amplified by off-manifold extrapolation induced by marginal masking reference values.

The variability due to random initialization can be reduced through model averaging (ensembling), which generalizes the best-model strategy used previously (selecting the model, or random initialization, performing best on the validation set). The extrapolation issue arises for the correlated pair $(15,16)$ under marginal masking. Marginal masking sets $X^{15}$ to its mean while $X^{16}$ retains its observed value, sending the input into a region the training data cover only sparsely (since in training $X^{15}\approx 0.85\cdot X^{16}$). The model must then extrapolate, which can induce positive feature effects and increase the frequency with which the feature is identified as significant. This is fundamentally a fitting and sampling issue: a given realization of $D_1$ may cover that region especially poorly, causing the fitted model to extrapolate unreliably there. To address this issue, we replace the marginal-mean reference value with a conditional mean given $X^{-\ell}$, estimated by ordinary least squares of $X^\ell$ on the remaining features.\footnote{An alternative to the conditional mean is a nearest-neighbor estimator.}  This choice tends to keep the masked input in a more densely covered region of the feature space and substantially reduces the rejection counts on $X^{15}$ and $X^{16}$ while preserving the ability to identify the DGP-relevant features. 

Table~\ref{table-robust-2000-ens-condfitted} reports the 2,000-trial experiment with fresh training and test data drawn in each trial, using a best-three-of-five ensemble (five networks trained per trial from independent initializations, the three with the lowest validation loss averaged) and the conditional-mean reference value, in which each continuous feature is masked by its conditional mean given the remaining features, estimated from the training data by ordinary least squares. The results demonstrate that combining the ensembling strategy with the conditional-mean reference value substantially reduces the rejection counts for DGP-null features and increases the counts for features entering the DGP. As shown in panel~(b) of Table~\ref{table-robust-2000-ens-condfitted}, a standard Bonferroni adjustment further reduces rejection counts for DGP-null features while only minimally affecting rejection counts for features entering the DGP.

The rejection counts for DGP-null features in Table~\ref{table-robust-2000-ens-condfitted} are moderately higher in the classification setting than in the regression setting. This suggests that DGP-null features in the classification model more frequently induce median feature effects close to the AICO boundary value $M^\ell=0$. Near this boundary, the rejection probability approaches the nominal significance level, so relatively small shifts in the feature-effect distribution due to fitting and sampling variability can increase the frequency with which DGP-null features are identified as significant. A plausible explanation is that the classification problem is intrinsically more difficult than the regression problem, making the fitted model more sensitive to finite-sample variation.

\begin{table}[h]
\centering
\begin{subtable}[t]{0.34\columnwidth}
\centering
\resizebox{\columnwidth}{!}{
\renewcommand{\arraystretch}{1}
\setlength{\tabcolsep}{8pt}
\begin{tabular}{@{}lcccc@{}}
\toprule
\multicolumn{5}{c}{\textbf{AICO}} \\
\midrule
\multirow{3}{*}{$\ell$}
& \multicolumn{2}{c}{\textbf{Regression}}
& \multicolumn{2}{c}{\textbf{Classification}} \\
\cmidrule(lr){2-3} \cmidrule(lr){4-5}
& 5\% & 1\% & 5\% & 1\% \\
\midrule
\textbf{1}  & 2000 & 2000 & 2000 & 2000 \\
\textbf{2}  & 2000 & 2000 & 2000 & 1999 \\
\textbf{3}  & 2000 & 2000 & 2000 & 2000 \\
\textbf{4}  & 2000 & 2000 & 1996 & 1996 \\
\textbf{5}  & 2000 & 2000 & 1999 & 1998 \\
\textbf{6}  & 2000 & 2000 & 2000 & 2000 \\
\textbf{7}  & 2000 & 2000 & 2000 & 2000 \\
\textbf{8}  & 2000 & 2000 & 2000 & 2000 \\
\textbf{9}  & 2000 & 2000 & 2000 & 2000 \\
\textbf{10} & 2000 & 2000 & 2000 & 2000 \\
\textbf{11} & 2000 & 2000 & 2000 & 2000 \\
\textbf{12} & 2000 & 2000 & 2000 & 2000 \\
\midrule
\textbf{13} & 3 & 0 & 5  & 1  \\
\textbf{14} & 1 & 0 & 5  & 1  \\
\textbf{15} & 0 & 0 & 2  & 1  \\
\textbf{16} & 1 & 0 & 5  & 1  \\
\textbf{17} & 1 & 0 & 15 & 10 \\
\textbf{18} & 0 & 0 & 14 & 9  \\
\textbf{19} & 0 & 0 & 14 & 8  \\
\bottomrule
\end{tabular}}
\subcaption{Raw counts.}
\end{subtable}\hspace{0.04\columnwidth}\begin{subtable}[t]{0.34\columnwidth}
\centering
\resizebox{\columnwidth}{!}{
\renewcommand{\arraystretch}{1}
\setlength{\tabcolsep}{8pt}
\begin{tabular}{@{}lcccc@{}}
\toprule
\multicolumn{5}{c}{\textbf{AICO}} \\
\midrule
\multirow{3}{*}{$\ell$}
& \multicolumn{2}{c}{\textbf{Regression}}
& \multicolumn{2}{c}{\textbf{Classification}} \\
\cmidrule(lr){2-3} \cmidrule(lr){4-5}
& 5\% & 1\% & 5\% & 1\% \\
\midrule
\textbf{1}  & 2000 & 2000 & 2000 & 2000 \\
\textbf{2}  & 2000 & 2000 & 1999 & 1997 \\
\textbf{3}  & 2000 & 2000 & 2000 & 2000 \\
\textbf{4}  & 2000 & 2000 & 1996 & 1996 \\
\textbf{5}  & 2000 & 2000 & 1998 & 1998 \\
\textbf{6}  & 2000 & 2000 & 2000 & 2000 \\
\textbf{7}  & 2000 & 2000 & 2000 & 2000 \\
\textbf{8}  & 2000 & 2000 & 2000 & 2000 \\
\textbf{9}  & 2000 & 2000 & 2000 & 2000 \\
\textbf{10} & 2000 & 2000 & 2000 & 2000 \\
\textbf{11} & 2000 & 2000 & 2000 & 2000 \\
\textbf{12} & 2000 & 2000 & 2000 & 2000 \\
\midrule
\textbf{13} & 0 & 0 & 0 & 0 \\
\textbf{14} & 0 & 0 & 0 & 0 \\
\textbf{15} & 0 & 0 & 1 & 1 \\
\textbf{16} & 0 & 0 & 0 & 0 \\
\textbf{17} & 0 & 0 & 9 & 5 \\
\textbf{18} & 0 & 0 & 6 & 3 \\
\textbf{19} & 0 & 0 & 3 & 3 \\
\bottomrule
\end{tabular}}
\subcaption{Bonferroni-adjusted at $\alpha/19$.}
\label{table-robust-2000-ens-condfitted-bonf}
\end{subtable}
\caption{Regression and classification model AICO test results for $2{,}000$
independent training/testing experiments at two standard significance levels
($\alpha = 5\%$ and $\alpha = 1\%$). The model is a best-three-of-five ensemble, and masking uses the
OLS-estimated conditional-mean reference value. We report the total number of
rejections (out of $2{,}000$) per feature $\ell$. Panel~(a) shows the raw counts
at nominal $\alpha$; panel~(b) applies a Bonferroni adjustment with per-feature
threshold $\alpha/19$. The training set size is $10^6$ and the
test set size is $N = 5 \cdot 10^5$.}
\label{table-robust-2000-ens-condfitted}
\end{table}

\subsection{Comparative testing performance for large numbers of trials}\label{robustness-comp-large}

Tables~\ref{table-fo-regression} and~\ref{table-fo-classification} compare the performance of AICO, CPT, HRT, and LOCO across ten independent training/testing experiments. Table~\ref{table-fo-regression-1000} below reports results for 1,000 independent experiments for the regression model~(\ref{mu}). The setting mirrors that of Section~\ref{sec-numerical}, with one exception: to accommodate the substantially longer computation times of CPT, HRT, and LOCO (see Fig.~\ref{eff}), the training set size was reduced from $10^6$ to $10^4$ and the test set size was reduced from $5 \cdot 10^5$ to $N = 5 \cdot 10^3$. The results confirm the strong performance of the AICO procedure in a setting with relatively small training and testing samples.

All procedures identify as significant the features entering the DGP in every experiment, but there are substantial differences in how frequently features that are null in the DGP are identified as significant.
Among the methods considered, AICO exhibits the lowest such frequencies overall, by a considerable margin.
The corresponding frequencies for CPT and HRT are moderately higher, while those for LOCO are substantially higher, consistent with the behavior observed in the smaller-scale experiments (see Table~\ref{table-fo-classification}).

\begin{table}[h]
\centering
\resizebox{0.5\columnwidth}{!}{
\renewcommand{\arraystretch}{1}
\setlength{\tabcolsep}{4.5pt}
\begin{tabular}{@{}lcccccccccccc@{}}
\toprule
\multirow{2}{*}{$\ell$} & \multicolumn{2}{c}{\textbf{AICO}} & \multicolumn{2}{c}{\textbf{CPT}} & \multicolumn{2}{c}{\textbf{HRT}} & \multicolumn{2}{c}{\textbf{LOCO}} \\
\cmidrule(lr){2-3} \cmidrule(lr){4-5} \cmidrule(lr){6-7} \cmidrule(lr){8-9}
 & 5\% & 1\% & 5\% & 1\% & 5\% & 1\% & 5\% & 1\% \\
\midrule
\textbf{1} & 1000 & 1000 & 1000 & 1000 & 1000 & 1000 & 1000 & 1000 \\
\textbf{2} & 1000 & 1000 & 1000 & 1000 & 1000 & 1000 & 1000 & 1000 \\
\textbf{3} & 1000 & 1000 & 1000 & 1000 & 1000 & 1000 & 1000 & 1000 \\
\textbf{4} & 1000 & 1000 & 1000 & 1000 & 1000 & 1000 & 1000 & 1000 \\
\textbf{5} & 1000 & 1000 & 1000 & 1000 & 1000 & 1000 & 1000 & 1000 \\
\textbf{6} & 1000 & 1000 & 1000 & 1000 & 1000 & 1000 & 1000 & 1000 \\
\textbf{7} & 1000 & 1000 & 1000 & 1000 & 1000 & 1000 & 1000 & 1000 \\
\textbf{8} & 1000 & 1000 & 1000 & 1000 & 1000 & 1000 & 1000 & 1000 \\
\textbf{9} & 1000 & 1000 & 1000 & 1000 & 1000 & 1000 & 1000 & 1000 \\
\textbf{10} & 1000 & 1000 & 1000 & 1000 & 1000 & 1000 & 1000 & 1000 \\
\textbf{11} & 1000 & 1000 & 1000 & 1000 & 1000 & 1000 & 1000 & 1000 \\
\textbf{12} & 1000 & 1000 & 1000 & 1000 & 1000 & 1000 & 1000 & 1000 \\
\midrule
\textbf{13} & 7 & 1 & 59 & 14 & 59 & 14 & 98 & 49 \\
\textbf{14} & 3 & 2 & 42 & 13 & 41 & 9 & 97 & 48 \\
\textbf{15} & 49 & 10 & 53 & 14 & 45 & 8 & 118 & 64 \\
\textbf{16} & 45 & 9 & 74 & 17 & 60 & 13 & 125 & 62 \\
\textbf{17} & 5 & 1 & 76 & 19 & 49 & 9 & 101 & 59 \\
\textbf{18} & 3 & 1 & 68 & 10 & 55 & 13 & 90 & 44 \\
\textbf{19} & 8 & 2 & 77 & 13 & 46 & 13 & 95 & 48 \\
\bottomrule
\end{tabular}}\caption{Regression model test results for 1,000 independent training/testing experiments at two standard significance levels ($\alpha = 5\%$ and $\alpha = 1\%$). The table compares AICO, CPT, HRT, and LOCO procedures, showing the total number of rejections across 1,000 trials. The training set size is $10^4$ and the test set size is $N=5 \cdot 10^3$.}
\label{table-fo-regression-1000}
\end{table}

\subsection{Finite-sample validity of AICO test}\label{app-validity}
Equation~(\ref{size}) establishes the exact finite-sample validity of the AICO test:
conditional on the training data $D_1$, a feature with median effect
$M^\ell\le M_0$ is rejected with probability at most $\alpha$, with equality
attained only at the boundary $M^\ell=M_0$. We demonstrate this property
numerically for the regression task (Section~\ref{sec-numerical}) and the
classification task (Appendix~\ref{classification-sect}), using $M_0=0$.

The AICO test concerns a model-relative null. In the data-generating process
(DGP), a feature is null when the response does not depend on it; in our design
these are features $13$ through $19$. For the AICO test, by contrast, a feature
is null when the feature-effect distribution induced by the fitted model and
masking specification has median $M^\ell\le 0$, so that masking the feature does
not systematically degrade predictive performance. The two notions need not coincide. Trained on finitely many samples, the fitted model may attach a small positive effect to a DGP-null feature. Moreover, because the feature effect depends on the masking specification, a feature that is null in the DGP need not be AICO-null under a given masking scheme, even when the fitted model accurately represents the underlying response mechanism. In either case, the resulting feature-effect distribution may satisfy $M^\ell>0$. A rejection is then a false
discovery relative to the DGP null, but not relative to the AICO null. Since the size property applies only when $M^\ell\le 0$, we restrict the
demonstration to fitted models for which this condition holds.

For each DGP-null feature $\ell$, we repeatedly draw a training set $D_1$ of
$10^6$ samples, fit $f$ as in Section~\ref{sec-numerical}
and Appendix~\ref{classification-sect}, and estimate $M^\ell$ on a fresh sample of
$10^8$ points, which is sufficiently large to accurately estimate the sign of 
$M^\ell$. Separately for each feature, we retain the first $100$ fitted
models for which the estimated median satisfies $M^\ell\le 0$; this retained set
generally differs across features. For each retained model, we estimate its
rejection probability by applying the AICO test to $10{,}000$ independent test
sets of size $N=5\cdot10^5$ and recording the fraction of rejections. Masking
uses the same reference-value specification as in the main experiments. For each
feature, Table~\ref{table-validity} reports the mean, standard deviation, and
maximum of these estimated rejection probabilities across the retained models,
together with the number of estimates exceeding $\alpha$.

\begin{table}[h]
\centering
\resizebox{\columnwidth}{!}{
\renewcommand{\arraystretch}{1.1}
\setlength{\tabcolsep}{4pt}
\begin{tabular}{@{}l cccc cccc cccc cccc@{}}
\toprule
\multicolumn{17}{c}{\textbf{AICO}} \\
\midrule
\multirow{3}{*}{$\ell$}
& \multicolumn{8}{c}{\textbf{Regression}}
& \multicolumn{8}{c}{\textbf{Classification}} \\
\cmidrule(lr){2-9} \cmidrule(lr){10-17}
& \multicolumn{4}{c}{$\alpha = 5\%$} & \multicolumn{4}{c}{$\alpha = 1\%$}
& \multicolumn{4}{c}{$\alpha = 5\%$} & \multicolumn{4}{c}{$\alpha = 1\%$} \\
\cmidrule(lr){2-5} \cmidrule(lr){6-9} \cmidrule(lr){10-13} \cmidrule(lr){14-17}
& Mean & SD & Max & \#Exceed & Mean & SD & Max & \#Exceed
& Mean & SD & Max & \#Exceed & Mean & SD & Max & \#Exceed \\
\midrule
\textbf{13} & 0.0003 & 0.0002 & 0.0011 & 0 & 0.0000 & 0.0000 & 0.0002 & 0 & 0.0046 & 0.0081 & 0.0469 & 0 & 0.0007 & 0.0015 & 0.0105 & 1 \\
\textbf{14} & 0.0003 & 0.0002 & 0.0010 & 0 & 0.0000 & 0.0001 & 0.0004 & 0 & 0.0047 & 0.0081 & 0.0439 & 0 & 0.0006 & 0.0014 & 0.0081 & 0 \\
\textbf{15} & 0.0305 & 0.0138 & 0.0629 & 9 & 0.0055 & 0.0032 & 0.0146 & 9 & 0.0031 & 0.0071 & 0.0430 & 0 & 0.0004 & 0.0012 & 0.0082 & 0 \\
\textbf{16} & 0.0322 & 0.0134 & 0.0613 & 9 & 0.0059 & 0.0030 & 0.0127 & 10 & 0.0049 & 0.0089 & 0.0478 & 0 & 0.0008 & 0.0018 & 0.0107 & 1 \\
\textbf{17} & 0.0003 & 0.0002 & 0.0010 & 0 & 0.0000 & 0.0000 & 0.0002 & 0 & 0.0041 & 0.0069 & 0.0422 & 0 & 0.0005 & 0.0012 & 0.0082 & 0 \\
\textbf{18} & 0.0002 & 0.0002 & 0.0007 & 0 & 0.0000 & 0.0000 & 0.0002 & 0 & 0.0060 & 0.0112 & 0.0553 & 1 & 0.0009 & 0.0020 & 0.0107 & 1 \\
\textbf{19} & 0.0003 & 0.0002 & 0.0009 & 0 & 0.0000 & 0.0000 & 0.0002 & 0 & 0.0053 & 0.0093 & 0.0383 & 0 & 0.0008 & 0.0018 & 0.0090 & 0 \\
\bottomrule
\end{tabular}}
\caption{Numerical demonstration of the finite-sample validity of the AICO test.
For each DGP-null feature $\ell$, among the $100$ fitted models retained for that
feature because their estimated median effect satisfies $\hat{M}^\ell\le 0$, we report
the mean, standard deviation, and maximum of the estimated rejection probability,
each estimated over $10{,}000$ independent test sets, together with the number
of retained models whose estimated rejection probability exceeds $\alpha$
(\#Exceed). The training set size is $10^6$ and the test set
size is $N=5\cdot10^5$.}
\label{table-validity}
\end{table}

The results are consistent with the finite-sample size property in both
settings. For every feature, the mean estimated rejection probability is at or
below $\alpha$, and it is far smaller for fitted models whose estimated median
effects are well below zero. The few estimates that exceed $\alpha$ occur for
fitted models whose estimated median effects are close to the boundary
$M^\ell=0$. At this boundary, the rejection probability is equal to or close to
$\alpha$, so finite Monte Carlo estimation can produce rejection-frequency
estimates slightly above $\alpha$. Increasing the number of repeated test sets
reduces the magnitude of these exceedances but need not eliminate their
occurrence. No fitted model whose estimated median effect is well below zero
produces a rejection-frequency estimate above the nominal level, consistent with
the exact finite-sample validity of the test.

The relatively large number of exceedances for regression features $15$ and $16$
may also be related to their correlation in the data-generating process.
Correlation can make it more difficult for the fitted model to distinguish among
features, potentially leading to estimated median feature effects that lie
closer to the boundary $M^\ell=0$ and thereby increasing the frequency of Monte
Carlo exceedances.


\section{Multiple data splits: Cross-fitting}\label{cross-fitting}
Data splitting into training and testing sets (see Section~\ref{setup}) can reduce statistical efficiency and testing power. 
These issues can be mitigated by using multiple data splits, so that each sample is used for both training and testing. 
Our testing procedure extends naturally to this cross-fitting setting. 

Consider a partition of the data $\{Z_i\}$ into $K \ge 2$ disjoint folds of approximately equal size. 
For each $k \in [K]$, the folds excluding $k$ form the training set $I_{1k}$ used to construct the model $f_k$, while fold $k$ serves as the test set $I_{2k}$. 
Based on $I_{1k}$ and $f_k$, we construct fold-specific feature effects $\Delta^\ell_k(Z_i)$ for $i \in I_{2k}$. 
Several approaches can then be used to implement the testing procedure and construct a $p$-value for feature $\ell$; see \ci{tansey2018holdout,janson,guo-shah}, among others. 

One approach is to perform fold-specific tests on $\{\Delta^\ell_k(Z_i)\}_{i \in I_{2k}}$ for each $k \in [K]$, yielding $p$-values $p_1,\ldots,p_K$. 
A simple aggregation rule rejects the overall null if at least one fold-specific null is rejected, with corresponding $p$-value $\min(p_1,\ldots,p_K)$ or, with Bonferroni correction, $\min(K \cdot \min(p_1,\ldots,p_K), 1)$. 
These aggregated $p$-values inherit the validity of the fold-specific $p$-values, and their conditional distributions given the fold-specific test statistics follow from the uniform law~(\ref{unif}).

An alternative approach applies the testing procedure to the pooled set of feature effects $\bigcup_{k=1}^K \{\Delta^\ell_k(Z_i): i \in I_{2k}\}$, yielding a single decision and $p$-value following Section \ref{sec-test}. 
The theoretical validity of this approach is unclear, since the $\Delta^\ell_k(Z_i)$ may be dependent across folds due to overlapping training samples.


As the number of folds $K$ increases, both approaches fit the model in each fold using a larger fraction $(1-1/K)$ of the data, which can improve the quality of the fitted model---and hence the power of the test---by more faithfully capturing subtle feature-response relationships. Cross-fitting also mitigates ``split lottery'' effects, wherein atypical train-test partitions that are unrepresentative of the underlying population can arise by chance, thereby reducing variability across random train-test partitions. These benefits must be balanced against additional computational cost as $K$ grows. Moreover, under the first aggregation approach, larger $K$ also decreases the per-fold testing sample size and tightens the Bonferroni correction, which can reduce power. Tansey et al.~\ci{tansey2018holdout} suggest that in practice $K=5$ folds are sufficient to attain high power for both aggregation approaches.

\subsection{Numerical results}
We re-run the regression and classification experiments of Section~\ref{sec-numerical} and Appendix~\ref{classification-sect} in a cross-fitting setting with $K = 5$ folds on the full sample of $1.5\cdot 10^6$ observations. In each fold, the neural network model and reference values are constructed using the corresponding training partition, and the resulting feature effects are evaluated on the held-out fold, as previously described. All remaining configurations of the experiment---including the data-generating process, network architecture and training procedure, and feature effect specification---are identical to those used in Section~\ref{sec-numerical}. 

\begin{table}[h]
\centering
\resizebox{0.5\columnwidth}{!}{
\renewcommand{\arraystretch}{1}
\setlength{\tabcolsep}{8pt}
\begin{tabular}{@{}lcccccccc@{}}
\toprule
\multirow{4}{*}{$\ell$}
& \multicolumn{4}{c}{\textbf{Regression}}
& \multicolumn{4}{c}{\textbf{Classification}} \\
\cmidrule(lr){2-5} \cmidrule(lr){6-9}
& \multicolumn{2}{c}{\multirow{2}{*}{\textbf{AICO}}}
& \multicolumn{2}{c}{\textbf{Pooled}}
& \multicolumn{2}{c}{\multirow{2}{*}{\textbf{AICO}}}
& \multicolumn{2}{c}{\textbf{Pooled}} \\
& \multicolumn{2}{c}{}
& \multicolumn{2}{c}{\textbf{AICO}}
& \multicolumn{2}{c}{}
& \multicolumn{2}{c}{\textbf{AICO}} \\
\cmidrule(lr){2-3} \cmidrule(lr){4-5} \cmidrule(lr){6-7} \cmidrule(lr){8-9}
& 5\% & 1\% & 5\% & 1\% & 5\% & 1\% & 5\% & 1\% \\
\midrule
\textbf{1}  & 10 & 10 & 10 & 10 & 10 & 10 & 10 & 10 \\
\textbf{2}  & 10 & 10 & 10 & 10 & 10 & 10 & 10 & 10 \\
\textbf{3}  & 10 & 10 & 10 & 10 & 10 & 10 & 10 & 10 \\
\textbf{4}  & 10 & 10 & 10 & 10 & 10 & 10 & 10 & 10 \\
\textbf{5}  & 10 & 10 & 10 & 10 & 10 & 10 & 10 & 10 \\
\textbf{6}  & 10 & 10 & 10 & 10 & 10 & 10 & 10 & 10 \\
\textbf{7}  & 10 & 10 & 10 & 10 & 10 & 10 & 10 & 10 \\
\textbf{8}  & 10 & 10 & 10 & 10 & 10 & 10 & 10 & 10 \\
\textbf{9}  & 10 & 10 & 10 & 10 & 10 & 10 & 10 & 10 \\
\textbf{10} & 10 & 10 & 10 & 10 & 10 & 10 & 10 & 10 \\
\textbf{11} & 10 & 10 & 10 & 10 & 10 & 10 & 10 & 10 \\
\textbf{12} & 10 & 10 & 10 & 10 & 10 & 10 & 10 & 10 \\
\midrule
\textbf{13} & 0  & 0  & 0  & 0  & 0  & 0  & 0  & 0  \\
\textbf{14} & 0  & 0  & 0  & 0  & 1  & 1  & 0  & 0  \\
\textbf{15} & 0  & 0  & 1  & 0  & 1  & 1  & 0  & 0  \\
\textbf{16} & 0  & 0  & 3  & 1  & 1  & 1  & 0  & 0  \\
\textbf{17} & 0  & 0  & 0  & 0  & 1  & 1  & 0  & 0  \\
\textbf{18} & 0  & 0  & 0  & 0  & 4  & 4  & 3  & 2  \\
\textbf{19} & 0  & 0  & 0  & 0  & 2  & 2  & 2  & 1  \\
\bottomrule
\end{tabular}}
\caption{AICO test results for $5$-fold cross-fitting for the regression and classification tasks of Section~\ref{sec-numerical} and Appendix~\ref{classification-sect}, with a dataset of size $1.5\cdot 10^6$. The entries report the number of rejections (out of 10) for each feature $\ell\in[d]$ at levels $\alpha=5\%$ and $\alpha=1\%$. The column ``AICO'' reports results obtained under the scheme that aggregates separate fold-specific test results ($p$-values). The column ``Pooled AICO'' reports results obtained under the scheme that entails pooling of fold-specific feature effects.}
\label{table-cv-aico}
\end{table}

Table~\ref{table-cv-aico} reports results under the two aggregation schemes for ten independent trials at the $5\%$ and $1\%$ levels.
The results are similar to the single train-test split results reported in the first two columns of Tables~\ref{table-fo-regression} and~\ref{table-fo-classification}.
Across all trials, features entering the DGP ($\ell=1,\ldots,12$) are identified as significant under both aggregation schemes for both regression and classification.
Features that are null in the DGP are identified as significant only occasionally.
This modest increase in such identifications could be due to fold-to-fold estimation variability: fitting $K$ fold-specific models can occasionally yield suboptimal fits, and the resulting noise may propagate into the computed feature effects.
Increasing $K$ enlarges the training fraction $(1-1/K)$ within each fold and can mitigate such suboptimal-fit events, although for the first aggregation approach this must be weighed against smaller per-fold test sets and a tighter Bonferroni aggregation.
Overall, the results are similar across the two aggregation schemes, and neither exhibits a clear advantage in these experiments.



\section{Panel data}\label{panel}
This appendix explains how panel data, which may exhibit cross-sectional or time-series dependence, can be handled in AICO. 
Consider panel data $\{Z_{i,t}\}$, where $Z_{i,t} = (X_{i,t}, Y_{i,t})$, with $i = 1,\ldots,m$ indexing units and $t = 1,\ldots,T$ indexing time. 
We partition the index set $\{(i,t): i \in [m], t \in [T]\}$ into two disjoint subsets, $I_1$ and $I_2$, such that each pair belongs to exactly one subset. 
This setup accommodates common data-splitting schemes, including temporal splits (distinguishing between in-sample and out-of-sample time periods), unit-based splits (assigning each unit’s entire time series to either $I_1$ or $I_2$), or combinations thereof. 
For an unbalanced panel, in which units may be observed at different times, only observed pairs $(i,t)$ are assigned to $I_1$ or $I_2$.

\subsection{Conditioning}
Conditioning entails focusing on a suitable subset $I \subseteq I_2$ when constructing the feature effect.
If some samples in $I_2$ are dependent, we can restrict attention to the subset $I$ of independent samples $(i,t) \in I_2$. 
This reduces the test set size but helps avoid inference distortions due to unaccounted dependence, see \ci{cameron}, for example. For instance, in the presence of temporal dependence, one possible choice is to construct $I$ by restricting attention to samples at a given time $t$, yielding the subset $I = \{(\cdot,t) \in I_2\}$ and corresponding feature effects $\{\Delta^\ell(Z_{i,t}) : (i,t) \in I\}$. This approach yields a sequence of tests and a corresponding time series of $p$-values, importance scores, and confidence intervals, whose evolution over time may provide additional insight.

For constructing masking reference values using the training data (see Section \ref{general}), the training set $I_1$ is treated as a collection of $|I_1|$ samples—with unit and time information tied to each—fully exploiting the richness of the panel structure. 
For each $(i,t) \in I$, the  masking reference values can draw information not only from all training samples $I_1$ but also from testing samples corresponding to the same unit but not
in the considered subset, $\{X_{i,\cdot}: (i,\cdot) \in I_2\setminus I\}$. This flexibility enables unit-specific and time-dependent reference values, such as lagged feature values (e.g., the previous period’s feature value) or historical aggregations, thereby enhancing the contextual relevance of the reference value.

Conditioning is applicable beyond temporal dependence. For example, under spatial dependence, one may restrict attention to at most one observation per region or neighborhood to promote approximate independence, although the additional randomness induced by subsampling should be taken into account when interpreting the results.

\subsection{Grouping}\label{grouping}
An alternative grouping approach enables joint testing at the unit level, across time-series observations. 
The idea is to aggregate feature effects within a unit $i$ so that any time-series dependence is contained within groups. 
See \ci{williamson-longi} for feature importance analysis in this context and \ci{liang-zeger}, among others, for analysis of related approaches in a regression setting.

We treat the testing set $I_2$ as a collection of $|I_2|$ samples and compute the masked and unmasked predictions 
$f(m^\ell(X_{i,t}))$ and $f(u^\ell(X_{i,t}))$ for $(i,t)\in I_2$, 
where $m^\ell(X_{i,t})$ and $u^\ell(X_{i,t})$ are masked and unmasked feature vectors constructed as in Section \ref{setup}. 
We group the predictions and corresponding true responses by unit $i$ to obtain the unit-level feature effect
\begin{equation}\label{unit-level}
L^*(g_i^\ell, Y_i^*) - L^*(h_i^\ell, Y_i^*),
\end{equation}
where $L^*$ is a trajectory-level loss function, 
$g_i^\ell = \{f(m^\ell(X_{i,\cdot})) : (i,\cdot) \in I_2\}$, 
$h_i^\ell = \{f(u^\ell(X_{i,\cdot})) : (i,\cdot) \in I_2\}$, 
and $Y_i^* = \{Y_{i,\cdot} : (i,\cdot) \in I_2\}$. 
Unlike the setting of Section \ref{setup}, where the loss function $L$ evaluates one prediction against one true response, 
the trajectory-level loss function $L^*$ evaluates the entire sequence of predictions against the corresponding sequence of true responses.

Trajectory-level loss functions can extend familiar loss metrics. 
For instance, one may use the \emph{mean} squared error, averaging squared errors across time steps. 
More complex specifications can exploit richer temporal structure, such as correlation (Information Coefficient, IC), 
the error in maximum drawdown, 
or other time-series evaluations tailored to specific domains.

More broadly, the grouping approach extends beyond temporal dependence structures to cross-sectional and spatial dependence structures. 
For spatially dependent data, one may group testing samples by region or neighborhood, under the assumption that neighborhoods are approximately independent. 
Likewise, for repeated-measures data, grouping by the underlying entity or experimental context can provide an effective way to account for within-group dependence while preserving the assumptions needed for inference across groups.

\subsection{Hybrid approach}
Conditioning and grouping can be combined. 
For instance, suppose we wish to analyze a feature’s impact on model performance over an extended period. 
A longer period helps smooth short-term noise but makes single-period conditioning insufficient. 
Conversely, a shorter period preserves data relevance but may preclude grouping across testing time horizons. 
This limitation can be addressed by first conditioning on a time frame of suitable length and then grouping the conditioned data by units corresponding to each time–unit pair. 
As with conditioning alone, this procedure can be repeated across overlapping time windows, yielding a rolling-window analysis with smoothed single-period results.

Another notable use case of the hybrid method is assessing per-unit consistency of feature performance—how consistent over time is a feature’s impact on model behavior for a given unit? 
A longer window mitigates short-term fluctuations, and when the multi-period loss function aggregates single-period losses, it also captures an additional dimension of per-unit consistency.

For example, consider a multi-period loss function $L^*$ defined as the average of a single-period function, such as the mean squared error. 
In this case, the unit-level feature effect Eq.~(\ref{unit-level}) is effectively the average of the unit–time-level effects $\Delta^\ell(Z_{i,t})$ and thus reflects a degree of consistency over time. 
The longer the time window, the stronger the emphasis on consistency over time, as a higher bar is set for variation in feature impact.


\section{Missing data}\label{missing}
In practice, many datasets contain missing feature values, which are handled via a variety of deletion or imputation schemes. 
In general, when samples with missing or imputed feature values can be identified,\footnote{A common strategy is to augment the input vector with missingness-indicator features to distinguish imputed from observed values during model training (see, for example, \ci{vanness2023missing}). } we recommend excluding them from the test set in a feature-wise manner: when testing feature $\ell$, we drop only samples $i$ for which $X^\ell_i$ is missing---regardless of the missingness of other features---to preserve sample size as much as possible. This approach avoids ambiguity in the unmasking operation: $u^\ell(\cdot)$ is intended to reintroduce the information in feature $\ell$ into the model input, which is ill-defined when the true value $X^\ell_i$ is missing. Moreover, this feature-wise exclusion reduces potential distortion in the feature effects due to overlap between imputed values and masking reference values. Finally, feature-wise exclusion reduces the extent to which AICO results depend on the chosen imputation scheme, due to the possibility that different imputations can yield different inferences.

Under Missing Completely At Random (MCAR), wherein missingness is independent of the data, our approach does not systematically bias the resulting inference. While it may decrease power and increase variance due to a reduced effective test sample size, our approach can drop substantially fewer samples than common strategies such as listwise deletion, wherein any sample with at least one missing feature value is removed. When missingness depends on the data—e.g., under Missing At Random (MAR) or Missing Not At Random (MNAR)—excluding samples with missing $X^\ell_i$ can change the target estimand. Following  \ci{von2025feature}, the resulting inference can still be interpreted as ``observed-data feature importance'', which quantifies importance under the observed-data distribution. In some settings, this estimand may be more relevant than ``full-data feature importance''—defined under the hypothetical distribution with perfect observation—for example, during model development when the missingness pattern in the retrospective dataset is expected to persist after deployment.

Multiple imputation (MI) and inverse-probability weighting (IPW) are also widely used for handling missing data in estimation problems \cite{williamson2026assessing,von2025feature,seaman2013review}. For example, Williamson and Huang~\ci{williamson2024flexible} combine MI with an algorithm-agnostic variable-importance measure to enable variable selection under missingness with flexible learning algorithms beyond finite-dimensional statistical models. MI can also be incorporated into AICO by running AICO on each imputed dataset and then pooling inferences for the median feature-effect estimate across imputations using Rubin’s rules, leveraging the asymptotic normality of the sample median (via the Central Limit Theorem for medians). IPW can also be incorporated into AICO by reweighting the feature effects, but this would require generalizing the sign test to a weighted setting \cite{mcdonald2016exact,ahn2012relative,larocque2007weighted,kassam1977nonparametric}, which we leave for future work. 

While more sophisticated imputation can outperform simple deletion in some settings, several studies suggest this is not guaranteed in general. For example, Seijo-Pardo et al.~\ci{seijo2018analysis} emphasize that constant-value imputations (e.g., mean imputation) can distort covariance structure and thereby bias downstream feature-importance estimates and rankings, whereas complete-case analysis avoids such distortions by relying only on observed data. Additionally, the broader machine learning survey of \ci{emmanuel2021survey} notes that deletion-based methods are commonly used in practice due to their simplicity, low computational cost, and minimal modeling assumptions, despite the reduction in sample size. These considerations are particularly relevant for AICO, where computational efficiency and minimal assumptions are central.

Note that when applying the grouping approach in Appendix~\ref{grouping}, the exclusion can be performed at the prediction level---i.e., on $f(m^\ell(X_{i,t}))$, $f(u^\ell(X_{i,t}))$, and $Y_{i, t}$---rather than at the feature-effect level Eq.~(\ref{unit-level}), to preserve sample size. For example, in a panel setting, if unit $i$ has a missing value only at a single time step $t$ (i.e., $X^\ell_{i,t}$ is missing, while $X^\ell_{i,s}$ is observed for all $s\neq t$), it may suffice to drop only the corresponding predictions $f(m^\ell(X_{i,t}))$ and $f(u^\ell(X_{i,t}))$ and the associated response $Y_{i,t}$ from unit $i$’s grouped trajectory. In other words, when only one point is missing in the time series, we drop that time point rather than the entire trajectory.


Finally, each AICO test is conditional on the fitted model $f$, so any bias in or limitations of the fitted model can carry over to the test and feature-importance results.
Accordingly, missing data and the method used to handle it in the training set affect AICO primarily through model fitting. 
Van Ness et al.~\ci{vanness2023missing} show that the Missing Indicator Method coupled with mean imputation, as adopted in our empirical experiments in Section \ref{sec-empirical}, is a strong approach while being substantially more efficient than more complicated imputation schemes. 


\FloatBarrier
\bibliographystyle{pnas-new}
\bibliography{biblio}

@article{dai-shao-chen,
	author = {Dai, Guorong and Shao, Lingxuan and Chen, Jinbo},
	journal = {Journal of the Royal Statistical Society Series B: Statistical Methodology},
	pages = {816-832},
	title = {Moving beyond population variable importance: concept, theory and applications of individual variable importance},
	volume = {87},
	year = {2025}}

@article{williamson2024flexible,
	author = {Williamson, Brian D and Huang, Ying},
	journal = {The International Journal of Biostatistics},
	number = {2},
	pages = {347--359},
	publisher = {De Gruyter},
	title = {Flexible variable selection in the presence of missing data},
	volume = {20},
	year = {2024}}

@article{emmanuel2021survey,
	author = {Emmanuel, Tlamelo and Maupong, Thabiso and Mpoeleng, Dimane and Semong, Thabo and Mphago, Banyatsang and Tabona, Oteng},
	journal = {Journal of Big Data},
	number = {1},
	pages = {140},
	publisher = {Springer},
	title = {A survey on missing data in machine learning},
	volume = {8},
	year = {2021}}

@inproceedings{seijo2018analysis,
	author = {Seijo-Pardo, Borja and Alonso-Betanzos, Amparo and Bennett, Kristin P and Bol{\'o}n-Canedo, Ver{\'o}nica and Guyon, Isabelle and Josse, Julie and Saeed, Mehreen},
	booktitle = {European Symposium on Artificial Neural Networks, Computational Intelligence and Machine Learning, ESANN},
	title = {Analysis of imputation bias for feature selection with missing data},
	year = {2018}}

@article{kassam1977nonparametric,
	author = {Kassam, Saleem A and Thomas, John B},
	journal = {SIAM Journal on Applied Mathematics},
	number = {3},
	pages = {649--652},
	publisher = {SIAM},
	title = {Nonparametric Weighted-Signs Tests for Location},
	volume = {32},
	year = {1977}}

@article{larocque2007weighted,
	author = {Larocque, Denis and Nevalainen, Jaakko and Oja, Hannu},
	journal = {Biometrika},
	number = {2},
	pages = {267--283},
	publisher = {Oxford University Press},
	title = {A weighted multivariate sign test for cluster-correlated data},
	volume = {94},
	year = {2007}}

@article{ahn2012relative,
	author = {Ahn, Chul and Hu, Fan and Lee, Seung-Chun},
	journal = {Drug Information Journal},
	number = {4},
	pages = {428--433},
	publisher = {SAGE Publications Sage CA: Los Angeles, CA},
	title = {Relative efficiency of unequal versus equal cluster sizes for the nonparametric weighted sign test estimators in clustered binary data},
	volume = {46},
	year = {2012}}

@article{mcdonald2016exact,
	author = {McDonald, Janie and Gerard, Patrick D and McMahan, Christopher S and Schucany, William R},
	journal = {Journal of Agricultural, Biological and Environmental Statistics},
	number = {4},
	pages = {698--712},
	publisher = {Springer},
	title = {Exact-permutation-based sign tests for clustered binary data via weighted and unweighted test statistics},
	volume = {21},
	year = {2016}}

@article{seaman2013review,
	author = {Seaman, Shaun R and White, Ian R},
	journal = {Statistical Methods in Medical Research},
	number = {3},
	pages = {278--295},
	publisher = {SAGE Publications Sage UK: London, England},
	title = {Review of inverse probability weighting for dealing with missing data},
	volume = {22},
	year = {2013}}

@article{williamson2026assessing,
	author = {Williamson, Brian D and Krakauer, Chloe and Johnson, Eric and Gruber, Susan and Shepherd, Bryan E and van der Laan, Mark J and Lumley, Thomas and Lee, Hana and Hern{\'a}ndez-Mu{\~n}oz, Jos{\'e} J and Zhao, Fengyu},
	journal = {Statistics in Medicine},
	number = {3-5},
	pages = {e70366},
	publisher = {Wiley Online Library},
	title = {Assessing Treatment Effects in Observational Data With Missing Confounders: A Comparative Study of Practical Doubly-Robust and Traditional Missing Data Methods},
	volume = {45},
	year = {2026}}

@inproceedings{von2025feature,
	author = {Von Kleist, Henrik and Wendland, Joshua and Shpitser, Ilya and Marr, Carsten},
	booktitle = {Forty-second International Conference on Machine Learning, ICML},
	title = {Feature Importance Metrics in the Presence of Missing Data},
	year = {2025}}

@article{liang-zeger,
	author = {Kung-Yee Liang and Scott Zeger},
	journal = {Biometrika},
	number = {1},
	pages = {13-22},
	title = {Longitudinal data analysis using generalized linear models},
	volume = {73},
	year = {1986}}

@article{cameron,
	author = {A. Colin Cameron and Douglas L. Miller},
	journal = {Journal of Human Resources},
	number = {2},
	pages = {317-372},
	title = {A Practitioner's Guide to Cluster-Robust Inference},
	volume = {50},
	year = {2015}}

@article{williamson-longi,
	author = {Brian D. Williamson and Erica E.M. Moodie and Gregory E. Simon and Rebecca C. Rossom and Susan M. Shortreed},
	journal = {Annals of Applied Statistics},
	number = {2},
	pages = {1340--1363},
	title = {Inference on summaries of a model-agnostic longitudinal variable importance trajectory with application to suicide prevention},
	volume = {20},
	year = {2026}}

@article{guo-shah,
	author = {F. Richard Guo and Rajen D. Shah},
	journal = {Journal of the Royal Statistical Society Series B: Statistical Methodology},
	number = {1},
	pages = {256-286},
	title = {Rank-transformed subsampling: inference for multiple data splitting and exchangeable p-values},
	volume = {87},
	year = {2025}}

@conference{vanness2023missing,
	author = {Van Ness, Mike and Tomas M. Bosschieter and Roberto Halpin-Gregorio and Madeleine Udell},
	booktitle = {Proceedings of the 29th ACM SIGKDD Conference on Knowledge Discovery and Data Mining},
	pages = {5004--5015},
	title = {The Missing Indicator Method: From Low to High Dimensions},
	year = {2023}}

@article{pnas-test,
	author = {Zhanrui Cai and Jing Lei and Kathryn Roeder},
	journal = {Proceedings of the National Academy of Sciences USA},
	number = {34},
	pages = {e2205518119},
	title = {Model-free prediction test with application to genomics data},
	volume = {119},
	year = {2022}}

@article{cox-75,
	author = {D. R. Cox},
	journal = {Biometrika},
	number = {2},
	pages = {441-444},
	title = {A note on data-splitting for the evaluation of significance levels},
	volume = {62},
	year = {1975}}

@article{williamson2,
	author = {B. D. Williamson and P. B. Gilbert and M. Carone and N. Simon},
	journal = {Biometrics},
	pages = {9--22},
	title = {Nonparametric variable importance assessment using machine learning techniques},
	volume = {77},
	year = {2021}}

@article{williamson,
	author = {B. D. Williamson and P. B. Gilbert and N. R. Simon and M. Carone},
	journal = {Journal of the American Statistical Association},
	number = {543},
	pages = {1645-1658},
	title = {A general framework for inference on algorithm-agnostic variable importance},
	volume = {118},
	year = {2023}}

@article{gan,
	author = {L. Gan and L. Zheng and G. I. Allen},
	journal = {arXiv preprint arXiv:2206.02088},
	title = {{LOCO} Feature Importance Inference without Data Splitting via Minipatch Ensembles},
	year = {2026}}

@article{lundborg,
	author = {A. R. Lundborg and I. Kim and R. D. Shah and R. J. Samworth},
	journal = {Annals of Statistics},
	number = {6},
	pages = {2851--2878},
	title = {The projected covariance measure for assumption-lean variable significance testing},
	volume = {52},
	year = {2024}}

@article{rinaldo,
	author = {Alessandro Rinaldo and Larry Wasserman and Max G'Sell},
	journal = {Annals of Statistics},
	number = {6},
	pages = {3438--3469},
	title = {Bootstrapping and sample splitting for high-dimensional, assumption-lean inference},
	volume = {47},
	year = {2019}}

@article{mentch-hooker,
	author = {Lucas Mentch and Giles Hooker},
	journal = {Journal of Machine Learning Research},
	pages = {1-41},
	title = {Quantifying Uncertainty in Random Forests via Confidence Intervals and Hypothesis Tests},
	volume = {17},
	year = {2016}}

@article{mentch,
	author = {Tim Coleman and Wei Peng and Lucas Mentch},
	journal = {Journal of Machine Learning Research},
	pages = {1-35},
	title = {Scalable and Efficient Hypothesis Testing with Random Forests},
	volume = {23},
	year = {2022}}

@article{binyu,
	author = {W. J. Murdoch and C. Singh and K. Kumbier and R. Abbasi-Asl and B. Yu},
	journal = {Proceedings of the National Academy of Sciences USA},
	number = {44},
	pages = {22071-22080},
	title = {Definitions, methods, and applications in interpretable machine learning},
	volume = {116},
	year = {2019}}

@article{hooker,
	author = {Giles Hooker and Lucas Mentch and Siyu Zhou},
	journal = {Statistics and Computing},
	pages = {82},
	title = {Unrestricted permutation forces extrapolation: variable importance requires at least one more model, or there is no free variable importance},
	volume = {31},
	year = {2021}}

@article{burkart-huber,
	author = {Nadia Burkart and Marco F. Huber},
	journal = {Journal of Artificial Intelligence Research},
	pages = {245--317},
	title = {A Survey on the Explainability of Supervised Machine Learning},
	volume = {70},
	year = {2021}}

@inproceedings{elliot-seq,
	author = {Daniel Y. Fu and Elliot L. Epstein and Eric Nguyen and Armin W. Thomas and Michael Zhang and Tri Dao and Atri Rudra and Christopher R{\'e}},
	booktitle = {Proceedings of the 40th International Conference on Machine Learning},
	pages = {10373--10391},
	title = {Simple hardware-efficient long convolutions for sequence modeling},
	year = {2023}}

@article{dlmr,
	author = {Sadhwani, Apaar and Giesecke, Kay and Sirignano, Justin},
	journal = {Journal of Financial Econometrics},
	number = {2},
	pages = {313--368},
	publisher = {Oxford University Press},
	title = {Deep learning for mortgage risk},
	volume = {19},
	year = {2021}}

@article{elliot,
	author = {Epstein, Elliot and Sadhwani, Apaar and Giesecke, Kay},
	journal = {arXiv preprint arXiv:2505.11243},
	title = {A Set-Sequence Model for Time Series},
	year = {2025}}

@article{friedman2001greedy,
	author = {Friedman, Jerome H},
	journal = {Annals of Statistics},
	pages = {1189--1232},
	volume = {29},
	publisher = {JSTOR},
	title = {Greedy function approximation: a gradient boosting machine},
	year = {2001}}

@article{bergstra2011algorithms,
	author = {Bergstra, James and Bardenet, R{\'e}mi and Bengio, Yoshua and K{\'e}gl, Bal{\'a}zs},
	journal = {Advances in Neural Information Processing Systems},
	title = {Algorithms for hyper-parameter optimization},
	volume = {24},
	year = {2011}}

@inproceedings{akiba2019optuna,
	author = {Akiba, Takuya and Sano, Shotaro and Yanase, Toshihiko and Ohta, Takeru and Koyama, Masanori},
	booktitle = {Proceedings of the 25th ACM SIGKDD International Conference on Knowledge Discovery \& Data Mining},
	pages = {2623--2631},
	title = {Optuna: A next-generation hyperparameter optimization framework},
	year = {2019}}

@inproceedings{chen2016xgboost,
	author = {Chen, Tianqi and Guestrin, Carlos},
	booktitle = {Proceedings of the 22nd ACM SIGKDD International Conference on Knowledge Discovery and Data Mining},
	pages = {785--794},
	title = {{XGBoost}: A scalable tree boosting system},
	year = {2016}}

@article{benjamini1995controlling,
	author = {Benjamini, Yoav and Hochberg, Yosef},
	journal = {Journal of the Royal Statistical Society: Series B (Methodological)},
	number = {1},
	pages = {289--300},
	publisher = {Wiley Online Library},
	title = {Controlling the false discovery rate: a practical and powerful approach to multiple testing},
	volume = {57},
	year = {1995}}

@article{habiger2015multiple,
	author = {Habiger, Joshua D},
	journal = {Journal of Statistical Planning and Inference},
	pages = {1--13},
	publisher = {Elsevier},
	title = {Multiple test functions and adjusted p-values for test statistics with discrete distributions},
	volume = {167},
	year = {2015}}

@article{cousido2022multiple,
	author = {Cousido-Rocha, Marta and de U{\~n}a-{\'A}lvarez, Jacobo and D{\"o}hler, Sebastian},
	journal = {Journal of the Royal Statistical Society Series C: Applied Statistics},
	number = {1},
	pages = {219--243},
	publisher = {Oxford University Press},
	title = {Multiple comparison procedures for discrete uniform and homogeneous tests},
	volume = {71},
	year = {2022}}

@article{kulinskaya2009fuzzy,
	author = {Kulinskaya, Elena and Lewin, Alex},
	journal = {Biometrika},
	number = {1},
	pages = {201--211},
	publisher = {Oxford University Press},
	title = {On fuzzy familywise error rate and false discovery rate procedures for discrete distributions},
	volume = {96},
	year = {2009}}

@inproceedings{lundberg2017unified,
	author = {Lundberg, Scott M and Lee, Su-In},
	booktitle = {Advances in Neural Information Processing Systems},
	pages = {4765--4774},
	title = {A unified approach to interpreting model predictions},
	year = {2017}}

@article{geyer-meeden,
	author = {Charles J. Geyer and Glen D. Meeden},
	issue = {4},
	journal = {Statistical Science},
	pages = {358--366},
	title = {Fuzzy and Randomized Confidence Intervals and P-Values},
	volume = {20},
	year = {2005}}

@inproceedings{bellot,
	author = {Bellot, Alexis and van der Schaar, Mihaela},
	booktitle = {Proceedings of the 33rd International Conference on Neural Information Processing Systems},
	pages = {2202--2211},
	title = {Conditional independence testing using generative adversarial networks},
	year = {2019}}

@article{janson,
	author = {Molei Liu and Eugene Katsevich and Lucas Janson and Aaditya Ramdas},
	issue = {2},
	journal = {Biometrika},
	pages = {277--293},
	title = {Fast and powerful conditional randomization testing via distillation},
	volume = {109},
	year = {2022}}

@article{wright,
	author = {David S. Watson and Marvin N. Wright},
	journal = {Machine Learning},
	pages = {2107--2129},
	title = {Testing Conditional Independence in Supervised Learning Algorithms},
	volume = {110},
	year = {2021}}

@article{bates,
	author = {Stephen Bates and Emmanuel Cand\`es and Lucas Janson and Wenshuo Wang},
	journal = {Journal of the American Statistical Association},
	number = {535},
	pages = {1413-1427},
	title = {Metropolized knockoff sampling},
	volume = {116},
	year = {2021}}

@article{fisher-rudin,
	author = {Aaron Fisher and Cynthia Rudin and Francesca Dominici},
	journal = {Journal of Machine Learning Research},
	number = {177},
	pages = {1-81},
	title = {All Models are Wrong, but Many are Useful: Learning a Variable's Importance by Studying an Entire Class of Prediction Models Simultaneously},
	volume = {20},
	year = {2019}}

@article{breiman,
	author = {Breiman, Leo},
	journal = {Machine Learning},
	number = {1},
	pages = {5-32},
	title = {Random Forests},
	volume = {45},
	year = {2001}}

@article{hama-mase-owen,
	author = {N. Hama and M. Mase and A. B. Owen},
	journal = {arXiv preprint arXiv:2211.08414},
	title = {Model free variable importance for high dimensional data},
	year = {2023}}

@book{asmussen-glynn,
	author = {Soren Asmussen and Peter Glynn},
	publisher = {Springer},
	title = {Stochastic Simulation: Algorithms and Analysis},
	year = {2007}}

@article{zielinski,
	author = {R. Zieli{\'n}ski and W. Zieli{\'n}ski},
	journal = {Statistics},
	number = {1},
	pages = {67-71},
	title = {Best exact nonparametric confidence intervals for quantiles},
	volume = {39},
	year = {2005}}

@book{efron,
	author = {Brad Efron},
	publisher = {SIAM},
	title = {The jackknife, the bootstrap, and other resampling plans},
	year = {1982}}

@article{falk-kaufmann,
	author = {Michael Falk and Edgar Kaufmann},
	journal = {Annals of Statistics},
	number = {1},
	pages = {485-495},
	title = {Coverage Probabilities of Bootstrap-Confidence Intervals for Quantiles},
	volume = {19},
	year = {1991}}

@article{chen-hall,
	author = {Song Xi Chen and Peter Hall},
	journal = {Annals of Statistics},
	number = {3},
	pages = {1166-1181},
	title = {Smoothed Empirical Likelihood Confidence Intervals for Quantiles},
	volume = {21},
	year = {1993}}

@article{wittkowski,
	author = {Knut M. Wittkowski},
	journal = {Journal of the Royal Statistical Society Series D: The Statistician},
	number = {2},
	pages = {93--96},
	title = {An Asymptotic {UMP} Sign Test for Discretised Data},
	volume = {38},
	year = {1989}}

@book{lehmann,
	author = {Erich Lehmann},
	publisher = {Springer},
	title = {Nonparametrics: {Statistical Methods Based on Ranks}},
	year = {2006}}

@article{deepknockoffs,
	author = {Romano, Yaniv and Sesia, Matteo and Cand\`es, Emmanuel J.},
	journal = {Journal of the American Statistical Association},
	number = {532},
	pages = {1861--1872},
	title = {Deep knockoffs},
	volume = {115},
	year = {2020}}

@article{noether,
	author = {G. Noether},
	journal = {Journal of the American Statistical Association},
	pages = {645-647},
	title = {Sample Size Determination for Some Common Nonparametric Tests},
	volume = {82},
	year = {1987}}

@article{lavergne2015,
	author = {Lavergne, P. and Maistre, S. and Patilea, V.},
	journal = {Electronic Journal of Statistics},
	pages = {643--678},
	title = {A significance test for covariates in nonparametric regression},
	volume = {9},
	year = {2015}}

@article{ait-sahalia-bickel,
	author = {Yacine A{\"i}t-Sahalia and Peter Bickel and Thomas M. Stoker},
	issue = {2},
	journal = {Journal of Econometrics},
	pages = {363-412},
	title = {Goodness-of-fit tests for kernel regression with an application to option implied volatilities},
	volume = {105},
	year = {2001}}

@article{delgado-manteiga,
	author = {Delgado, M. A. and Gonz{\'a}lez Manteiga, W.},
	journal = {Annals of Statistics},
	pages = {1469--1507},
	publisher = {JSTOR},
	title = {Significance testing in nonparametric regression based on the bootstrap},
	volume = {29},
	year = {2001}}

@article{fan-li,
	author = {Fan, Yanqin and Li, Qi},
	journal = {Econometrica},
	pages = {865--890},
	publisher = {JSTOR},
	title = {Consistent model specification tests: omitted variables and semiparametric functional forms},
	volume = {64},
	year = {1996}}

@article{horel-giesecke,
	author = {Enguerrand Horel and Kay Giesecke},
	journal = {Journal of Machine Learning Research},
	pages = {1--29},
	title = {Significance Tests for Neural Networks},
	volume = {21},
	year = {2020}}

@article{fallahgoul,
	author = {Hasan Fallahgoul and Vincentius Franstianto and Xin Lin},
	issue = {1},
	journal = {Journal of Econometrics},
	pages = {105574},
	title = {Asset pricing with neural networks: Significance tests},
	volume = {238},
	year = {2024}}

@article{berrett,
	author = {Thomas Berrett and Yi Wang and Rina Foygel Barber and Richard Samworth},
	journal = {Journal of the Royal Statistical Society Series B: Statistical Methodology},
	number = {1},
	pages = {175--197},
	title = {The conditional permutation test for independence while controlling for confounders},
	volume = {82},
	year = {2020}}

@article{verdinelli-wasserman,
	author = {Isabella Verdinelli and Larry Wasserman},
	journal = {Statistical Science},
	number = {4},
	pages = {623-636},
	title = {Feature Importance: A Closer Look at {Shapley} Values and {LOCO}},
	volume = {39},
	year = {2024}}

@article{Lei2018predictive,
	author = {Jing Lei and Max G'Sell and Alessandro Rinaldo and Ryan J. Tibshirani and Larry Wasserman},
	journal = {Journal of the American Statistical Association},
	number = {523},
	pages = {1094-1111},
	publisher = {Taylor & Francis},
	title = {Distribution-Free Predictive Inference for Regression},
	volume = {113},
	year = {2018}}

@inproceedings{ribeiro2016should,
	author = {Ribeiro, Marco Tulio and Singh, Sameer and Guestrin, Carlos},
	booktitle = {Proceedings of the 22nd ACM SIGKDD international conference on knowledge discovery and data mining},
	organization = {ACM},
	pages = {1135--1144},
	title = {Why should {I} trust you?: Explaining the predictions of any classifier},
	year = {2016}}

@article{white2001statistical,
	author = {White, Halbert and Racine, Jeffrey},
	journal = {IEEE Transactions on Neural Networks},
	number = {4},
	pages = {657--673},
	publisher = {IEEE},
	title = {Statistical inference, the bootstrap, and neural-network modeling with application to foreign exchange rates},
	volume = {12},
	year = {2001}}

@article{lavergne2000nonparametric,
	author = {Lavergne, Pascal and Vuong, Quang},
	journal = {Econometric Theory},
	number = {4},
	pages = {576--601},
	publisher = {Cambridge University Press},
	title = {Nonparametric significance testing},
	volume = {16},
	year = {2000}}

@article{candes2018panning,
	author = {Cand\`es, Emmanuel and Fan, Yingying and Janson, Lucas and Lv, Jinchi},
	journal = {Journal of the Royal Statistical Society Series B: Statistical Methodology},
	number = {3},
	pages = {551--577},
	publisher = {Wiley Online Library},
	title = {Panning for gold: `model-{X}' knockoffs for high dimensional controlled variable selection},
	volume = {80},
	year = {2018}}

@article{tansey2018holdout,
	author = {Tansey, Wesley and Veitch, Victor and Zhang, Haoran and Rabadan, Raul and Blei, David M},
	journal = {Journal of Computational and Graphical Statistics},
	number = {1},
	pages = {151--162},
	title = {The holdout randomization test for feature selection in black box models},
	volume = {31},
	year = {2022}}

@book{lehmann2006testing,
	author = {Lehmann, Erich and Romano, Joseph},
	publisher = {Springer},
	title = {Testing Statistical Hypotheses},
	year = {2005}}

\end{document}